\documentclass[%
 reprint,
superscriptaddress,
 amsmath,amssymb,
 aps,
 prx,
]{revtex4-2}

\usepackage{graphicx}
\usepackage{dcolumn}
\usepackage{bm}

\usepackage{mathtools}

\DeclareMathOperator*{\argmax}{argmax}

\begin{document}

\preprint{APS/123-QED}

\title{Conditioned quantum-assisted deep generative surrogate \\ for particle-calorimeter interactions}%

\author{J. Quetzalcoatl Toledo-Marin}
\affiliation{TRIUMF, Vancouver, BC V6T 2A3, Canada
}%
\affiliation{
Perimeter Institute for Theoretical Physics,
Waterloo, Ontario, N2L 2Y5, Canada
}%
\author{Sebastian Gonzalez}
\thanks{These authors contributed equally and are listed in alphabetical order.}%
\affiliation{%
 TRIUMF, Vancouver, BC V6T 2A3, Canada
}%
\author{Hao Jia}
\thanks{These authors contributed equally and are listed in alphabetical order.}%
\affiliation{%
 Department of Physics and Astronomy, University of British Columbia, \\
 Vancouver, BC V6T 1Z1, Canada
}%
\author{Ian Lu}
\thanks{These authors contributed equally and are listed in alphabetical order.}%
\affiliation{%
 TRIUMF, Vancouver, BC V6T 2A3, Canada
}%
\author{Deniz Sogutlu}
\affiliation{%
 TRIUMF, Vancouver, BC V6T 2A3, Canada
}%
\author{Abhishek Abhishek}
\affiliation{Department of Electrical and Computer Engineering, \\
University of British Columbia, Vancouver, BC V6T 1Z4, Canada}
\author{Colin Gay}
\affiliation{%
 Department of Physics and Astronomy, University of British Columbia, \\
 Vancouver, BC V6T 1Z1, Canada
}%
\author{Eric Paquet}
\affiliation{Digital Technologies Research Centre,
National Research Council, \\
1200 Montreal Road, Building M-50,
Ottawa, Ontario, K1A 0R6, Canada}
\author{Roger Melko}
\affiliation{
Perimeter Institute for Theoretical Physics,
Waterloo, Ontario, N2L 2Y5, Canada
}%
\author{Geoffrey C. Fox}
\affiliation{%
 University of Virginia,
 Computer Science and Biocomplexity Institute, \\
 994 Research Park Blvd, Charlottesville,
 Virginia, 22911, USA
}%
\author{Maximilian Swiatlowski}
\affiliation{%
 TRIUMF, Vancouver, BC V6T 2A3, Canada
}%
\author{Wojciech Fedorko}
\affiliation{%
 TRIUMF, Vancouver, BC V6T 2A3, Canada
}%




\date{\today}

\begin{abstract}
Particle collisions at accelerators such as the Large Hadron Collider, recorded and analyzed by experiments such as ATLAS and CMS, enable exquisite measurements of the Standard Model and searches for new phenomena. Simulations of collision events at these detectors have played a pivotal role in shaping the design of future experiments and analyzing ongoing ones. However, the quest for accuracy in Large Hadron Collider (LHC) collisions comes at an imposing computational cost, with projections estimating the need for millions of CPU-years annually during the High Luminosity LHC (HL-LHC) run \cite{collaboration2022atlas}. Simulating a single LHC event with \textsc{Geant4} currently devours around 1000 CPU seconds, with simulations of the calorimeter subdetectors in particular imposing substantial computational demands \cite{rousseau2023experimental}. To address this challenge, we propose a conditioned quantum-assisted deep generative model. Our model integrates a conditioned variational autoencoder (VAE) on the exterior with a conditioned Restricted Boltzmann Machine (RBM) in the latent space, providing enhanced expressiveness compared to conventional VAEs. The RBM nodes and connections are meticulously engineered to enable the use of qubits and couplers on D-Wave's Pegasus-structured \textit{Advantage} quantum annealer (QA) for sampling. We introduce a novel method for conditioning the quantum-assisted RBM using \textit{flux biases}. We further propose a novel adaptive mapping to estimate the effective inverse temperature in quantum annealers. The effectiveness of our framework is illustrated using Dataset 2 of the CaloChallenge \cite{calochallenge}. 
\end{abstract}

\maketitle


\section{Introduction}

By the end of the decade, the LHC is expected to begin an upgraded ``High Luminosity'' phase, which will ultimately increase the collision rate by a factor of 10 higher than the initial design. Increasing the number of collisions will generate more experimental data, enabling the observation of rare processes and increased precision in measurements, furthering our understanding of the universe.
The path toward the HL-LHC presents great technological challenges and commensurate innovations to overcome them.
Monte Carlo simulations of collision events at the ATLAS experiment have played a key role in the design of future experiments and, particularly, in the analysis of current ones. 
However, these simulations are computationally intensive, projected to reach into millions of CPU-years per year during the HL-LHC run \cite{collaboration2022atlas}.
Simulating a single event with \textsc{Geant4} \cite{agostinelli2003geant4} in an LHC experiment requires roughly 1000 CPU seconds. The calorimeter simulation is by far dominating the total simulation time \cite{rousseau2023experimental}. To address this challenge, deep generative models are being developed to act as particle-calorimeter interaction surrogates, with the potential to reduce the simulation overall time by orders of magnitude. The key point to take into consideration is that one particle impacting a calorimeter can lead to thousands of secondary particles, collectively known as showers, to be traced through the detector, while only the total energy deposit per sensitive element (a cell) is actually measured in the experiment. Hence, through the generation of these showers, non-negligible computational resources are being employed in the detailed recording of the path of these particles. The problem being addressed is whether one can bypass the path-tracing step in the simulation and generate the cell energy deposits directly from a set of well-defined parameters (\textit{e.g.}, type of particle, incidence energy, incidence angle, \textit{etc.}) via sampling from a deep generative model. 

There is a large and growing body of literature addressing this critical problem via deep generative models. The earliest methods developed to address this challenge were of the kind of Generative Adversarial Networks \cite{de2017learning, paganini2018accelerating, paganini2018calogan}, which are now an integral part of the simulation pipeline \cite{atlas2020fast, aad2022atlfast3} of some experiments. Different deep generative frameworks have been proposed since then, including VAEs \cite{buhmann2021decoding, buhmann2021decoding, atlas2022deep, salamani2023metahep}, Normalizing Flows \cite{krause2021caloflow, buckley2024inductive}, Transformers \cite{favaro2024calodream}, Diffusion models \cite{ mikuni2024caloscore,kobylianskii2024calograph, liu2024calo} and combinations thereof \cite{amram2023denoising, madulacalolatent}. A noteworthy development in the field is the CaloChallenge-2022 endeavour \cite{krause2024calochallenge}, which not only catalyzed research efforts but has also enabled better quantitative comparison across different frameworks. Similarly, defining benchmarks and metrics has been a rather active topic of research \cite{kansal2023evaluating, ahmad2024comprehensive}. A common feature of these models is the fast generation of showers via sampling on GPUs, but further speed increases may still be possible with alternative computing paradigms. In our work, we develop deep generative models which can be naturally encoded onto Quantum Annealers (QAs), allowing for potentially significant improvements in the speed of shower simulation.

In previous work by this group \cite{hoque2023caloqvae}, a proof of concept of a quantum-assisted deep discrete variational autoencoder calorimeter surrogate called CaloQVAE was presented, whereby the performance in synthetic data generation is similar to its classical counterpart. As a follow up, in Ref \cite{gonzalez2024caloqvae}, we improved our framework by re-engineering the encoder and decoder architectures by introducing two-dimensional convolutions as well as replacing the two-partite graph in the RBM with four-partite graph, enabling us to use D-wave’s quantum annealer AdvantageSystem \cite{boothby2020next}. 
The main contributions of this paper are: \textit{i)} we condition the prior on specific incident energies which allows us to disentangle the latent space embeddings. Most relevant is that we propose a novel method to enforce this condition scheme on the qubits by means of the QA's \textit{flux bias} parameters. \textit{ii)} We also propose a novel adaptive method to estimate the effective temperature in the quantum annealer. This method is a simple map with a stable fixed point on the effective temperature, and we show it to be more stable with a faster convergence than the KL-divergence-based method used previously. \textit{iii)} The new model uses 3D convolutional layers for the encoder and decoder as well as periodic boundaries to account for the cylindrical geometry of the shower, leading to an improved performance when compared to its previous counterpart. We illustrate our framework by using Dataset 2 of the CaloChallenge.
Henceforth, we refer to our framework as Calo4pQVAE.

The paper is organized as follows: In the Methods section we present the dataset we used, the preprocessing steps employed. This section also introduces the Calo4pQVAE framework, which is detailed in three subsections: the 4p-VAE, where we give a description over the classical framework; the 4-partite RBM where we describe in detail the sampling and training procedure for conditioned and non-conditioned RBM; and in Quantum Annealer we give a brief overview of the motivations behind quantum annealers, explain how they work, and discuss how we incorporate them into our framework, including sampling methods and conditioning techniques. In the Results section we outline the training process of our model and define the criteria used to select the best-performing model. We also showcase the performance of the model using various quantitative and qualitative metrics. We interpret the results and provide a thorough discussion on the current limitations of our approach, propose next steps for improvement, and address the technical challenges ahead.

\begin{figure}
    \centering
    \includegraphics[width=0.99\linewidth]{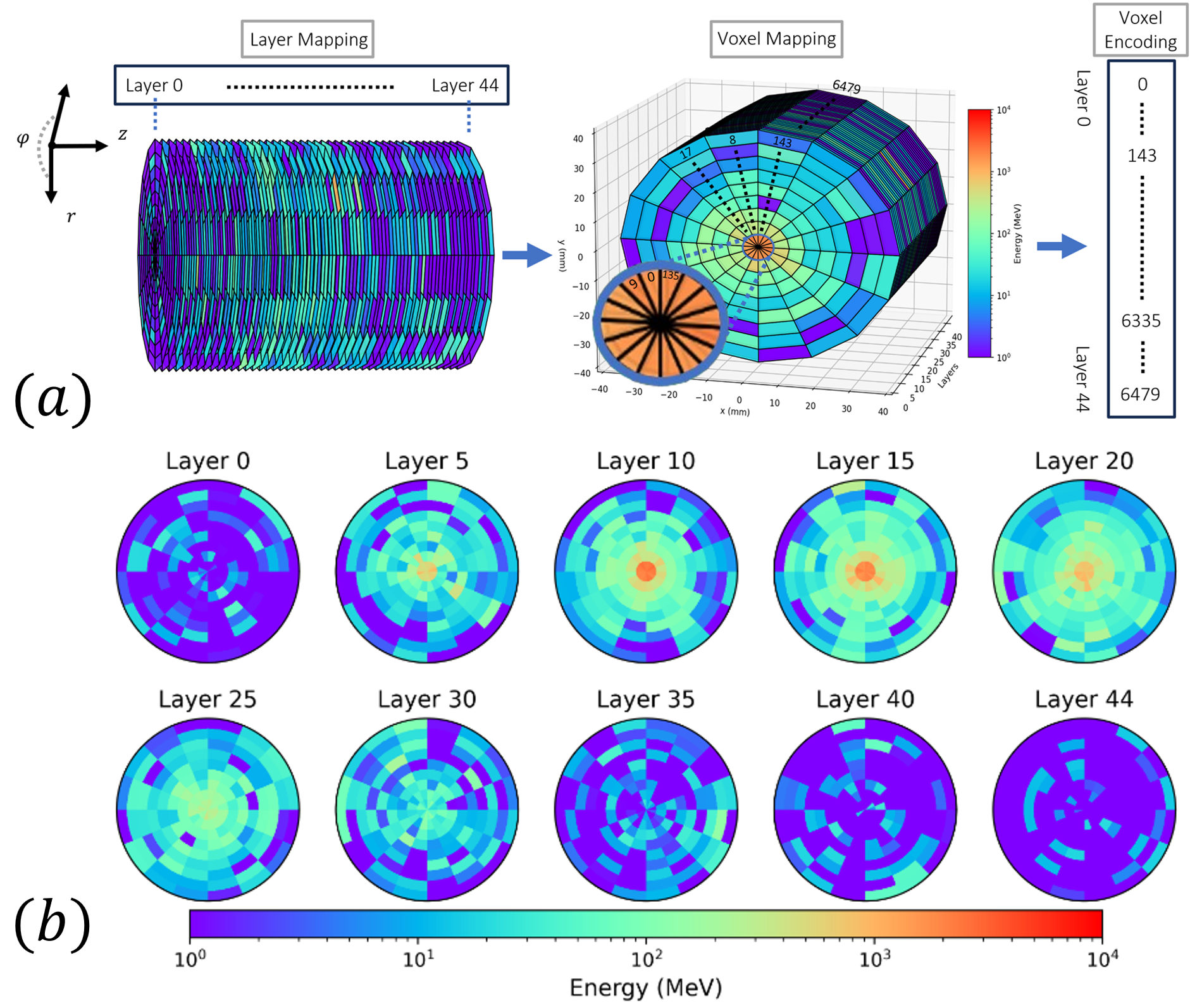}
    \caption{\textbf{(a)} Calochallenge dataset showers are voxelized using cylindrical coordinates $(r,\varphi,z)$ such that the showers evolve in the $z$ direction. For any given event, each voxel value corresponds to the energy (MeV) in that vicinity. Dataset 2 contains $100$k events and the voxelized cylinder has 45 stacked layers. Each layer has 144 voxels composed of $16$ angular bins and $9$ radial bins. The data set is parsed onto a 1D vector following the common way to eat a pizza, \textit{i.e.}, grab a slice and start from the inside towards the crust. Each 1D vector has $45\times 9 \times 16 = 6480$ voxels per each event. \textbf{(b)} Visualization of the voxels in an event in the dataset.}
    \label{fig:dataset}
\end{figure}

\section{Methods}
The CaloChallenge 2022 \cite{calochallenge} comprises three distinct datasets, each designed to facilitate research and testing in the field of calorimeter simulations. All three datasets are derived from \textsc{Geant4} simulations. The first dataset, referred to as \textit{Dataset 1}, represent photon and charged pion showers within a specified $\eta$ \footnote{Particle physics experiments typically use a right-hand coordinate system with the interaction point at the center of the detector and the $z$-axis pointing along the beam pipe. The $x$-axis points to the center of the accelerating ring, and the $y$-axis points upwards. Polar coordinates ($r$, $\phi$) are used to describe the transverse directions of the detector, with $\phi$ being the azimuthal angle around the $z$-axis. The pseudorapidity $\eta$ is defined relative to the polar angle $\theta$ as $\eta = - \ln{\tan{(\theta/2)}}$.} range. The dataset covers a discrete range of incident energies, ranging from 256 MeV to 4 TeV, logarithmically spaced out evenly with varying sizes at higher energies. The calorimeter geometry is that of the ATLAS detector.

In the case of \textit{Dataset 2} it is comprised by two files containing 100,000 \textsc{Geant4}-simulated electron showers each. These showers encompass a wide energy range, spanning from 1 GeV to 1 TeV sampled from a log-uniform distribution. The detector geometry features a concentric cylinder structure with 45 layers as shown in Fig. \ref{fig:dataset}, each consisting of active (silicon) and passive (tungsten) material. The dataset is characterized by high granularity, with $45 \times 16 \times 9 = 6,480$ voxels. The cylinder is 36 radiation lengths deep and spans nine Moli\`ere radii in diameter \cite{calochallenge}.
Lastly, \textit{Dataset 3} contains four files, each housing 50,000 \textsc{Geant4}-simulated electron showers. Similar to Dataset 2, these showers encompass energies ranging from 1 GeV to 1 TeV. The detector geometry remains consistent with Dataset 2 but exhibits significantly higher granularity, boasting 18 radial and 50 angular bins per layer, totaling $45 \times 50 \times 18 = 40,500$ voxels.

These datasets collectively offer a comprehensive resource for researchers interested in the development and evaluation of generative models and simulations within the field of calorimetry in High Energy experiments. The datasets are publicly available and accessible via \cite{d1,d2,d3}. For our results, we consider Dataset 2 and we leave the testing of our framework using the remaining datasets for future work.

\subsection{Data preprocessing}
Before feeding the shower and incident energy data to the model, we apply several transformations to the shower and the incident particle energy \textit{on-the-fly}. Given an event shower, $\bm{v}$, and corresponding incident energy, $e$, we first reduce the voxel energy, $v_i$, per event by dividing it by the incident energy, $e$, \textit{viz.} $E_i = v_i/e$. Notice that $E_i \in \left[0,1 \right]$, where the left and right bounds correspond to when the voxel energy is zero and equal to the incident energy, respectively. To remove the strict bounds, we define $u_i = \delta + (1-2\delta)E_i$, where $\delta=10^{-7}$, to prevent discontinuities during the \textit{logit} transformation. Specifically, we use the transformation $x_i=\ln u_i/(1-u_i) - \ln \delta/(1-\delta)$, where the second term preserves the zero values in the transformed variable, \textit{i.e.}, when the voxel energy is zero, $v_i = 0$, the transformed variable $x_i=0$.

The incident energy is used as a conditioning parameter and we transform it by applying a logarithmic function followed by scaling it between 0 and 1. These transformations have been used before in the same context \cite{krause2021caloflow, amram2023denoising, mikuni2024caloscore}. However, in our case, we modify the process to preserve the zeroes in the transformed variables, $x_i$, and therefore omit the last step of standardizing the new variables, in contrast with other approaches.

\subsection{4p-QVAE}
\begin{figure*}
    \centering
    \includegraphics[width=0.99\linewidth]{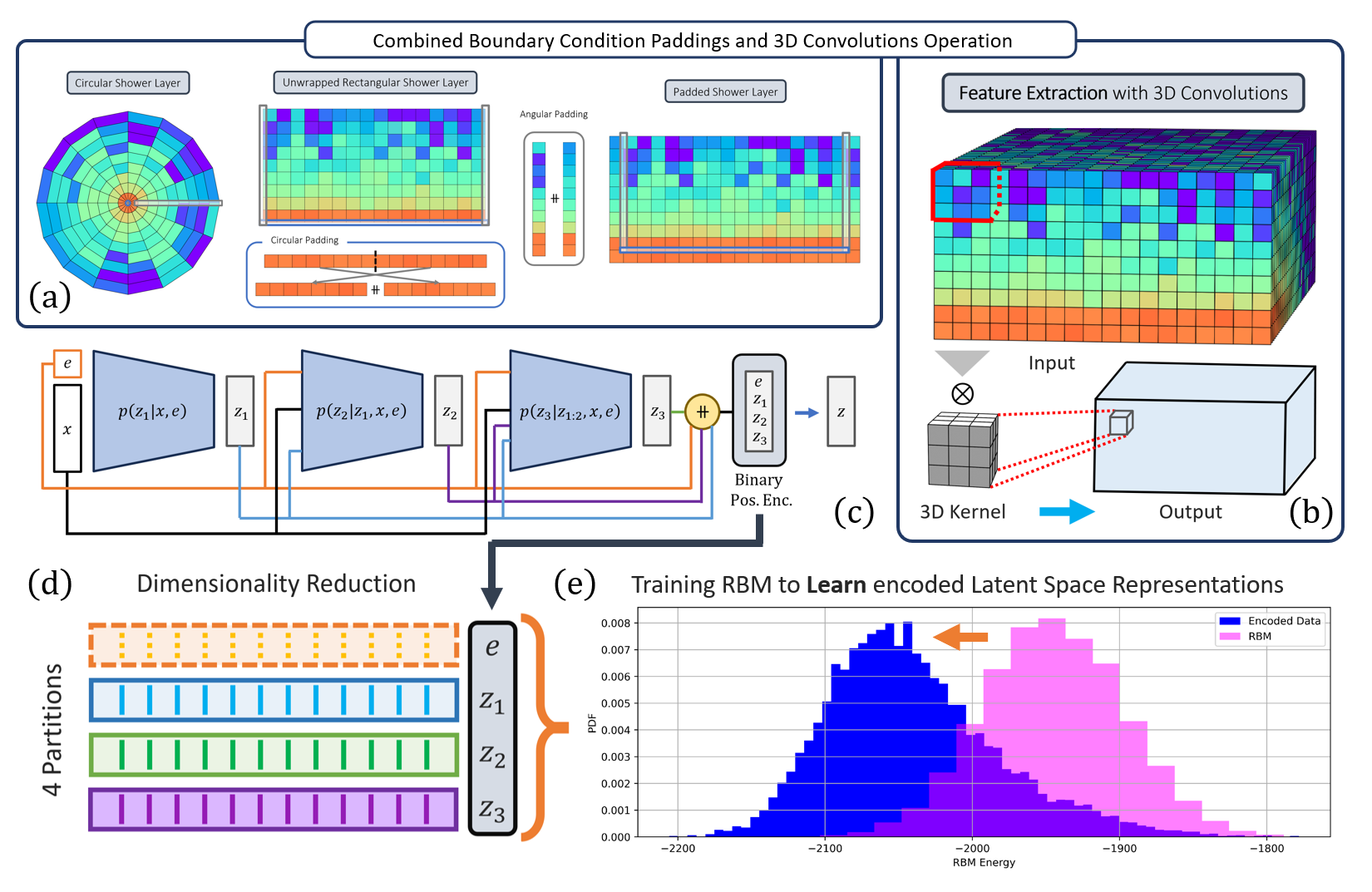}
    \caption{Diagram of the encoding framework. \textbf{(a)} We unwrap the cylindrical shower into a tensor of rank 3 with indices. We account the angular periodicity of the cylindrical geometry by padding the tensor in $theta$ dimension, such that the size becomes $45 \times 18 \times 9$. To account for the neighboring voxels in the center of the cylinder, we pad the tensor in the corresponding radial dimension. We pad it by taking the centermost voxels, splitting it in half and permuting the two halves. \textbf{(b)} These operations are performed several times, ech prior to a 3D convolution operation for feature extraction. \textbf{(c)} The encoder embeds hierarchy levels, \textit{i.e.}, the first encoder generates a fraction of the encoded data, which is then fed to the second encoder (together with the input) to generate the remaining fraction of the encoded data. The encoded data is used to train the QPU RBM. The encoded data and the incidence energy is passed to the decoder to reconstruct the energy per voxel. \textbf{(d)} The Calo4pQVAE uses a discrete binary latent space and assumes a Boltzmann distribution for prior. The energy function in the Boltzmann distribution corresponds to a sparse 4-partite graph, which allows the direct mapping to Pegasus-structured Advantage quantum annealer. }
    \label{fig:NN_diagram}
\end{figure*}

The Calo4pQVAE can be conceptualized as a variational autoencoder (VAE) with a restricted Boltzmann machine (RBM) as its prior. The modularity of our framework allows for the replacement of any component, such as the encoder, decoder, or the RBM, facilitating flexibility in its configuration \cite{winci2020path}. 
The encoder, also referred to as the \textit{approximating posterior}, is denoted as $q_\phi(\bm{z} | \bm{x}, e)$, while the latent space prior distribution is denoted as $p_\theta(\bm{z})$. The decoder, responsible for generating data from the latent variables, is represented as $p_\theta(\bm{x}|\bm{z},e)$. We train the model to generate synthetic shower events given a specific incidence energy. In other words, we are interested in finding $p_\theta(\bm{x}|\bm{z},e)$, such that $\int p_\theta(\bm{x}|\bm{z},e)p_\theta(\bm{z}) d\bm{z}$ matches the empirical dataset distribution. We use $\phi$ to denote the encoder parameters, $\theta$ for the prior and decoder parameters, and $\bm{z}$ the latent space vector. We denote as $\bm{x}$ a one-dimensional vector, such that $\bm{x} \in R^n$, and we say each element $x_i$ contains the energy measured at the $i$th voxel for that specific instance (or event), defined by the bijective transformation described in the previous subsection, while $e$ is incidence energy of the event. During training, the model takes as input an instance of $\bm{x}$ and $e$.
The input data is encoded into the latent space via the encoder. One key difference between VAEs and autoencoders is that in the latter, the same input generally yields the same encoded representation, provided the encoder does not use stochastic filters, such as dropout. In contrast, VAEs generate a different encoded representation with each pass of the same input. This is because the output of a VAE encoder consists of the parameters of a distribution from which the encoded data is sampled. The encoded data is then passed through the decoder which reconstructs the shower event vector, $\hat{\bm{x}}$. The incident particle energy is the label of the event and conditions the encoder and decoder, as depicted in Fig. \ref{fig:NN_diagram}. To condition the encoder and decoder with the incidence energy, we tested two different methods: simple concatenation of the label with the one-dimensional energy per voxel vector, and positional encoding similar to the techniques used by Meta \cite{gehring2017convolutional} and Google \cite{vaswani2017attention}. Despite experimenting with these different encoding schemes, we observed no significant difference in performance. Therefore, we opted to use the simpler concatenation method henceforth and leave the positional encoding methods for future work when dealing with a greater number of features in the dataset. 

\begin{figure}
    \centering
    \includegraphics[width=0.99\linewidth]{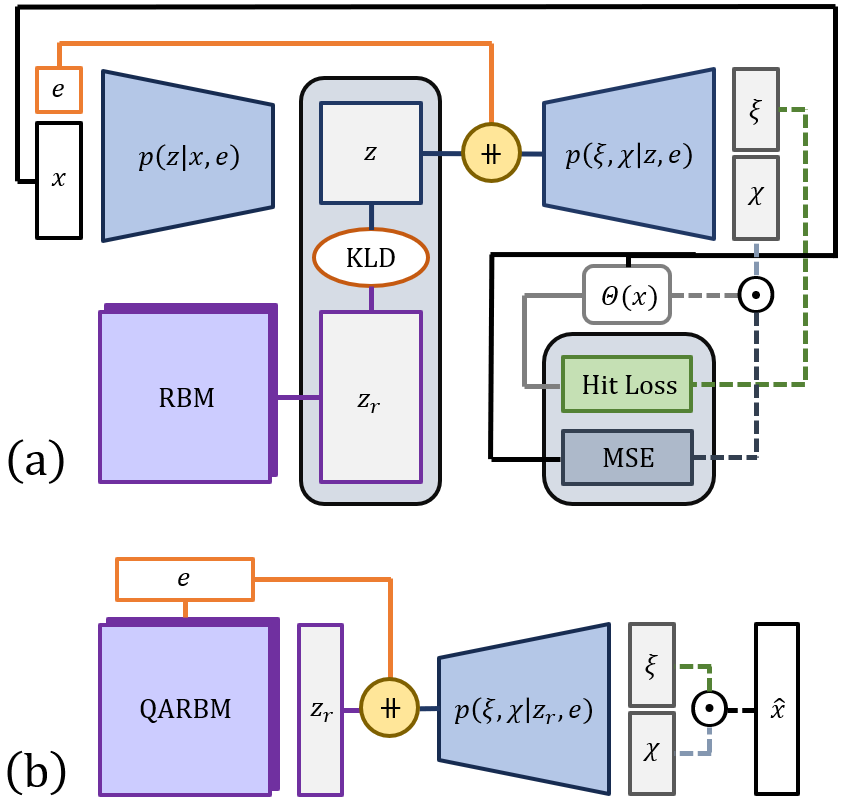}
    \caption{Sketch of Calo4pQVAE. \textbf{(a)} The input data is composed by the energy per voxel, $x$ and the incidence energy, $e$. During training, the data flows through the encoder gets encoded into a latent space, $z$, it then goes through the decoder and generates a reconstruction of the voxels per energy, while the incidence energy is the label of the event and conditions the encoder and decoder. The decoder outputs the activation vector, $\chi$ and the hits vector $\xi$. The model is trained via the optimization of the mean squared error between the input shower and the reconstructed shower, the binary cross entropy (hit loss) between the hits vector and the input shower and the Kulbach-Liebler divergence which is composed by the entropy of the encoded sample and the restricted Boltzmann machine log-likelihood. \textbf{(b)} For inference, we sample from the RBM or the QA conditioned to an incidence energy, afterwards the sample goes through the decoder to generate a shower. }
    \label{fig:arch}
\end{figure}

The encoder architecture uses hierarchy levels, as described in \cite{winci2020path}. The encoder is composed by three sub-encoders. The first generates one-fourth of the encoded data, which is then fed to the second sub-encoder along with the input data to generate another quarter of the encoded data. This process is repeated with the third sub-encoder, each time integrating all previously generated data and the original input to produce the final encoded output, as depicted in Fig. \ref{fig:NN_diagram}. The purpose of these hierarchy levels is to enforce conditional relationships among latent units by introducing conditioning among latent nodes, \textit{viz}. 
\begin{eqnarray}
    q_\phi(\bm{z}|\bm{x}) &\equiv& q_\phi(\bm{z}_1,\bm{z}_2, \bm{z}_3, \bm{z}_4|\bm{x}) \\
    &=& q_\phi^{(1)}(\bm{z}_1|\bm{x}) \prod_{\alpha=2}^4 q_\phi^{(\alpha)}(\bm{z}_\alpha|\lbrace \bm{z}_i \rbrace_{i=1}^{\alpha-1},\bm{x}) \nonumber
\end{eqnarray}
The choice of three sub-encoders is designed on mimicking the connections between the four partitions in latent space.
Our results suggest that this hierarchical approach indeed fosters correlations between latent units, leading to a Boltzmann-like distribution. Additionally, this hierarchical structure introduces multiple paths for gradient backpropagation, similar to residual networks \cite{he2016deep}, enhancing the model’s learning capability.  The fourth partition conditions the RBM and consists on a binary representation of the incident particle energy, which we describe in detail in Appendix \ref{AppSec:IncEn}.

Besides the reconstruction of the event, the encoded data is also used to train the 4-partite RBM via the Kullback-Liebler (KL) divergence. As in traditional VAEs (full derivations can be found in Ref. \cite{kingma2013auto} and in Appendix \ref{App:1}), we optimize the \textit{evidence lower bound} (ELBO), $\mathcal{L}_{\phi, \theta}(\bm{x})$, to train the Calo4pQVAE. Explicitly, the ELBO function is:
\begin{eqnarray}
     \mathcal{L}_{\phi, \theta}(\bm{x}) &=& \langle \ln p_\theta (\bm{x}|\bm{z})  \rangle_{q_\phi (\bm{z}|\bm{x})}
     - \langle \ln \frac{ q_\phi(\bm{z}|\bm{x})}{ p_\theta (\bm{z}) } \rangle_{q_\phi (\bm{z}|\bm{x})} \; . \label{eq:ELBO} \; 
\end{eqnarray}
The first term in Eq. \eqref{eq:ELBO} represents the reconstruction accuracy and the second term corresponds to the KL divergence.
To define a functional form for $p_\theta (\hat{\bm{x}}|\bm{z})$, certain assumptions about the distribution are typically made. A common assumption is that the likelihood of the reconstruction is Gaussian distributed, which simplifies to optimizing the mean squared error (MSE).  

In the case of calorimeter data, one important aspect is differentiating between voxels that are hit ($x_i \neq 0$) from those that are not hit ($x_j=0$) in a given event. Accurately training deep learning models to produce zero values can be challenging, as typical activation functions for regressions struggle to yield consistent zero outputs. Here we tackle the challenge by separating zero values from non-zero values by factorizing the data vector $\bm{x}$ into two components, \textit{i.e.}, $\bm{x} = \bm{\chi} \odot \bm{\xi}$ where $\bm{\chi} \in R^n$ represents the continuous energy values of the hits and $\xi_i = \Theta(x_i)$ is a binary vector indicating the presence of hits, with $\Theta(\bullet)$ representing the Heaviside function.

Next, we consider a joint probability $p_{\theta}(\bm{\chi},\bm{\xi}|\bm{z})$ as the probability of the voxels being hit according to $\bm{\xi}$ and energy $\bm{\chi}$. We can express the joint probability as $p_{\theta}(\bm{\chi},\bm{\xi}|\bm{z})=p_{\theta}(\bm{\chi}|\bm{\xi},\bm{z})p_{\theta}(\bm{\xi}|\bm{z})$. We model the event hitting probability, $p_{\theta}(\bm{\xi}|\bm{z})$, as a Bernoulli distribution 
$\prod_{i=1}^n p_{\xi_i}^{\xi_i}(1-p_{\xi_i})^{1-\xi_i}$.
The number of particles in the electromagnetic shower follows approximately a Poisson distribution. Furthermore, via the saddle point approximation, for large number of particles in the shower the multivariate Poisson distribution becomes a multivariate Gaussian distribution with the mean equal to the variance.
One can show that the variables $\lbrace \chi_i \rbrace_{i=1}^n $ are also approximately Gaussian distributed 
provided $v_i/e \ll 1$ (see App. \ref{AppSec:GaussApprox}).
In the present paper, we assume a variance equal to unity in our generative model, as we observed better performance for this choice. Hence, the joint distribution $p(\bm{\chi}, \bm{\xi}|\bm{z})$ can be formally expressed as
\begin{eqnarray}
    p_\theta(\bm{\chi}, \bm{\xi}|\bm{z}) &=& \prod_{i=1}^n \left( \xi_i \frac{1}{\sqrt{2\pi}}e^{-\frac{(\chi_i-x_{i})^2}{2}} \right. \nonumber \\
    &+& \left. (1-\xi_i) \delta(\chi_i) \right) p_{\xi_i}^{\xi_i}(1-p_{\xi_i})^{1-\xi_i} \; .
    \label{eq:generative_model_0}
\end{eqnarray}
Maximizing the log-likelihood of $p_\theta(\bm{\chi}, \bm{\xi}|\bm{z})$ is rather difficult due to the factor containing the Dirac delta function. It may well be possible to replace the Dirac delta function with a smooth parameterized function. Yet, there is a simpler way where we instead mask $\bm{\chi}$ with $\bm{\xi}$, in which case we can neglect the term containing the Dirac delta function. The key point to stress here is that by means of the mask we split the tasks between generating zeroes and non-zeroes, and generating the voxel energy via $\bm{\xi}$ and $\bm{\chi}$, respectively. Therefore, the joint distribution in Eq. \eqref{eq:generative_model_0} becomes \footnote{We removed the $\xi$ coefficient due to normalization.}:
\begin{eqnarray}
    p_\theta(\bm{\chi}, \bm{\xi}|\bm{z}) &=& \prod_{i=1}^n \frac{1}{\sqrt{2\pi}}e^{-\frac{(\chi_i \cdot \xi_i-x_{i})^2}{2}} \nonumber \\
    && p_{\xi_i}^{\xi_i}(1-p_{\xi_i})^{1-\xi_i} \; . \label{eq:generative_model}
\end{eqnarray}
The reconstruction term transforms to
\begin{eqnarray}
    \langle \ln p_\theta (\bm{\chi}, \bm{\xi}|\bm{z})  \rangle_{q_\phi (\bm{z}|\bm{x})} = \langle \sum_{i=1}^n \left[ - \left( \chi_i \cdot \xi_i - x_i \right)^2 \right.  \nonumber \\
    \left. + \xi_i \ln(p_{\xi_i}) + (1-\xi_i)\ln(1-p_{\xi_i}) + \text{const} \right] \rangle_{q_\phi (\bm{z}|\bm{x})} \; . \label{eq:recon}
\end{eqnarray} 

While Eq. \eqref{eq:generative_model} outlines the functional structure of the reconstruction of our generative model, the formulation of loss function requires further consideration. Specifically, there are three key aspects to address. First, the Bernoulli-distributed hits vector is designed to capture the hits distribution in the dataset, yet binary entropy alone does not adequately represent this. Next, the MSE term in Eq. \eqref{eq:recon} is biased towards low energy when the hits vector reconstruction, $\bm{\xi}$, incorrectly predicts values during training. To tackle both concerns, we replace $\xi \rightarrow \Theta(x_i)$, which converts the binary entropy into a binary cross entropy (BCE). The third consideration involves how BCE penalizes errors. In its standard form, BCE penalizes all incorrect hit predictions equally. However, it is intuitive that this should not be the case. For instance, predicting a voxel hit of 0 when the true value is 100 GeV should be penalized more heavily than predicting 0 when the true value is 1 GeV. To account for this, we experimented with reweighting the BCE terms proportional to the voxel value. Despite this modification, we did not observe a significant improvement. 

Maximizing the ELBO function leads to maximizing the log-likelihood of $p_\theta(x)$. In practice, VAEs are trained by minimizing the negative ELBO function, from which it is straightforward to notice that the first term in the r.h.s. of Eq. \eqref{eq:recon} becomes mean squared error (MSE), while the second and third terms correspond to the binary cross entropy. After taking these considerations into account, flipping the sign in the ELBO function and neglecting constant terms, Eq. \eqref{eq:recon} becomes:
\begin{eqnarray}
    - \langle \ln p_\theta (\bm{\chi}, \bm{\xi}|\bm{z})  \rangle_{q_\phi (\bm{z}|\bm{x})} = \langle \sum_{i=1}^n  \left( \chi_i(\theta) \cdot \Theta(x_i) - x_i \right)^2 \rangle_{q_\phi (\bm{z}|\bm{x})}   \nonumber \\
    -  \langle \Theta(x_i) \ln(p_{\xi_i}(\theta)) + (1-\Theta(x_i))\ln(1-p_{\xi_i}(\theta)) \rangle_{q_\phi (\bm{z}|\bm{x})} \; . \label{eq:recon2}
\end{eqnarray}
In Eq. \eqref{eq:recon2}, we use the \textit{Gumbel trick} for the hits vector. We explain the approach in what follows.

The second term in the r.h.s. of Eq. \eqref{eq:ELBO} is known as the VAE regularizer. This term can be parsed as:
\begin{equation}
    \langle \ln \frac{ q_\phi(\bm{z}|\bm{x})}{ p_\theta (\bm{z}) } \rangle_{q_\phi (\bm{z}|\bm{x})} = \langle \ln q_\phi(\bm{z}|\bm{x}) \rangle_{q_\phi (\bm{z}|\bm{x})} - \langle \ln p_\theta (\bm{z}) \rangle_{q_\phi (\bm{z}|\bm{x})} \label{eq:regularizer}
\end{equation}
Let us focus on the first term on the r.h.s. of Eq. \eqref{eq:regularizer}. When taking the gradient of the entropy with respect to the model parameters one faces two issues: \textit{i)} due to the discrete nature of the latent variables $z$, the gradient becomes singular; and \textit{ii)} given a function evaluated on a random variable, it is ill-defined taking the gradient of the function with respect to the parameters of the random variable's probability density function. Moreover, taking the gradient of a discrete estimator, $\sum_{\bm{z} \sim q_\phi(\bm{z}|\bm{x})}(...)$, with respect to $\phi$ is not well-defined, since the gradient variables appear in the summation argument. The first issue can be easily addressed by simply considering a continuous step-like function, such as a sigmoid, $\sigma(\bullet)$. Specifically, we replace $\bm{z}$ with $ \bm{\zeta}$, where $\zeta_i = \sigma(\left[\text{encoded data point}\right]_i \cdot \beta)$, and $\beta$ is an annealing parameter, such that $\lim_{\beta\rightarrow \infty} \bm{\zeta} = \bm{z}$. The second issue can be addressed by means of the so-called \textit{Gumbel trick} \cite{maddison2016concrete}. The Gumbel trick has become an umbrella term which refers to a set of methods to sample from discrete probabilities or to estimate its partition function. In our case, we simply generate latent variables $\bm{\zeta}$ via
\begin{equation}
    \zeta_i = \sigma( (l_i(\phi, x) + \sigma^{-1}(\rho_i)) \beta) \quad \text{for all } i=1,...,m\; , \label{eq:GumbelTrick}
\end{equation}
where $\lbrace \rho_i \rbrace_{i=1}^m$ is a set of i.i.d. uniform random numbers, and $\lbrace l_i(\phi,\bm{x}) \rbrace_{i=1}^m$ is a set of logits, \textit{i.e.}, a set of unbounded real numbers generated from the encoder. The connection with Gumbel distributed random numbers is due to the fact that $\sigma^{-1}(\rho) \sim G_1 - G_2$, where $G_1$ and $G_2$ are two Gumbel distributed random numbers. Moreover, under the recipe given in Eq. \eqref{eq:GumbelTrick} one is guaranteed that in the discrete regime of $\bm{\zeta}$ (\textit{i.e.}, $\beta \rightarrow \infty$), $P(\zeta_i = 1) = \sigma(l_i(\phi,x))$ \cite{maddison2016concrete, balog2017lost, khoshaman2018gumbolt}. Taking into account the previous, we can then express the entropy as
\begin{eqnarray}
    \langle \ln q_\phi(\bm{\zeta}|\bm{x}) \rangle_{q_\phi (\bm{z}|\bm{x})} = \langle \sum_{i=1}^m \zeta(l_i,\beta, \rho_i) \ln \sigma(l_i) \nonumber \\
    + (1-\zeta(l_i,\beta, \rho_i)) \ln (1-\sigma(l_i)) \rangle_{q_\phi (\bm{z}|\bm{x})}  \label{eq:EntropyEstimator}
\end{eqnarray}

The second term in Eq. \eqref{eq:regularizer} can be expanded as follows. By design the functional form of $p_\theta(\bm{z})$ is that of a Boltzmann distribution in a heat bath at temperature $T=1$ ($k_B=1$). Therefore, we have:
\begin{equation}
    \langle \ln p_\theta (\bm{z}) \rangle_{q_\phi (\bm{z}|\bm{x})} = - \langle E(\bm{z}) \rangle_{q_\phi (\bm{z}|\bm{x})} -  \ln Z  \; . \label{eq:BoltzmannDistLL}
\end{equation}
The first term in the r.h.s. is the energy function averaged over the approximate posterior, whereas the second term is the free energy, which is independent of the encoded samples. 
The free energy can be replaced by the internal energy, $U$, minus the entropy, $S$, times the temperature, \textit{viz.} $U-TS$. We use the \textit{high temperature} gradient approximation to replace the second term in the r.h.s. of Eq. \eqref{eq:BoltzmannDistLL} with the internal energy, which is a common practice in approximating the free energy when training RBMs. In the high temperature gradient approximation, as the energy in the system saturates, the specific heat converges to zero. We demostrate in Appendix \ref{sec:HTGA} that when $\langle E(\bm{z}) \rangle \langle \partial E(\bm{z})/\partial \phi  \rangle = \langle E(\bm{z}) \partial E(\bm{z})/\partial \phi \rangle$, the specific heat is zero, which allows us to rewrite the previous Equation as
\begin{equation}
    \langle \ln p_\theta (z) \rangle_{q_\phi (\bm{z}|\bm{x})} = - \langle E(\bm{z}) \rangle_{q_\phi (\bm{z}|\bm{x})} + U \; . \label{eq:BoltzmannDistLLSimp}
\end{equation}
In this regime, the entropy is solely configurational, scales linearly with the number of units in the RBM and is independent of the energy parameters. 

Computing $U$ in Eq. \eqref{eq:BoltzmannDistLLSimp} requires calculating the partition function which becomes intractable beyond a few tens of nodes due to the exponential scaling of the number of states with the number of nodes. Instead, we estimate $U$ using uncorrelated samples. To obtain these uncorrelated samples, one performs Markov Chain Monte Carlo simulations also known as block Gibbs sampling. In the following section we elaborate on how we sample from the 4-partite RBM.

\subsection{4-partite Restricted Boltzmann Machine} \label{sec:4pRBM}
Let us consider a 4-partite restricted Boltzmann machine. For this purpose, we denote each of the four layers as $v,h,s$ and $t$. The energy function is defined by:
\begin{eqnarray}
    E(\mathbf{v},\mathbf{h},\mathbf{s},\mathbf{t}) = - a_i v_i - b_i h_i - c_i s_i - d_i t_i \nonumber \\
    - v_i W^{(0,1)}_{ij} h_j - v_i W^{(0,2)}_{ij} s_j \nonumber \\
    - v_i W^{(0,3)}_{ij} t_j - h_i W^{(1,2)}_{ij} s_j \nonumber \\
    - h_i W^{(1,3)}_{ij} t_j - s_i W^{(2,3)}_{ij} t_j \; ,
    \label{eq:RBM_energy}
\end{eqnarray}
where we are using the double indices convention for summation.

The Boltzmann distribution has the following form:
\begin{equation}
    p(\mathbf{v},\mathbf{h},\mathbf{s},\mathbf{t}) = \frac{\exp(- E(\mathbf{v},\mathbf{h},\mathbf{s},\mathbf{t}))}{Z} \; .
\end{equation}
where $Z$ is the partition function,
\begin{equation}
    Z = \sum_{\lbrace \mathbf{v},\mathbf{h},\mathbf{s},\mathbf{t} \rbrace} \exp(- E(\mathbf{v},\mathbf{h},\mathbf{s},\mathbf{t})) \; .
\end{equation}
Similar to the 2-partite RBM, we can express the distribution over any one of the layers conditioned to the remaining three in terms of the ratio of the Boltzmann distribution and the marginalized distribution over the layer of interest. Without any loss in generality, let us assume the layer of interest is $h$, then the probability distribution on $p(\mathbf{h} | \mathbf{v}, \mathbf{s}, \mathbf{t})$ is given by:
\begin{eqnarray}
    p(\mathbf{h} | \mathbf{v}, \mathbf{s}, \mathbf{t}) &=& \frac{p(\mathbf{v}, \mathbf{h}, \mathbf{s}, \mathbf{t})}{p(\mathbf{v}, \mathbf{s}, \mathbf{t})} \nonumber \\
    &=& \frac{\exp(- E(\mathbf{v},\mathbf{h},\mathbf{s},\mathbf{t}))}{\sum_{\mathbf{h}} \exp(- E(\mathbf{v},\mathbf{h},\mathbf{s},\mathbf{t}))} \nonumber \\
    &=& \prod_i \frac{\exp(h_i \cdot (b_i +B_i))}{1+\exp(b_i +B_i)} \label{eq:p_H}
\end{eqnarray}
with
\begin{equation}
    B_i =  W^{(01)}_{ji}v_j + W^{(12)}_{ij}s_j + W^{(13)}_{ij}t_j
\end{equation}
We can rewrite Eq. \eqref{eq:p_H} in a rather compact form. Furthermore, we do the same for the case where we condition any of the four partitions on the remaining three. We obtain: 
\begin{subequations}
\begin{align}
    p(\mathbf{v}=\mathbf{1} | \mathbf{h}, \mathbf{s}, \mathbf{t}) = \prod_i \sigma(a_i +A_i) \label{eq:BGS_not_thisone} \; , \\
    p(\mathbf{h}=\mathbf{1} | \mathbf{v}, \mathbf{x}, \mathbf{y}) = \prod_i \sigma(b_i +B_i) \; , \\
    p(\mathbf{s}=\mathbf{1} | \mathbf{v}, \mathbf{h}, \mathbf{t}) = \prod_i \sigma(c_i +C_i) \; , \\
    p(\mathbf{t}=\mathbf{1} | \mathbf{v}, \mathbf{h}, \mathbf{s}) = \prod_i \sigma(d_i +D_i) \; .
\end{align}
\label{eq:BGS}
\end{subequations}
where
\begin{subequations}
\begin{align}
    A_i = W^{(01)}_{ij}h_j + W^{(02)}_{ij}s_j + W^{(03)}_{ij}t_j \; , \\
    B_i =  W^{(01)}_{ji}v_j + W^{(12)}_{ij}s_j + W^{(13)}_{ij}t_j \; ,\\
    C_i = W^{(02)}_{ji}v_j + W^{(12)}_{ji}h_j + W^{(23)}_{ij}t_j \; ,\\
    D_i = W^{(03)}_{ji}v_j + W^{(13)}_{ji}h_j + W^{(23)}_{ji}s_j\; .
\end{align}
\end{subequations}
We then propose a Gibbs sampling process to sample from the joint distribution $p(\mathbf{v}, \mathbf{h}, \mathbf{s}, \mathbf{t})$ of the 4-partite RBM. Notice that the joint distribution can be factorized as $p(\mathbf{v}, \mathbf{h}, \mathbf{s}, \mathbf{t}) = p(\mathbf{v} | \mathbf{h}, \mathbf{s}, \mathbf{t}) p(\mathbf{h}, \mathbf{s}, \mathbf{t})$. Extending the approach used in 2-partite RBM, we perform Markov chain Monte Carlo by sequentially updating each partition conditioned on the remaining, \textit{i.e.}, the prior in step $n+1$ is deemed the posterior in step $n$.
The Gibbs sampling process, at iteration $n$, is done in the following steps:
\begin{enumerate}
    \item Sample partition $\bm{v}$: $p(\bm{v} | \bm{h}^{(n)}, \bm{s}^{(n)}, \bm{t}^{(n)})$
    \item Sample partition $\bm{h}$: $p(\bm{h} | \bm{v}^{(n+1)}, \bm{s}^{(n)}, \bm{t}^{(n)})$
    \item Sample partition $\bm{s}$: $p(\bm{s} | \bm{v}^{(n+1)}, \bm{h}^{(n+1)}, \bm{t}^{(n)})$
    \item Sample partition $\bm{t}$: $p(\bm{t} | \bm{v}^{(n+1)}, \bm{h}^{(n+1)}, \bm{s}^{(n+1)})$
\end{enumerate}
Notice that one block Gibbs sampling step corresponds to four sampling steps. By repeating this process iteratively, we generate samples that approximate the joint distribution.

We have shown how to generalize an RBM to a four-partite graph. It is important to emphasize that, as with any deep generative model, the features of the dataset are expressed in latent space. In this context, each state in latent space encodes specific features of the dataset. Different techniques such as t-SNE \cite{van2008visualizing} and LSD \cite{toledo2023using}, can illustrate how dataset features are embedded in latent space after training. These methods allow for the conditioning of the \textit{prior} on specific features. There is a rather straightforward way to condition the \textit{prior} on given dataset parameters during training, which is widely used across different frameworks. Generally, this involves using specific channels to condition the prior and specifics depend on the actual framework. An accurate and robust method to sample from latent space constrained to specific features is crucial for practical applications in generative deep models. In the present context, a calorimeter surrogate requires input related to particle type, incident particle energy and incident angle, among other parameters. In the following, we show how to condition the 4-partite RBM sampling process.



\subsubsection{Conditioned 4-partite Restricted Boltzmann Machine}
In this section we outline the approach to condition the 4-partite RBM. In this context, the conditioned RBM refers to the scenario where a subset of nodes in the RBM is kept fixed during the block Gibbs sampling process. We adapt the methodology from \cite{mnih2012conditional} to our Calo4pQVAE. A crucial element of our approach is the use of one complete partition to deterministically encode the data features. Although utilizing the entire partition for feature encoding might appear excessive, it is necessary due to the sparsity of the quantum annealer. Exploring the optimal number of nodes required to efficiently encode the features is beyond the scope of this paper. Moreover, as we move forward towards more complex datasets, we anticipate an increase in the number of features that need to be encoded, underscoring the importance of our chosen approach as a baseline.

Notice that by treating one partition as a conditioning parameter, we effectively modify the self-biases of the remaining partitions. To fix ideas, we consider partition $v$ as the conditioning partition, therefore we may rewrite Eq. \eqref{eq:RBM_energy} as
\begin{eqnarray}
    E(\mathbf{h},\mathbf{s},\mathbf{t}| \mathbf{v}) = - (b_i +  v_j W^{(0,1)}_{ji}  )h_i - (c_i + v_j W^{(0,2)}_{ji} )s_i \nonumber \\  
    - (d_i + v_j W^{(0,3)}_{ji}) t_i - h_i W^{(1,2)}_{ij} s_j \nonumber \\
    - h_i W^{(1,3)}_{ij} t_j - s_i W^{(2,3)}_{ij} t_j \; .
\end{eqnarray}
The block Gibbs sampling procedure is the same as before via Eqs. \eqref{eq:BGS} although we skip Eq. \eqref{eq:BGS_not_thisone}. The authors in \cite{mnih2012conditional} caution on the use of contrastive divergence when training a conditioned RBM due to the possibility of vanishing gradient even when the conditioned RBM might not be fully trained. This happens when mixing times are larger than the number of Gibbs sampling steps. Despite the previous, we use contrastive divergence in our framework and we show how the estimated log-likelihood saturates after $250$ epochs. We further elaborate and discuss this approach in the discussion section.

\subsection{Quantum Annealers}
A quantum annealer (QA) is an array of superconducting flux quantum bits with programmable spin–spin couplings and biases with an annealing parameter \cite{johnson2011quantum}. The motivation for QAs comes from the adiabatic approximation \cite{sakurai2017jim}, which asserts that if a quantum system is in an eigenstate of its Hamiltonian (which describes the total energy of the system), and the Hamiltonian changes slowly enough, then the system will remain in an eigenstate of the Hamiltonian, although the state itself may change (see Appendix \ref{App:QA} for a formal derivation). QAs were initially thought of as a faster method to find the ground state of complex problems that could be mapped onto a Hamiltonian $H$ \cite{ronnow2014defining}. This can be done by initializing the system in the ground state of some Hamiltonian $H_0$, which is easy to prepare both theoretically and experimentally. In addition, by design the commutator $[H,H_0]\neq 0$. The QA interpolates between the two Hamiltonians via 
\begin{eqnarray}
    H_{QA} = \frac{\mathcal{A}(s)}{2} H_0 + \frac{\mathcal{B}(s)}{2} H \label{eq:DwaveHam}
\end{eqnarray}
with
\begin{equation}
    \begin{cases}
        H_0 = - \sum_{i} \hat{\sigma}_x^{(i)} \\
        H =  \sum_{i} \Delta_i \hat{\sigma}_z^{(i)} + \sum_{i>j} J_{ij} \hat{\sigma}_z^{(i)} \hat{\sigma}_z^{(j)} \label{eq:QA}
    \end{cases}
\end{equation}
such that the annealing parameters, $\mathcal{A}(s)$ and $\mathcal{B}(s)$, are two slowly-varying controllable parameters constrained to $\mathcal{A}(0) \gg \mathcal{B}(0)$ and $\mathcal{A}(1) \ll \mathcal{B}(1)$ and $s\in [0,1]$ \cite{hauke2020perspectives}. In practice, quantum annealers have a strong interaction with the environment which lead to thermalization and decoherence. Evidence suggests that the culprit are the $\sigma_z$ operator which couple to the environment \cite{benedetti2016estimation}. There has been efforts towards mitigating decoherence via \textit{zero noise extrapolation} methods \cite{amin2023quantum}. 

Thermalization and decoherence are usually unwanted features in quantum systems as it destroys the quantum state. In our case, these features allows us to replace the RBM with the QA. In other words, RBMs are classical simulations of QAs. A non-desired feature in our framework correspond to system arrest or freeze-out during annealing \cite{amin2015searching}, akin to glass melts subject to a rapid quench \cite{debenedetti2001supercooled}. Similar to glasses, the annealing time and protocol can have a dramatic impact on the end state \cite{marshall2019power}. It has been shown that the distribution in this freeze-out state can be approximated with a Boltzmann distribution \cite{winci2020path}.

\begin{figure}
    \centering
    \includegraphics[width=1.0\linewidth]{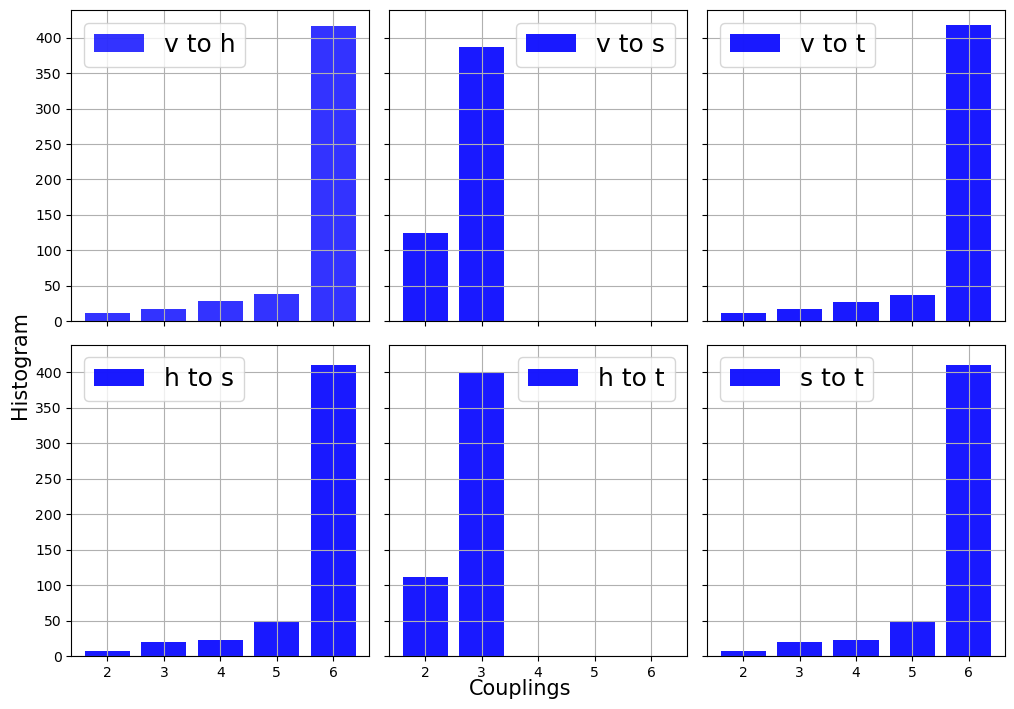}
    \caption{Quadripartite RBM weight matrices. Each panel correspond to the histogram of connections between partition A and partition B (see legend).}
    \label{fig:peg_wMat}
\end{figure}

In our pipeline we use Dwave's \textit{Advantage\_system6.4} \cite{DWaveSystems3} which is composed by $5627$ qubits and are coupled such that it forms a quadri-partite graph. Typically each qubit is coupled with 16 other qubits. In Fig. \ref{fig:peg_wMat} we show the histogram for number of connections between each of the four partitions. 

Notice that the RBM data are binary vectors such that each element can take value of $0$ or $1$, whereas the qubits can have values $-1$ or $1$. Hence, to map the RBM onto the QA one redefines the RBM variables $x_i \rightarrow (\sigma_z^{(i)}+1)/2$ and rearrange the terms in the RBM Hamiltonian to be mapped onto the QA (this is explicitly shown in Appendix \ref{App:QA}). The previous variable change will lead to an energy offset between the RBM and the QA, which henceforth we neglect. It is important to stress that the QA is in a heat bath with temperature $T_{QA} \lesssim 15 mK$ \cite{DWaveSystems2}, while the annealing parameter has an upper bound $\mathcal{B}(1) \approx 5.0\cdot 10^{-24} J$ \cite{DWaveSystems}, which implies that the prefactor $\beta_{QA} = \frac{\mathcal{B}(1)}{2k_{B}T_{QA}} \approx 12$ while in the RBM by design $\beta_{RBM}=1$. To ensure that both, the RBM and the QA describe the same Boltzmann distribution with the same temperature, one can either re-scale the QA Hamiltonian by $1/\beta_{QA}$ or the RBM Hamiltonian by $\beta_{QA}$. Currently, there is not a way to measure the QA prefactor, however there are different ways to estimate it \cite{benedetti2016estimation, raymond2016global}. In addition, there is not a direct way to control this prefactor. The most common way to estimate this prefactor is via the Kullback-Liebler divergence between the QA and the RBM distributions, whereby one introduces a tuning parameter $\beta$ as a prefactor in the RBM distribution which is tuned to minimize the Kullback-Liebler divergence. It is in the process of estimating the $\beta_{QA}$ that we also indirectly control the parameter by rescaling the Hamiltonian. The previous can be mathematically expressed as

\begin{figure}
    \centering
    \includegraphics[width=0.9\linewidth]{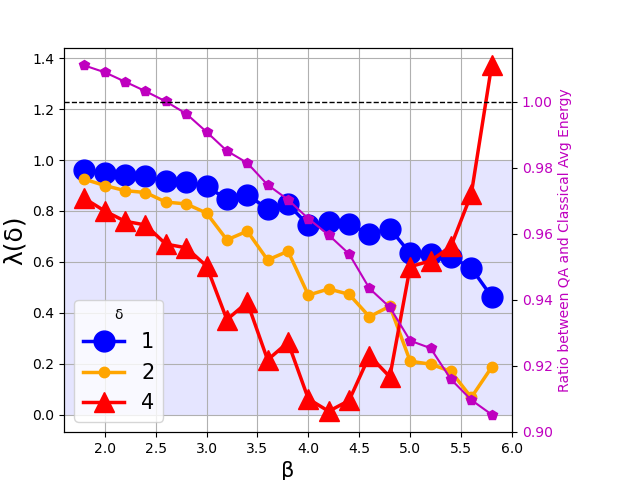}
    \caption{$\lambda(\delta)$ (see Eq. \eqref{eq:stability_maintext}) \textit{vs} $\beta$ for different $\delta$ values (see legend). The stability region is shaded in light blue. Different $\delta$ values affect the stability depending on the $\beta$ values. For instance,  for low $\beta$ values, large $\delta$ parameters leads to better stability; conversely, large $\beta$ values with large $\delta$ parameter leads to instabilities. The purple pentagons correspond to the average energy ratio between QPU samples and classical samples. The intersection between the black dashed line and the purple curve define the fixed point.}
    \label{fig:stability_plot}
\end{figure}

\begin{equation}
    \beta_{t+1} = \beta_t - \eta (\langle H \rangle_{QA} - \langle H \rangle_{RBM}) \; . \label{eq:beta}
\end{equation}
The previous converges when the average energy from the RBM matches that from the QA, in which case, $\beta = \beta_{QA}$. 
Due to coupling-dependent temperature fluctuations, $\beta_{QA}$ is expected to change in the process of training the model. For this reason the method described by Eq. \eqref{eq:beta} should be employed, in principle, after each parameter update. Notice that optimizing $\beta$ in Eq. \eqref{eq:beta} requires samples from the RBM, furthermore, it is the classical RBM temperature which is being tuned to match the distribution of the QA. In our case, we want the opposite, 
to fit the QA distribution to that of the RBM and not the other way around. To address this point, $H(x)$ is replaced with an annealed Hamiltonian $H(x,\beta)=H(x)/\beta$, \textit{i.e.}, instead of tuning the temperature in the classical RBM, to match the distribution in the QA, one iteratively rescales the Hamiltonian in the QA to effectively tune the QA's temperature to match that of the classical RBM.  This approach, as already mentioned, is well-known \cite{benedetti2016estimation, raymond2016global, winci2020path} and has been proven to be empirically robust yet slow to converge.
We therefore propose a different protocol which empirically converges faster than the KL divergence:
\begin{equation}
    \beta_{t+1} = \beta_t \left( \frac{\langle H \rangle_{QA}}{\langle H \rangle_{RBM}} \right)^{\delta} \; . \label{eq:HaoMethod}
\end{equation}
Notice that the previous map has a fixed point at $\beta_t = \beta_{QA}$. The condition for a stable fixed point is $\lambda(\delta) < 1$, where
\begin{eqnarray}
    \lambda(\delta) = 
    \begin{cases}
        |1+ \frac{\sigma^2_{QA}}{\langle H \rangle_{B(1)}}|, \; \delta=1 \\
        |1+ \delta \frac{\sigma^2_{QA}}{\langle H \rangle_{QA}}|, \; \delta \neq 1 \; .
    \end{cases} \label{eq:stability_maintext}
\end{eqnarray}
In Fig. \ref{fig:stability_plot} we show Eq. \eqref{eq:stability_maintext} \textit{vs} $\beta$ for different values of $\delta$. The values of $\beta$ chosen for this plot correspond to where we typically find the fixed point. We call $\delta$ a stability parameter since we can tune it to stabilize the mapping per iteration and modulate the fixed point attractor feature to ultimately converge in a smaller number of iterations, as we show in the next section. The purple markers in Fig. \ref{fig:stability_plot} correspond to the ratio of the average RBM energy obtained from the QA and that obtained from classical sampling. When the ratio equals one, the mapping in Eq. \eqref{eq:HaoMethod} is at the fixed point.
In Appendix \ref{App:QA} we provide a fully detailed derivation and in the next section we compare both methods.

Up to this point, we have detailed the process of replacing the RBM with the QA in the Calo4pQVAE. As previously mentioned, we utilize a conditioned RBM to enable the sampling of showers with specific characteristics. In the next section, we will explore various approaches for conditioning the QA, aiming to achieve targeted sampling of particle showers with desired features.


\subsubsection{Conditioned Quantum Annealer}
Quantum annealers were designed to find solutions to optimization problems by identifying a Boolean vector that satisfies a given set of constraints. Typically, QAs are not intended to be conditioned or manipulated in terms of fixing specific qubits during the annealing process. However, there are different approaches that can be employed to fix a set of qubits during the annealing process. In this context, we will present two such approaches. For clarity, let us revisit the concept of conditioning in the realm of QAs. Our goal is to utilize QAs in a manner that allows us to fix a subset of qubits, denoted as $\sigma_z^{(k)}$ (see Eq. \eqref{eq:QA}), \textit{a priori}, such that these qubits remain in their predetermined states throughout and after the annealing process.

\textbf{Reverse annealing with zero transverse field for conditioning qubits.} This approach requires control over the biases in $H_0$ from Eq. \eqref{eq:QA}, where $H_0 = \sum_i \kappa_i \hat{\sigma}_{x}^{(i)}$ and $\lbrace \kappa_i \rbrace$ are directly specified by the user. Initially, we set the qubits encoding the condition $\sigma_z^{(k)}$ while the rest of the qubits are randomly initialized. We also set the biases $\kappa_k = 0$ to ensure that the transverse field does not alter the state of $\sigma_z^{(k)}$. Subsequently, we perform reverse annealing, by starting from $s=1$ and reversing the annealing process towards $s=0$, before completing the annealing process as usual. The primary drawbacks in this approach are: 
\begin{itemize}
    \item Speedup compromise: The annealing process is effectively doubled in duration due to the reverse annealing step.
    \item Condition destructed by thermal fluctuations: There is a possibility of the conditioned state being altered due to thermal fluctuations.
\end{itemize} 
Both drawbacks can be mitigated by decreasing the annealing time, as this would not only reduce the overall duration but also minimize the destruction of the conditioned-encoding state by thermal fluctuations. However, as previously noted, reducing the annealing time can lead to the issue of dynamical arrest, where the system becomes trapped in a local minimum \cite{amin2015searching}. Given that our current framework relies on thermodynamic fluctuations, we leave this approach for future exploration and proceed to discuss an alternative method that does not suffer from these drawbacks and is more practical for immediate implementation.

\textbf{Conditioning qubits through flux biases.} This approach relies on using the external flux bias, $\Phi_k^x$, as effective biases to fix the qubits during annealing. The external flux bias were introduced as a practical remedy to render the biases $h_i$ time-independent during the annealing by having $B(s) h_i = 2 \Phi_i^x I_p(s)$, where $I_p(s)$ denotes the magnitude of the supercurrent flowing about the rf-SQUID loop. This flux bias is applied to the qubit loop about which the supercurrent flows \cite{harris2010experimental}. Flux biases work as effective biases on qubits. Hence, before the annealing we specify the encoded incidence energy via the flux biases. Through out our experiments, we always made sure that the partition encoding the incidence energy in the QA after annealing matched the encoded incidence energy before the annealing.

\begin{figure}
    \centering
    \includegraphics[width=0.9\linewidth]{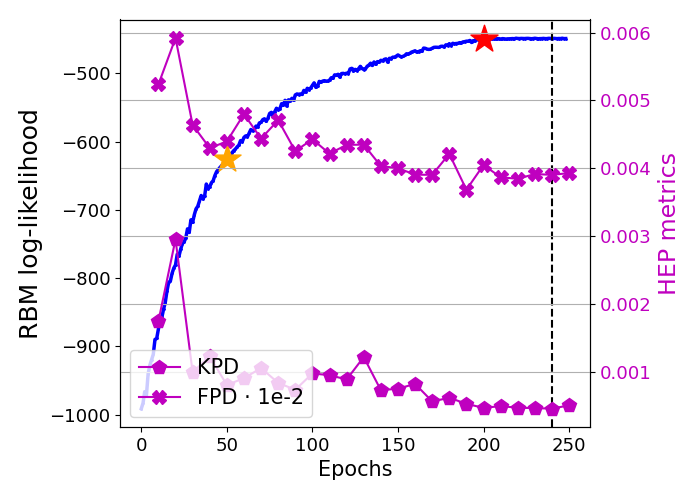}
    \caption{Restricted Boltzmann Machine log-likelihood \textit{vs} epochs. \textit{Annealed importance sampling} and \textit{reverse annealed importance sampling} methods were used to estimate the partition function. The annealed step  was set to $1/30$. The yellow star marks the epoch were the annealed training parameters have reached their final values. The red star marks the point after which the encoder and decoder training parameters are frozen. In magenta, we plot the KPD and FPD (right axis) \textit{vs} epochs.}
    \label{fig:RBM-LL}
\end{figure}

\section{Results}
In this section, we present the results concerning training and evaluating our model on Dataset 2. Therefore, we present as results the training aspects of our framework.

\begin{table}[!t]
\renewcommand{\arraystretch}{1.3}
\caption{Fr\'echet Particle Distance (FPD) and Kernel Particle Distance (KPD) metrics, implemented in the JetNet \cite{jetlib} library and adapted for the CaloChallenge \cite{calochallenge}.}
\label{table_metrics}
\centering
\begin{tabular}{c c c }
\hline
 Calo4pQVAE & FPD & KPD \\
\hline
 MCMC (100k) & $(390.35 \pm 1.85) \times 10^{-3}$ & $(0.46 \pm 0.05) \times 10^{-3}$ \\
\hline
\end{tabular}
\end{table}

\textbf{Training:} We train our model for $250$ epochs. Each epoch typically has $625$ updating steps. The number of block Gibbs sampling steps was set to $3000$. During the first $50$ epochs we anneal the model parameters, such as those used in the Gumbel trick, from smooth to a sharp step, to mitigate the discontinuities in the gradient related to the use of discrete variables. After the annealing parameters have reached their final values, we train the model for the next $150$ epochs, after which we freeze the encoder and decoder parameters at epoch $200$, while the prior distribution parameters keep being updated. We use an Nvidia A100 GPU. Our experiments show that this last step was necessary for the RBM log-likelihood to saturate. In Fig. \ref{fig:RBM-LL} we show the RBM log-likelihood \textit{vs} epochs. The yellow star points to when the annealed parameters reached their final values, whereas the purple star points to when the encoder and decoder parameter were frozen, after which it is clear the log-likelihood saturates. To estimate the log-likelihood, after training, we used \textit{annealed importance sampling} and \textit{reverse annealed importance sampling} methods \cite{salakhutdinov2008learning, burda2015accurate} with an annealed step set to $1/30$. As the model updates its parameters, the encoded data used to estimate the log-likelihood will also be modified from epoch to epoch, hence, we stored the validation encoded data for each epoch in order to accurately estimate the RBM log-likelihood. During training we save an instance of the model every ten epochs.

\begin{figure}
    \centering
    \includegraphics[width=0.99\linewidth]{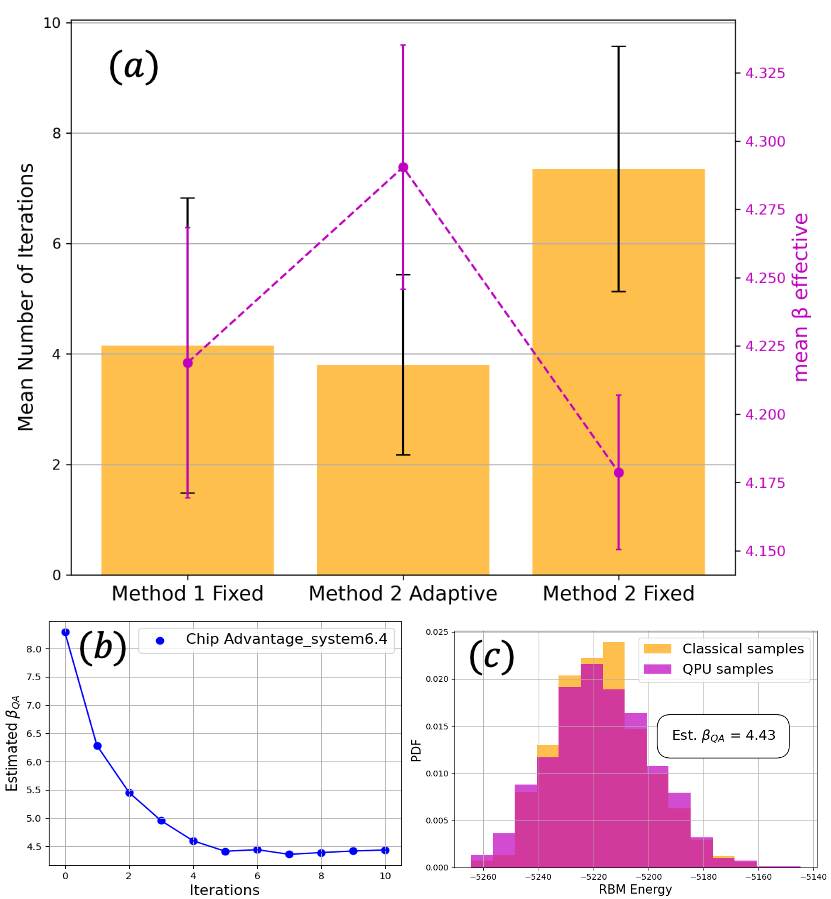}
    \caption{\textbf{a)} Mean number of iterations to meet the reduced standard error threshold in Eq. \eqref{eq:thrshld} using different iterative methods. Method 1 uses the KL divergence as in Eq. \eqref{eq:beta}. Method 2 and 3 use Eq. \eqref{eq:HaoMethod}, in addition Method 2 adapts the $\delta$ parameter after each iteration such that $\lambda \approx 0$ (see Eq. \eqref{eq:stability_maintext}). The dashed purple line corresponds to the ratio between final minus initial effective $\beta$ and the number of iterations, such that higher values implies faster convergence. The bars correspond to standard deviation. \textbf{b)} Estimated inverse temperature \textit{vs} iterations using method 2 adaptive. \textbf{c)} RBM histogram using classically generated and QA-generated samples after estimated temperature convergence.}
    \label{fig:beta_comparison}
\end{figure}

\textbf{Validation:} To validate our model, we use Fr\'echet physics distance (FPD) and the kernel physics distance (KPD) \cite{kansal2023evaluating}. These are \textit{integral probability} metrics for high energy physics, specifically designed to be sensitive to modelling of shower shape variables. In Fig. \ref{fig:RBM-LL} we show these metrics \textit{vs} epochs. Henceforth, the results are generated using the $240$-epoch model, corresponding to the best KPD metric with good FPD metric in the saturated RBM log-likelihood regime.
We further validate our model's reconstruction, classical sampling and QA sampling with \textsc{Geant4} data. To sample using the Advantage\_system6.4 QA \cite{boothby2020next}, we first estimate the effective $\beta$ by iteratively using Eq. \eqref{eq:HaoMethod} until the absolute difference between the QA's energy and that of the RBM is smaller than the reduced standard error, \textit{viz.},
\begin{equation}
    | \langle H \rangle_{QA} - \langle H \rangle_{RBM} | < \frac{2}{\sqrt{N}} \frac{\sigma_{QA} \sigma_{RBM}}{\sigma_{RBM} + \sigma_{QA}} \; . \label{eq:thrshld}
\end{equation}
In Fig. \ref{fig:beta_comparison} we show the mean number of iterations to meet the reduced standard error threshold in Eq. \eqref{eq:thrshld} using different iterative methods. Method 1 uses the KL divergence as in Eq. \eqref{eq:beta}. Method 2 uses Eq. \eqref{eq:HaoMethod}, where Method 2 \textit{adaptive} adapts the $\delta$ parameter after each iteration such that $\lambda \approx 0$ from Eq. \eqref{eq:stability_maintext}. The dashed purple line corresponds to the ratio between final minus initial effective $\beta$ and the number of iterations, such that higher values implies faster convergence.

\begin{figure*}
    \centering
    \includegraphics[width=0.79\linewidth]{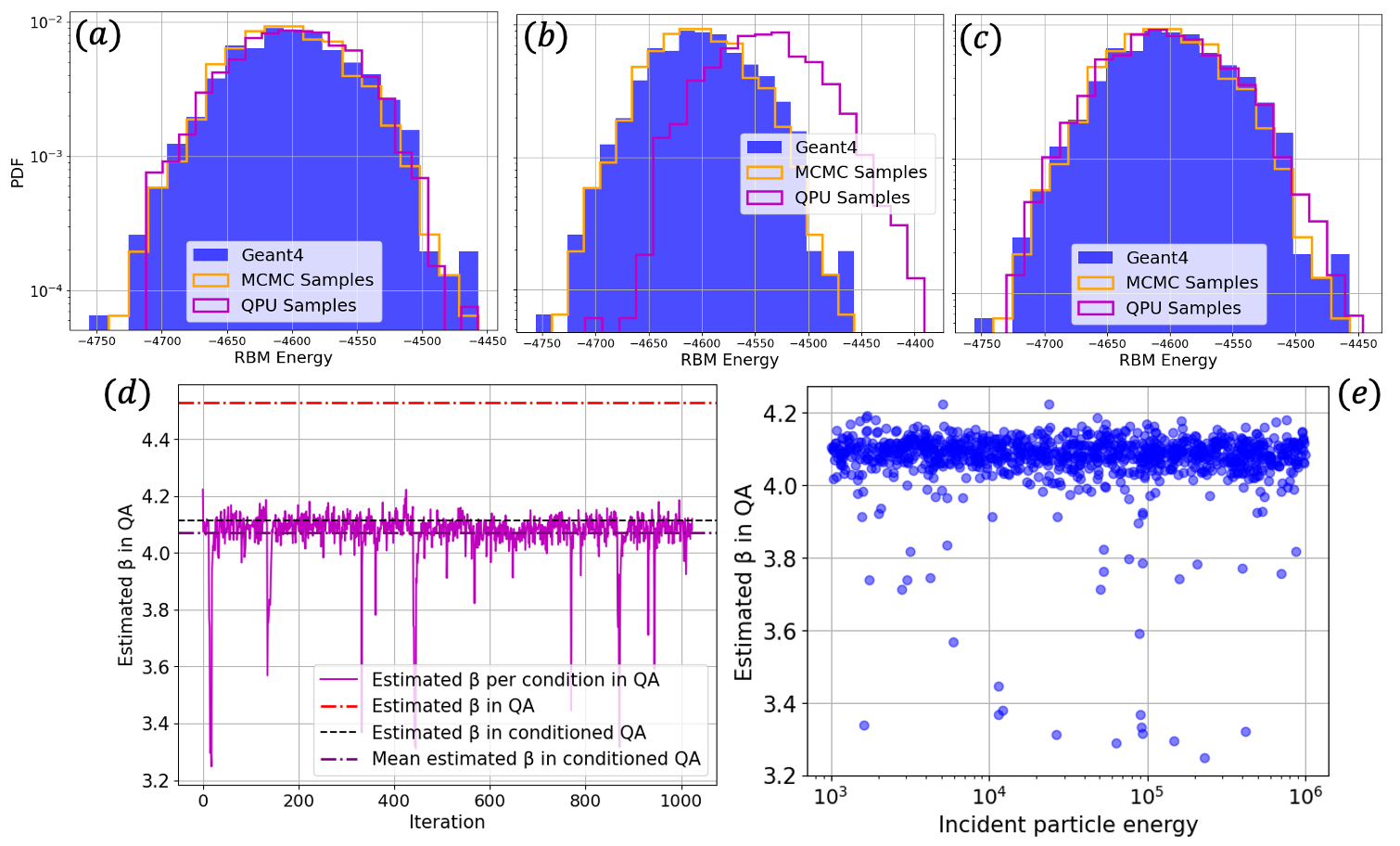}
    \caption{RBM energy histogram obtained from encoded \textsc{Geant4} showers, classically generated samples and QA-generated samples. \textbf{a)}  QA samples obtained via inverse temperature estimation per incident energy condition, \textbf{b)} single estimated inverse temperature, \textbf{c)} single estimated inverse temperature and a sleep time of 2.5s between API call. \textbf{d)} Estimated inverse temperature per incident energy condition in solid purple, estimated inverse temperature in the absence of flux biases in dashed red, estimated inverse temperature for single incident energy condition used in \textbf{b)} and \textbf{c)}. \textbf{e)} Estimated inverse temperature \textit{vs} incident energy condition.}
    \label{fig:hists_cond}
\end{figure*}

In each API call the user specifies the bias and coupler parameters, as well as the number of samples to be generated. The QA is programmed once using these biases and couplers and then performs the annealing process sequentially for the number of iterations requested, returning samples for each anneal. In addition, the user has the option to specify the flux bias parameters with each API call.

In the case of conditioned sampling using the QA, we estimate the temperature under flux biases. 
We observed that using flux biases increases the QA's effective temperature. Additionally, programming the QA increases temperature fluctuations. 
In Fig. \ref{fig:hists_cond} (a-c) we present the RBM energy histogram obtained from encoded \textsc{Geant4} showers, classically generated samples and QA-generated showers. Specifically, in Fig. \ref{fig:hists_cond} a) we estimate the inverse temperature for each incident energy condition, as outlined previously, before generating the QA sample. Fig. \ref{fig:hists_cond} d) shows the estimated inverse temperature per incident energy condition. Notably, we observe rather large but rare thermal fluctuations of the order of $10\%$.  In Fig. \ref{fig:hists_cond} e) we show the same estimated inverse temperature shown in Fig. \ref{fig:hists_cond} d) versus the incident energy condition, suggesting that these large thermal fluctuations are independent of the incident energy condition. To further investigate these effects, we repeated the process of generating conditioned samples using the QA, but estimated the QA inverse temperature only once at the beginning of the sampling process. This estimate is depicted as a dashed black line Fig. \ref{fig:hists_cond} d). In Fig. \ref{fig:hists_cond} b) we show the RBM energy, where a small shift in the energy of the QA samples is noticeable. Using the same inverse temperature obtained in the previous setting, we generated new samples and introduced a $2.5$s pause between QA API calls, that is per sample. In Fig. \ref{fig:hists_cond} c) we show the RBM energy obtained from these new samples, from which it is clear that the energy shift has disappeared. Additionally, in Fig. \ref{fig:hists_cond} d) we show the estimated inverse temperature in the absence of flux biases in dashed red. We conclude that there are two main contributions to the thermal fluctuations: the programming of the QA and the flux biases. The former can be substantially mitigated by either pausing the sampling process between samples or estimating the inverse temperature per sample. The latter can be accounted for by estimating the inverse temperature in the presence of flux biases. We discuss this further in the Discussion section.

\begin{figure*}[hbtp]
\centering
\includegraphics[width=0.9\linewidth]{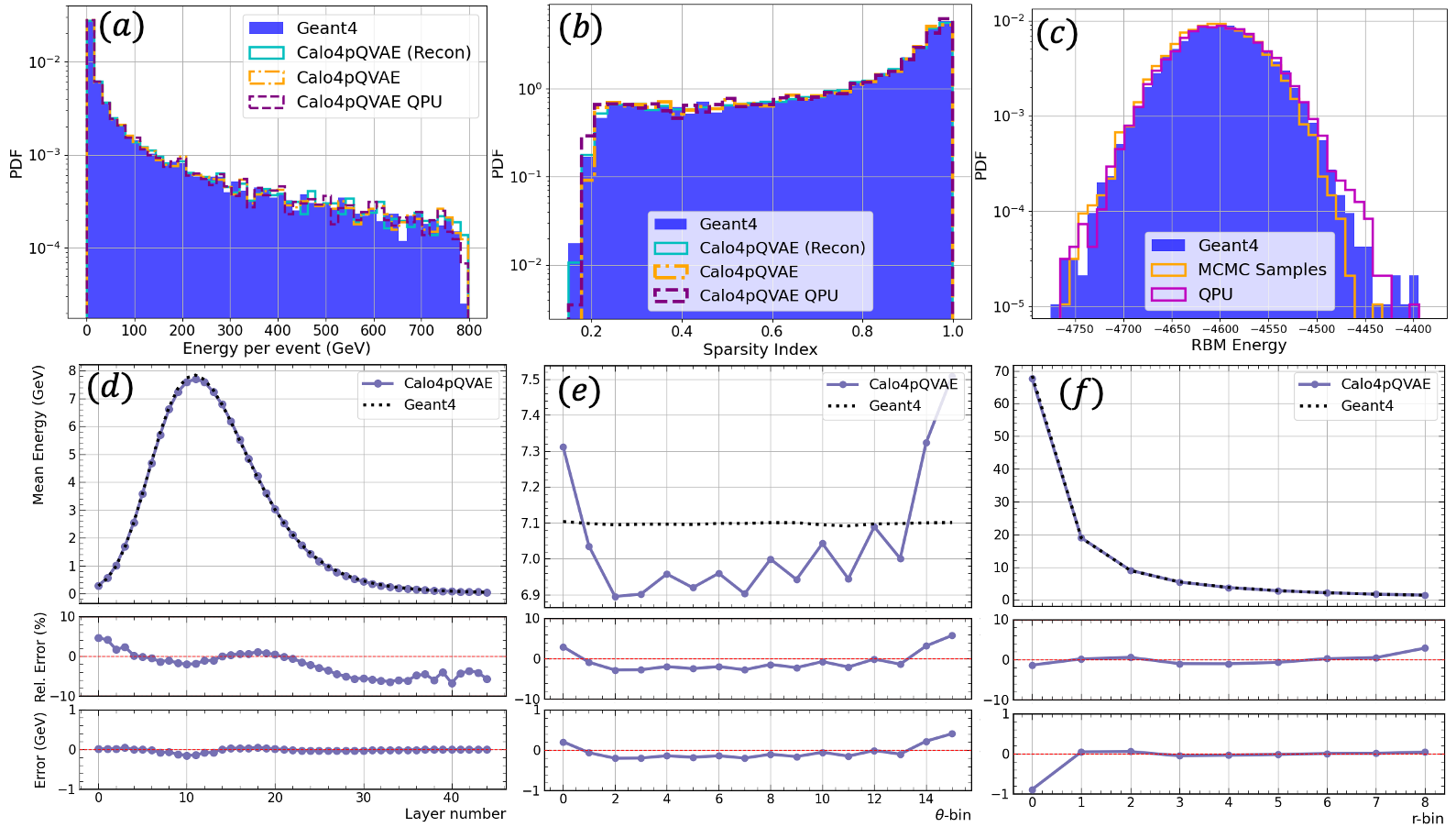}
\caption{ Comparison between \textsc{Geant4} samples, reconstructed samples using Calo4pQVAE, classically generated and QA-generated samples. \textbf{a)} Energy per event histogram, \textbf{b)} sparsity index per event, \textbf{c)} RBM energy per event. Comparison between \textsc{Geant4} and classically generated samples. Mean event energy \textit{vs} layer number \textbf{(panel d)}, \textit{vs} angular number \textbf{(panel e)}, and \textit{vs} radial number \textbf{(panel f)}. Each panel shows the relative error and the error underneath. The average is take from 100k events.} \label{fig:plot-panel}
\end{figure*}

After estimating the effective $\beta$ in the QA, we use the validation dataset composed of $10,000$ data points to benchmark the model. In Fig. \ref{fig:plot-panel}  (a) we show the shower energy histograms, while in Fig. \ref{fig:plot-panel}  (b) we show the sparsity histogram. The sparsity index is a measure of the sparsity of the shower and we define it as the ratio between zero-value voxels in the shower and total number of voxels. Each of the histograms include the \textsc{Geant4} data and the model's reconstruction. Furthermore, we test the generative model by generating samples from the RBM and feeding these samples together with the incidence energy from the validation dataset to the decoder and generating shower samples. Similarly, we test the quantum-assisted generative model by generating samples from the QA instead of the classical RBM. Both, the classically generated samples and the quantum-assisted generated ones are included in Figs. \ref{fig:plot-panel} (a) and (b) and are labelled \textit{Sample} and \textit{Sample w/ QPU}, respectively. In Fig. \ref{fig:plot-panel} (c) we show the RBM energy distribution corresponding to the encoded validation dataset and, both, the classically and QA sampled data. 

To further validate our model, we generate a synthetic dataset composed by $100,000$ events with the same incidence energy distribution as the training dataset. We compare our synthetic dataset with Dataset 2's test set \cite{calochallenge} by computing the mean energy in the $r,\theta$ and $z$ directions of the cylinder (see Fig. \ref{fig:dataset}). We show these results in Figs. \ref{fig:plot-panel} (d-f).

\section{Discussion}
In the previous section we outlined the process of training our conditioned quantum-assisted Calo4pQVAE and described a novel and faster method to estimate the QA's effective inverse temperature before sampling using the QPU. We evaluated the performance of our model by means of the KPD and FPD metrics, achieving results on the same order and one order of magnitude higher than CaloDiffusion, respectively, as shown in Table \ref{table_metrics} \cite{amram2023denoising}. Furthermore, when compared with the models in the CaloChallenge under this metric, our framework performs better than more than half of the 18 models considered \cite{krause2024calochallenge}. Our results demonstrate that \textit{i)} our framework is able of reconstructing the showers and preserving sparsity, and \textit{ii)} the prior is effectively learning the structure of the encoded data. 

Another critical metric to evaluate is the shower generation time, which in our framework depends on the decoder processing time and the RBM generation time. Estimating the RBM generation time is not straightforward and there are multiple approaches for this task. This estimation depends on the size of the RBM, its coordination number and the values of the couplings and biases, among other things. Furthermore, it has been shown that the RBM mixing time increases with training \cite{decelle2021equilibrium}. On the other hand, as mentioned earlier, QA can suffer from freeze-out, depending on features such as the coupling and bias values, the coordination number and the annealing time. Therefore, benchmarking the classical and quantum-assisted generation times requires careful consideration, and we leave this research for a future paper. Ultimately, both frameworks are several orders of magnitude faster than \textsc{Geant4}.

We previously mentioned that we set the number of block Gibbs sampling steps to 3000. We observed that with this number of steps, the RBM log-likelihood versus epochs increased monotonically.
In this case, generating $1024$ samples classically takes approximately $1$ s with $\mathcal{O}(1000)$ MCMC steps. Conversely, when sampling with QA, there are three characteristic timescales to consider: the programming time (approximately $ 10 $ ms), the annealing time (approximately $ 20 \mu s$) and the readout time (approximately $ 100 \mu s$). This results in a total generation time of approximately $0.1$ s for $\mathcal{O}(1000)$ samples, \textit{i.e.}, one order of magnitude faster than the classical method, assuming the QPU programming step is performed only once. These estimates per sample are shown in Table \ref{table_example}, which are highly competitive \cite{krause2024calochallenge}. However, in our current framework, as mentioned earlier, the flux biases conditioning is done during the programming step, effectively increasing the overall time to approximately $10$s.

Future iterations of D-wave's QA are expected to decouple the flux biases conditioning from the programming step. Therefore, we anticipate that the quantum-assisted framework to be competitive if the flux biasing conditioning remain below the order of milliseconds. Additionally, uncoupling the flux biases step from the QA programming will help mitigate the undesired large temperature fluctuations. Furthermore, since the self-correlation time is bound to increase with the coordination number, we expect the competitiveness of QAs to improve as the number of couplers per qubit increases.

\begin{table}[!t]
\renewcommand{\arraystretch}{1.3}
\caption{Shower generation time estimates using \textsc{Geant4} \cite{mikuni2024caloscore}, Calo4pQVAE on GPUs (assuming $3k$ BGS) and Calo4pQVAE with QPU without conditioning. 
}
\label{table_example}
\centering
\begin{tabular}{c c c c c}
\hline
 & \textsc{Geant4} & GPU (A100) & QPU & Anneal time\\
\hline
Time & $\mathcal{O}(0.1)-\mathcal{O}(10^2)$ s & $\sim 2$ ms & $\sim 0.2$ ms & $\sim 0.02$ ms \\
\hline
\end{tabular}
\end{table}


In the immediate future, we will focus on three key aspects:
\begin{itemize}
    \item \textbf{Exploring RBM Configurations}: In the present work, we set the number of units in the RBM to 512 units per partition, which implies a compression factor in the Calo4pQVAE of approximately $30\%$, similar to that in Ref. \cite{winci2020path}. Increasing this number will require increasing the number of block Gibbs sampling steps, as shown in \cite{decelle2021equilibrium, fernandez2023disentangling}. We will explore different RBM sizes with varying numbers of block Gibbs sampling steps to optimize performance and computational efficiency.
    \item \textbf{Enhancing the Decoder Module}: We will investigate the use of hierarchical structures and skip connections in the decoder module, as these can have a positive effect on performance, as observed in diffusion models \cite{amram2023denoising}. Implementing these architectures may improve the model's ability to reconstruct complex data patterns.
    \item \textbf{Upgrading to the Latest Quantum Annealer}: We plan to replace the current QA with D-Wave's latest version, Advantage2\_prototype2.4 \cite{boothby2021zephyr}. Although it has a smaller number of qubits, it offers a greater number of couplers per qubit and reduced noise, which could enhance the quality of quantum simulations and overall model performance.
\end{itemize}
As a final comment, our framework combines deep generative models with quantum simulations via Quantum Annealers (QAs). We speculate that the transition to a QA presents new opportunities not only in the \textit{noisy intermediate-scale quantum} stage but also by paving the way toward utilizing large-scale quantum-coherent simulations \cite{king2024computational} as priors in deep generative models.

\section{Acknowledgments}
We gratefully acknowledge Jack Raymond, Mohammad Amin, Trevor Lanting and Mark Johnson for their feedback and discussions.
We gratefully acknowledge funding from the National Research Council (Canada) via Agreement AQC-002, Natural Sciences and Engineering Research Council (Canada), in particular, via Grants SAPPJ-2020-00032 and SAPPJ-2022-00020.
This research was supported in part by Perimeter Institute for Theoretical Physics. Research at Perimeter Institute is supported by the Government of Canada through the Department of Innovation, Science and Economic Development and by the Province of Ontario through the Ministry of Research, Innovation and Science. The University of Virginia acknowledges support from NSF 2212550 OAC Core: Smart Surrogates for High Performance Scientific Simulations and DE-SC0023452: FAIR Surrogate Benchmarks Supporting AI and Simulation Research. JQTM acknowledges a Mitacs Elevate Postdoctoral Fellowship (IT39533) with Perimeter Institute for Theoretical Physics.

\bibliography{references}

\begin{thebibliography}{64}%
\makeatletter
\providecommand \@ifxundefined [1]{%
 \@ifx{#1\undefined}
}%
\providecommand \@ifnum [1]{%
 \ifnum #1\expandafter \@firstoftwo
 \else \expandafter \@secondoftwo
 \fi
}%
\providecommand \@ifx [1]{%
 \ifx #1\expandafter \@firstoftwo
 \else \expandafter \@secondoftwo
 \fi
}%
\providecommand \natexlab [1]{#1}%
\providecommand \enquote  [1]{``#1''}%
\providecommand \bibnamefont  [1]{#1}%
\providecommand \bibfnamefont [1]{#1}%
\providecommand \citenamefont [1]{#1}%
\providecommand \href@noop [0]{\@secondoftwo}%
\providecommand \href [0]{\begingroup \@sanitize@url \@href}%
\providecommand \@href[1]{\@@startlink{#1}\@@href}%
\providecommand \@@href[1]{\endgroup#1\@@endlink}%
\providecommand \@sanitize@url [0]{\catcode `\\12\catcode `\$12\catcode
  `\&12\catcode `\#12\catcode `\^12\catcode `\_12\catcode `\%12\relax}%
\providecommand \@@startlink[1]{}%
\providecommand \@@endlink[0]{}%
\providecommand \url  [0]{\begingroup\@sanitize@url \@url }%
\providecommand \@url [1]{\endgroup\@href {#1}{\urlprefix }}%
\providecommand \urlprefix  [0]{URL }%
\providecommand \Eprint [0]{\href }%
\providecommand \doibase [0]{https://doi.org/}%
\providecommand \selectlanguage [0]{\@gobble}%
\providecommand \bibinfo  [0]{\@secondoftwo}%
\providecommand \bibfield  [0]{\@secondoftwo}%
\providecommand \translation [1]{[#1]}%
\providecommand \BibitemOpen [0]{}%
\providecommand \bibitemStop [0]{}%
\providecommand \bibitemNoStop [0]{.\EOS\space}%
\providecommand \EOS [0]{\spacefactor3000\relax}%
\providecommand \BibitemShut  [1]{\csname bibitem#1\endcsname}%
\let\auto@bib@innerbib\@empty
\bibitem [{\citenamefont {{ATLAS
  Collaboration}}(2022)}]{collaboration2022atlas}%
  \BibitemOpen
  \bibfield  {author} {\bibinfo {author} {\bibnamefont {{ATLAS
  Collaboration}}},\ }\href@noop {} {\emph {\bibinfo {title} {ATLAS software
  and computing HL-LHC roadmap}}},\ \bibinfo {type} {Tech. Rep.}\ (\bibinfo
  {institution} {Technical report, CERN, Geneva. http://cds. cern.
  ch/record/2802918},\ \bibinfo {year} {2022})\BibitemShut {NoStop}%
\bibitem [{\citenamefont {Rousseau}(2023)}]{rousseau2023experimental}%
  \BibitemOpen
  \bibfield  {author} {\bibinfo {author} {\bibfnamefont {D.}~\bibnamefont
  {Rousseau}},\ }\bibfield  {title} {\bibinfo {title} {Experimental particle
  physics and artificial intelligence},\ }in\ \href@noop {} {\emph {\bibinfo
  {booktitle} {Artificial Intelligence for Science: A Deep Learning
  Revolution}}}\ (\bibinfo  {publisher} {World Scientific},\ \bibinfo {year}
  {2023})\ pp.\ \bibinfo {pages} {447--464}\BibitemShut {NoStop}%
\bibitem [{\citenamefont {{Michele Faucci Giannelli, Gregor Kasieczka, Claudius
  Krause, Ben Nachman, Dalila Salamani, David Shih, Anna
  Zaborowska}}(2022)}]{calochallenge}%
  \BibitemOpen
  \bibfield  {author} {\bibinfo {author} {\bibnamefont {{Michele Faucci
  Giannelli, Gregor Kasieczka, Claudius Krause, Ben Nachman, Dalila Salamani,
  David Shih, Anna Zaborowska}}},\ }\href@noop {} {\bibinfo {title} {Fast
  calorimeter simulation challenge 2022 - dataset 1,2 and 3 [data set].
  zenodo.}},\ \bibinfo {howpublished}
  {\url{https://doi.org/10.5281/zenodo.8099322},
  \url{https://doi.org/10.5281/zenodo.6366271},
  \url{https://doi.org/10.5281/zenodo.6366324}} (\bibinfo {year} {2022}),\
  \bibinfo {note} {online; accessed TO FILL}\BibitemShut {NoStop}%
\bibitem [{\citenamefont {Agostinelli}\ \emph {et~al.}(2003)\citenamefont
  {Agostinelli}, \citenamefont {Allison}, \citenamefont {Amako}, \citenamefont
  {Apostolakis}, \citenamefont {Araujo}, \citenamefont {Arce}, \citenamefont
  {Asai}, \citenamefont {Axen}, \citenamefont {Banerjee}, \citenamefont
  {Barrand} \emph {et~al.}}]{agostinelli2003geant4}%
  \BibitemOpen
  \bibfield  {author} {\bibinfo {author} {\bibfnamefont {S.}~\bibnamefont
  {Agostinelli}}, \bibinfo {author} {\bibfnamefont {J.}~\bibnamefont
  {Allison}}, \bibinfo {author} {\bibfnamefont {K.~a.}\ \bibnamefont {Amako}},
  \bibinfo {author} {\bibfnamefont {J.}~\bibnamefont {Apostolakis}}, \bibinfo
  {author} {\bibfnamefont {H.}~\bibnamefont {Araujo}}, \bibinfo {author}
  {\bibfnamefont {P.}~\bibnamefont {Arce}}, \bibinfo {author} {\bibfnamefont
  {M.}~\bibnamefont {Asai}}, \bibinfo {author} {\bibfnamefont {D.}~\bibnamefont
  {Axen}}, \bibinfo {author} {\bibfnamefont {S.}~\bibnamefont {Banerjee}},
  \bibinfo {author} {\bibfnamefont {G.}~\bibnamefont {Barrand}}, \emph
  {et~al.},\ }\bibfield  {title} {\bibinfo {title} {Geant4—a simulation
  toolkit},\ }\href@noop {} {\bibfield  {journal} {\bibinfo  {journal} {Nuclear
  instruments and methods in physics research section A: Accelerators,
  Spectrometers, Detectors and Associated Equipment}\ }\textbf {\bibinfo
  {volume} {506}},\ \bibinfo {pages} {250} (\bibinfo {year}
  {2003})}\BibitemShut {NoStop}%
\bibitem [{\citenamefont {de~Oliveira}\ \emph {et~al.}(2017)\citenamefont
  {de~Oliveira}, \citenamefont {Paganini},\ and\ \citenamefont
  {Nachman}}]{de2017learning}%
  \BibitemOpen
  \bibfield  {author} {\bibinfo {author} {\bibfnamefont {L.}~\bibnamefont
  {de~Oliveira}}, \bibinfo {author} {\bibfnamefont {M.}~\bibnamefont
  {Paganini}},\ and\ \bibinfo {author} {\bibfnamefont {B.}~\bibnamefont
  {Nachman}},\ }\bibfield  {title} {\bibinfo {title} {Learning particle physics
  by example: location-aware generative adversarial networks for physics
  synthesis},\ }\href@noop {} {\bibfield  {journal} {\bibinfo  {journal}
  {Computing and Software for Big Science}\ }\textbf {\bibinfo {volume} {1}},\
  \bibinfo {pages} {4} (\bibinfo {year} {2017})}\BibitemShut {NoStop}%
\bibitem [{\citenamefont {Paganini}\ \emph
  {et~al.}(2018{\natexlab{a}})\citenamefont {Paganini}, \citenamefont
  {de~Oliveira},\ and\ \citenamefont {Nachman}}]{paganini2018accelerating}%
  \BibitemOpen
  \bibfield  {author} {\bibinfo {author} {\bibfnamefont {M.}~\bibnamefont
  {Paganini}}, \bibinfo {author} {\bibfnamefont {L.}~\bibnamefont
  {de~Oliveira}},\ and\ \bibinfo {author} {\bibfnamefont {B.}~\bibnamefont
  {Nachman}},\ }\bibfield  {title} {\bibinfo {title} {Accelerating science with
  generative adversarial networks: an application to 3d particle showers in
  multilayer calorimeters},\ }\href@noop {} {\bibfield  {journal} {\bibinfo
  {journal} {Physical review letters}\ }\textbf {\bibinfo {volume} {120}},\
  \bibinfo {pages} {042003} (\bibinfo {year} {2018}{\natexlab{a}})}\BibitemShut
  {NoStop}%
\bibitem [{\citenamefont {Paganini}\ \emph
  {et~al.}(2018{\natexlab{b}})\citenamefont {Paganini}, \citenamefont
  {de~Oliveira},\ and\ \citenamefont {Nachman}}]{paganini2018calogan}%
  \BibitemOpen
  \bibfield  {author} {\bibinfo {author} {\bibfnamefont {M.}~\bibnamefont
  {Paganini}}, \bibinfo {author} {\bibfnamefont {L.}~\bibnamefont
  {de~Oliveira}},\ and\ \bibinfo {author} {\bibfnamefont {B.}~\bibnamefont
  {Nachman}},\ }\bibfield  {title} {\bibinfo {title} {Calogan: Simulating 3d
  high energy particle showers in multilayer electromagnetic calorimeters with
  generative adversarial networks},\ }\href@noop {} {\bibfield  {journal}
  {\bibinfo  {journal} {Physical Review D}\ }\textbf {\bibinfo {volume} {97}},\
  \bibinfo {pages} {014021} (\bibinfo {year} {2018}{\natexlab{b}})}\BibitemShut
  {NoStop}%
\bibitem [{\citenamefont {{ATLAS collaboration}}\ \emph
  {et~al.}(2020)\citenamefont {{ATLAS collaboration}} \emph
  {et~al.}}]{atlas2020fast}%
  \BibitemOpen
  \bibfield  {author} {\bibinfo {author} {\bibnamefont {{ATLAS collaboration}}}
  \emph {et~al.},\ }\bibfield  {title} {\bibinfo {title} {Fast simulation of
  the {ATLAS} calorimeter system with generative adversarial networks},\
  }\href@noop {} {\bibfield  {journal} {\bibinfo  {journal} {ATLAS PUB Note,
  CERN, Geneva}\ } (\bibinfo {year} {2020})}\BibitemShut {NoStop}%
\bibitem [{\citenamefont {Aad}\ \emph {et~al.}(2022)\citenamefont {Aad},
  \citenamefont {Abbott}, \citenamefont {Abbott}, \citenamefont {Abud},
  \citenamefont {Abeling}, \citenamefont {Abhayasinghe}, \citenamefont {Abidi},
  \citenamefont {Aboulhorma}, \citenamefont {Abramowicz}, \citenamefont {Abreu}
  \emph {et~al.}}]{aad2022atlfast3}%
  \BibitemOpen
  \bibfield  {author} {\bibinfo {author} {\bibfnamefont {G.}~\bibnamefont
  {Aad}}, \bibinfo {author} {\bibfnamefont {B.}~\bibnamefont {Abbott}},
  \bibinfo {author} {\bibfnamefont {D.~C.}\ \bibnamefont {Abbott}}, \bibinfo
  {author} {\bibfnamefont {A.~A.}\ \bibnamefont {Abud}}, \bibinfo {author}
  {\bibfnamefont {K.}~\bibnamefont {Abeling}}, \bibinfo {author} {\bibfnamefont
  {D.~K.}\ \bibnamefont {Abhayasinghe}}, \bibinfo {author} {\bibfnamefont
  {S.~H.}\ \bibnamefont {Abidi}}, \bibinfo {author} {\bibfnamefont
  {A.}~\bibnamefont {Aboulhorma}}, \bibinfo {author} {\bibfnamefont
  {H.}~\bibnamefont {Abramowicz}}, \bibinfo {author} {\bibfnamefont
  {H.}~\bibnamefont {Abreu}}, \emph {et~al.},\ }\bibfield  {title} {\bibinfo
  {title} {Atlfast3: the next generation of fast simulation in {ATLAS}},\
  }\href@noop {} {\bibfield  {journal} {\bibinfo  {journal} {Computing and
  software for big science}\ }\textbf {\bibinfo {volume} {6}},\ \bibinfo
  {pages} {7} (\bibinfo {year} {2022})}\BibitemShut {NoStop}%
\bibitem [{\citenamefont {Buhmann}\ \emph {et~al.}(2021)\citenamefont
  {Buhmann}, \citenamefont {Diefenbacher}, \citenamefont {Eren}, \citenamefont
  {Gaede}, \citenamefont {Kasieczka}, \citenamefont {Korol},\ and\
  \citenamefont {Kr{\"u}ger}}]{buhmann2021decoding}%
  \BibitemOpen
  \bibfield  {author} {\bibinfo {author} {\bibfnamefont {E.}~\bibnamefont
  {Buhmann}}, \bibinfo {author} {\bibfnamefont {S.}~\bibnamefont
  {Diefenbacher}}, \bibinfo {author} {\bibfnamefont {E.}~\bibnamefont {Eren}},
  \bibinfo {author} {\bibfnamefont {F.}~\bibnamefont {Gaede}}, \bibinfo
  {author} {\bibfnamefont {G.}~\bibnamefont {Kasieczka}}, \bibinfo {author}
  {\bibfnamefont {A.}~\bibnamefont {Korol}},\ and\ \bibinfo {author}
  {\bibfnamefont {K.}~\bibnamefont {Kr{\"u}ger}},\ }\bibfield  {title}
  {\bibinfo {title} {Decoding photons: Physics in the latent space of a bib-ae
  generative network},\ }in\ \href@noop {} {\emph {\bibinfo {booktitle} {EPJ
  Web of Conferences}}},\ Vol.\ \bibinfo {volume} {251}\ (\bibinfo
  {organization} {EDP Sciences},\ \bibinfo {year} {2021})\ p.\ \bibinfo {pages}
  {03003}\BibitemShut {NoStop}%
\bibitem [{\citenamefont {{ATLAS collaboration}}\ \emph
  {et~al.}(2022)\citenamefont {{ATLAS collaboration}} \emph
  {et~al.}}]{atlas2022deep}%
  \BibitemOpen
  \bibfield  {author} {\bibinfo {author} {\bibnamefont {{ATLAS collaboration}}}
  \emph {et~al.},\ }\bibfield  {title} {\bibinfo {title} {Deep generative
  models for fast photon shower simulation in {ATLAS}},\ }\href@noop {}
  {\bibfield  {journal} {\bibinfo  {journal} {arXiv preprint arXiv:2210.06204}\
  } (\bibinfo {year} {2022})}\BibitemShut {NoStop}%
\bibitem [{\citenamefont {Salamani}\ \emph {et~al.}(2023)\citenamefont
  {Salamani}, \citenamefont {Zaborowska},\ and\ \citenamefont
  {Pokorski}}]{salamani2023metahep}%
  \BibitemOpen
  \bibfield  {author} {\bibinfo {author} {\bibfnamefont {D.}~\bibnamefont
  {Salamani}}, \bibinfo {author} {\bibfnamefont {A.}~\bibnamefont
  {Zaborowska}},\ and\ \bibinfo {author} {\bibfnamefont {W.}~\bibnamefont
  {Pokorski}},\ }\bibfield  {title} {\bibinfo {title} {Metahep: Meta learning
  for fast shower simulation of high energy physics experiments},\ }\href@noop
  {} {\bibfield  {journal} {\bibinfo  {journal} {Physics Letters B}\ }\textbf
  {\bibinfo {volume} {844}},\ \bibinfo {pages} {138079} (\bibinfo {year}
  {2023})}\BibitemShut {NoStop}%
\bibitem [{\citenamefont {Krause}\ and\ \citenamefont
  {Shih}(2021)}]{krause2021caloflow}%
  \BibitemOpen
  \bibfield  {author} {\bibinfo {author} {\bibfnamefont {C.}~\bibnamefont
  {Krause}}\ and\ \bibinfo {author} {\bibfnamefont {D.}~\bibnamefont {Shih}},\
  }\bibfield  {title} {\bibinfo {title} {Caloflow: fast and accurate generation
  of calorimeter showers with normalizing flows},\ }\href@noop {} {\bibfield
  {journal} {\bibinfo  {journal} {arXiv preprint arXiv:2106.05285}\ } (\bibinfo
  {year} {2021})}\BibitemShut {NoStop}%
\bibitem [{\citenamefont {Buckley}\ \emph {et~al.}(2024)\citenamefont
  {Buckley}, \citenamefont {Pang}, \citenamefont {Shih},\ and\ \citenamefont
  {Krause}}]{buckley2024inductive}%
  \BibitemOpen
  \bibfield  {author} {\bibinfo {author} {\bibfnamefont {M.~R.}\ \bibnamefont
  {Buckley}}, \bibinfo {author} {\bibfnamefont {I.}~\bibnamefont {Pang}},
  \bibinfo {author} {\bibfnamefont {D.}~\bibnamefont {Shih}},\ and\ \bibinfo
  {author} {\bibfnamefont {C.}~\bibnamefont {Krause}},\ }\bibfield  {title}
  {\bibinfo {title} {Inductive simulation of calorimeter showers with
  normalizing flows},\ }\href@noop {} {\bibfield  {journal} {\bibinfo
  {journal} {Physical Review D}\ }\textbf {\bibinfo {volume} {109}},\ \bibinfo
  {pages} {033006} (\bibinfo {year} {2024})}\BibitemShut {NoStop}%
\bibitem [{\citenamefont {Favaro}\ \emph {et~al.}(2024)\citenamefont {Favaro},
  \citenamefont {Ore}, \citenamefont {Schweitzer},\ and\ \citenamefont
  {Plehn}}]{favaro2024calodream}%
  \BibitemOpen
  \bibfield  {author} {\bibinfo {author} {\bibfnamefont {L.}~\bibnamefont
  {Favaro}}, \bibinfo {author} {\bibfnamefont {A.}~\bibnamefont {Ore}},
  \bibinfo {author} {\bibfnamefont {S.~P.}\ \bibnamefont {Schweitzer}},\ and\
  \bibinfo {author} {\bibfnamefont {T.}~\bibnamefont {Plehn}},\ }\bibfield
  {title} {\bibinfo {title} {Calodream - detector response emulation via
  attentive flow matching},\ }\href@noop {} {\bibfield  {journal} {\bibinfo
  {journal} {arXiv preprint arXiv:2405.09629}\ } (\bibinfo {year}
  {2024})}\BibitemShut {NoStop}%
\bibitem [{\citenamefont {Mikuni}\ and\ \citenamefont
  {Nachman}(2024)}]{mikuni2024caloscore}%
  \BibitemOpen
  \bibfield  {author} {\bibinfo {author} {\bibfnamefont {V.}~\bibnamefont
  {Mikuni}}\ and\ \bibinfo {author} {\bibfnamefont {B.}~\bibnamefont
  {Nachman}},\ }\bibfield  {title} {\bibinfo {title} {Caloscore v2: single-shot
  calorimeter shower simulation with diffusion models},\ }\href@noop {}
  {\bibfield  {journal} {\bibinfo  {journal} {Journal of Instrumentation}\
  }\textbf {\bibinfo {volume} {19}}\bibinfo  {number} { (02)},\ \bibinfo
  {pages} {P02001}}\BibitemShut {NoStop}%
\bibitem [{\citenamefont {Kobylianskii}\ \emph {et~al.}(2024)\citenamefont
  {Kobylianskii}, \citenamefont {Soybelman}, \citenamefont {Dreyer},\ and\
  \citenamefont {Gross}}]{kobylianskii2024calograph}%
  \BibitemOpen
\bibfield  {number} {  }\bibfield  {author} {\bibinfo {author} {\bibfnamefont
  {D.}~\bibnamefont {Kobylianskii}}, \bibinfo {author} {\bibfnamefont
  {N.}~\bibnamefont {Soybelman}}, \bibinfo {author} {\bibfnamefont
  {E.}~\bibnamefont {Dreyer}},\ and\ \bibinfo {author} {\bibfnamefont
  {E.}~\bibnamefont {Gross}},\ }\bibfield  {title} {\bibinfo {title}
  {Calograph: Graph-based diffusion model for fast shower generation in
  calorimeters with irregular geometry},\ }\href@noop {} {\bibfield  {journal}
  {\bibinfo  {journal} {arXiv preprint arXiv:2402.11575}\ } (\bibinfo {year}
  {2024})}\BibitemShut {NoStop}%
\bibitem [{\citenamefont {Liu}\ \emph {et~al.}(2024)\citenamefont {Liu},
  \citenamefont {Shimmin}, \citenamefont {Liu}, \citenamefont {Shlizerman},
  \citenamefont {Li},\ and\ \citenamefont {Hsu}}]{liu2024calo}%
  \BibitemOpen
  \bibfield  {author} {\bibinfo {author} {\bibfnamefont {Q.}~\bibnamefont
  {Liu}}, \bibinfo {author} {\bibfnamefont {C.}~\bibnamefont {Shimmin}},
  \bibinfo {author} {\bibfnamefont {X.}~\bibnamefont {Liu}}, \bibinfo {author}
  {\bibfnamefont {E.}~\bibnamefont {Shlizerman}}, \bibinfo {author}
  {\bibfnamefont {S.}~\bibnamefont {Li}},\ and\ \bibinfo {author}
  {\bibfnamefont {S.-C.}\ \bibnamefont {Hsu}},\ }\bibfield  {title} {\bibinfo
  {title} {Calo-vq: Vector-quantized two-stage generative model in calorimeter
  simulation},\ }\href@noop {} {\bibfield  {journal} {\bibinfo  {journal}
  {arXiv preprint arXiv:2405.06605}\ } (\bibinfo {year} {2024})}\BibitemShut
  {NoStop}%
\bibitem [{\citenamefont {Amram}\ and\ \citenamefont
  {Pedro}(2023)}]{amram2023denoising}%
  \BibitemOpen
  \bibfield  {author} {\bibinfo {author} {\bibfnamefont {O.}~\bibnamefont
  {Amram}}\ and\ \bibinfo {author} {\bibfnamefont {K.}~\bibnamefont {Pedro}},\
  }\bibfield  {title} {\bibinfo {title} {Denoising diffusion models with
  geometry adaptation for high fidelity calorimeter simulation},\ }\href@noop
  {} {\bibfield  {journal} {\bibinfo  {journal} {Physical Review D}\ }\textbf
  {\bibinfo {volume} {108}},\ \bibinfo {pages} {072014} (\bibinfo {year}
  {2023})}\BibitemShut {NoStop}%
\bibitem [{\citenamefont {Madula}\ and\ \citenamefont
  {Mikuni}(2023)}]{madulacalolatent}%
  \BibitemOpen
  \bibfield  {author} {\bibinfo {author} {\bibfnamefont {T.}~\bibnamefont
  {Madula}}\ and\ \bibinfo {author} {\bibfnamefont {V.~M.}\ \bibnamefont
  {Mikuni}},\ }\bibfield  {title} {\bibinfo {title} {Calolatent: Score-based
  generative modelling in the latent space for calorimeter shower generation},\
  }\href@noop {} {\bibfield  {journal} {\bibinfo  {journal} {Neurips
  ML4PhysicalSciences}\ } (\bibinfo {year} {2023})}\BibitemShut {NoStop}%
\bibitem [{\citenamefont {Krause}\ \emph {et~al.}(2024)\citenamefont {Krause},
  \citenamefont {Giannelli}, \citenamefont {Kasieczka}, \citenamefont
  {Nachman}, \citenamefont {Salamani}, \citenamefont {Shih}, \citenamefont
  {Zaborowska}, \citenamefont {Amram}, \citenamefont {Borras}, \citenamefont
  {Buckley} \emph {et~al.}}]{krause2024calochallenge}%
  \BibitemOpen
  \bibfield  {author} {\bibinfo {author} {\bibfnamefont {C.}~\bibnamefont
  {Krause}}, \bibinfo {author} {\bibfnamefont {M.~F.}\ \bibnamefont
  {Giannelli}}, \bibinfo {author} {\bibfnamefont {G.}~\bibnamefont
  {Kasieczka}}, \bibinfo {author} {\bibfnamefont {B.}~\bibnamefont {Nachman}},
  \bibinfo {author} {\bibfnamefont {D.}~\bibnamefont {Salamani}}, \bibinfo
  {author} {\bibfnamefont {D.}~\bibnamefont {Shih}}, \bibinfo {author}
  {\bibfnamefont {A.}~\bibnamefont {Zaborowska}}, \bibinfo {author}
  {\bibfnamefont {O.}~\bibnamefont {Amram}}, \bibinfo {author} {\bibfnamefont
  {K.}~\bibnamefont {Borras}}, \bibinfo {author} {\bibfnamefont {M.~R.}\
  \bibnamefont {Buckley}}, \emph {et~al.},\ }\bibfield  {title} {\bibinfo
  {title} {Calochallenge 2022: A community challenge for fast calorimeter
  simulation},\ }\href@noop {} {\bibfield  {journal} {\bibinfo  {journal}
  {arXiv preprint arXiv:2410.21611}\ } (\bibinfo {year} {2024})}\BibitemShut
  {NoStop}%
\bibitem [{\citenamefont {Kansal}\ \emph {et~al.}(2023)\citenamefont {Kansal},
  \citenamefont {Li}, \citenamefont {Duarte}, \citenamefont {Chernyavskaya},
  \citenamefont {Pierini}, \citenamefont {Orzari},\ and\ \citenamefont
  {Tomei}}]{kansal2023evaluating}%
  \BibitemOpen
  \bibfield  {author} {\bibinfo {author} {\bibfnamefont {R.}~\bibnamefont
  {Kansal}}, \bibinfo {author} {\bibfnamefont {A.}~\bibnamefont {Li}}, \bibinfo
  {author} {\bibfnamefont {J.}~\bibnamefont {Duarte}}, \bibinfo {author}
  {\bibfnamefont {N.}~\bibnamefont {Chernyavskaya}}, \bibinfo {author}
  {\bibfnamefont {M.}~\bibnamefont {Pierini}}, \bibinfo {author} {\bibfnamefont
  {B.}~\bibnamefont {Orzari}},\ and\ \bibinfo {author} {\bibfnamefont
  {T.}~\bibnamefont {Tomei}},\ }\bibfield  {title} {\bibinfo {title}
  {Evaluating generative models in high energy physics},\ }\href@noop {}
  {\bibfield  {journal} {\bibinfo  {journal} {Physical Review D}\ }\textbf
  {\bibinfo {volume} {107}},\ \bibinfo {pages} {076017} (\bibinfo {year}
  {2023})}\BibitemShut {NoStop}%
\bibitem [{\citenamefont {Ahmad}\ \emph {et~al.}(2024)\citenamefont {Ahmad},
  \citenamefont {Venkataswamy},\ and\ \citenamefont
  {Fox}}]{ahmad2024comprehensive}%
  \BibitemOpen
  \bibfield  {author} {\bibinfo {author} {\bibfnamefont {F.~Y.}\ \bibnamefont
  {Ahmad}}, \bibinfo {author} {\bibfnamefont {V.}~\bibnamefont
  {Venkataswamy}},\ and\ \bibinfo {author} {\bibfnamefont {G.}~\bibnamefont
  {Fox}},\ }\bibfield  {title} {\bibinfo {title} {A comprehensive evaluation of
  generative models in calorimeter shower simulation},\ }\href@noop {}
  {\bibfield  {journal} {\bibinfo  {journal} {arXiv preprint arXiv:2406.12898}\
  } (\bibinfo {year} {2024})}\BibitemShut {NoStop}%
\bibitem [{\citenamefont {Hoque}\ \emph {et~al.}(2024)\citenamefont {Hoque},
  \citenamefont {Jia}, \citenamefont {Abhishek}, \citenamefont {Fadaie},
  \citenamefont {Toledo-Mar{\'\i}n}, \citenamefont {Vale}, \citenamefont
  {Melko}, \citenamefont {Swiatlowski},\ and\ \citenamefont
  {Fedorko}}]{hoque2023caloqvae}%
  \BibitemOpen
  \bibfield  {author} {\bibinfo {author} {\bibfnamefont {S.}~\bibnamefont
  {Hoque}}, \bibinfo {author} {\bibfnamefont {H.}~\bibnamefont {Jia}}, \bibinfo
  {author} {\bibfnamefont {A.}~\bibnamefont {Abhishek}}, \bibinfo {author}
  {\bibfnamefont {M.}~\bibnamefont {Fadaie}}, \bibinfo {author} {\bibfnamefont
  {J.~Q.}\ \bibnamefont {Toledo-Mar{\'\i}n}}, \bibinfo {author} {\bibfnamefont
  {T.}~\bibnamefont {Vale}}, \bibinfo {author} {\bibfnamefont {R.~G.}\
  \bibnamefont {Melko}}, \bibinfo {author} {\bibfnamefont {M.}~\bibnamefont
  {Swiatlowski}},\ and\ \bibinfo {author} {\bibfnamefont {W.~T.}\ \bibnamefont
  {Fedorko}},\ }\bibfield  {title} {\bibinfo {title} {Caloqvae: Simulating
  high-energy particle-calorimeter interactions using hybrid quantum-classical
  generative models},\ }\href@noop {} {\bibfield  {journal} {\bibinfo
  {journal} {The European Physical Journal C}\ }\textbf {\bibinfo {volume}
  {84}},\ \bibinfo {pages} {1} (\bibinfo {year} {2024})}\BibitemShut {NoStop}%
\bibitem [{\citenamefont {Gonzalez}\ \emph {et~al.}(2024)\citenamefont
  {Gonzalez}, \citenamefont {Jia}, \citenamefont {Toledo-Marin}, \citenamefont
  {Hoque}, \citenamefont {Abhishek}, \citenamefont {Lu}, \citenamefont
  {Sogutlu}, \citenamefont {Andersen}, \citenamefont {Gay}, \citenamefont
  {Paquet}, \citenamefont {Melko}, \citenamefont {Fox}, \citenamefont
  {Swiatlowski},\ and\ \citenamefont {Fedorko}}]{gonzalez2024caloqvae}%
  \BibitemOpen
  \bibfield  {author} {\bibinfo {author} {\bibfnamefont {S.}~\bibnamefont
  {Gonzalez}}, \bibinfo {author} {\bibfnamefont {H.}~\bibnamefont {Jia}},
  \bibinfo {author} {\bibfnamefont {J.~Q.}\ \bibnamefont {Toledo-Marin}},
  \bibinfo {author} {\bibfnamefont {S.}~\bibnamefont {Hoque}}, \bibinfo
  {author} {\bibfnamefont {A.}~\bibnamefont {Abhishek}}, \bibinfo {author}
  {\bibfnamefont {I.}~\bibnamefont {Lu}}, \bibinfo {author} {\bibfnamefont
  {D.}~\bibnamefont {Sogutlu}}, \bibinfo {author} {\bibfnamefont
  {S.}~\bibnamefont {Andersen}}, \bibinfo {author} {\bibfnamefont
  {C.}~\bibnamefont {Gay}}, \bibinfo {author} {\bibfnamefont {E.}~\bibnamefont
  {Paquet}}, \bibinfo {author} {\bibfnamefont {R.}~\bibnamefont {Melko}},
  \bibinfo {author} {\bibfnamefont {G.}~\bibnamefont {Fox}}, \bibinfo {author}
  {\bibfnamefont {M.}~\bibnamefont {Swiatlowski}},\ and\ \bibinfo {author}
  {\bibfnamefont {W.}~\bibnamefont {Fedorko}},\ }\bibfield  {title} {\bibinfo
  {title} {Calo4pqvae: Quantum-assisted 4-partite vae surrogate for high energy
  particle-calorimeter interactions},\ }\href@noop {} {\bibfield  {journal}
  {\bibinfo  {journal} {2024 IEEE International Conference on Quantum Computing
  and Engineering (QCE)}\ } (\bibinfo {year} {2024})}\BibitemShut {NoStop}%
\bibitem [{\citenamefont {Boothby}\ \emph {et~al.}(2020)\citenamefont
  {Boothby}, \citenamefont {Bunyk}, \citenamefont {Raymond},\ and\
  \citenamefont {Roy}}]{boothby2020next}%
  \BibitemOpen
  \bibfield  {author} {\bibinfo {author} {\bibfnamefont {K.}~\bibnamefont
  {Boothby}}, \bibinfo {author} {\bibfnamefont {P.}~\bibnamefont {Bunyk}},
  \bibinfo {author} {\bibfnamefont {J.}~\bibnamefont {Raymond}},\ and\ \bibinfo
  {author} {\bibfnamefont {A.}~\bibnamefont {Roy}},\ }\bibfield  {title}
  {\bibinfo {title} {Next-generation topology of {D-wave} quantum processors},\
  }\href@noop {} {\bibfield  {journal} {\bibinfo  {journal} {arXiv preprint
  arXiv:2003.00133}\ } (\bibinfo {year} {2020})}\BibitemShut {NoStop}%
\bibitem [{Note1()}]{Note1}%
  \BibitemOpen
  \bibinfo {note} {Particle physics experiments typically use a right-hand
  coordinate system with the interaction point at the center of the detector
  and the $z$-axis pointing along the beam pipe. The $x$-axis points to the
  center of the accelerating ring, and the $y$-axis points upwards. Polar
  coordinates ($r$, $\phi $) are used to describe the transverse directions of
  the detector, with $\phi $ being the azimuthal angle around the $z$-axis. The
  pseudorapidity $\eta $ is defined relative to the polar angle $\theta $ as
  $\eta = - \ln {\tan {(\theta /2)}}$.}\BibitemShut {Stop}%
\bibitem [{\citenamefont {Giannelli}\ \emph {et~al.}(2023)\citenamefont
  {Giannelli}, \citenamefont {Kasieczka}, \citenamefont {Krause}, \citenamefont
  {Nachman}, \citenamefont {Salamani}, \citenamefont {Shih},\ and\
  \citenamefont {Zaborowska}}]{d1}%
  \BibitemOpen
  \bibfield  {author} {\bibinfo {author} {\bibfnamefont {M.~F.}\ \bibnamefont
  {Giannelli}}, \bibinfo {author} {\bibfnamefont {G.}~\bibnamefont
  {Kasieczka}}, \bibinfo {author} {\bibfnamefont {C.}~\bibnamefont {Krause}},
  \bibinfo {author} {\bibfnamefont {B.}~\bibnamefont {Nachman}}, \bibinfo
  {author} {\bibfnamefont {D.}~\bibnamefont {Salamani}}, \bibinfo {author}
  {\bibfnamefont {D.}~\bibnamefont {Shih}},\ and\ \bibinfo {author}
  {\bibfnamefont {A.}~\bibnamefont {Zaborowska}},\ }\bibfield  {title}
  {\bibinfo {title} {{Fast Calorimeter Simulation Challenge 2022 - Dataset
  1}},\ }\href {https://doi.org/10.5281/zenodo.8099322}
  {10.5281/zenodo.8099322} (\bibinfo {year} {2023})\BibitemShut {NoStop}%
\bibitem [{\citenamefont {Faucci~Giannelli}\ \emph
  {et~al.}(2022{\natexlab{a}})\citenamefont {Faucci~Giannelli}, \citenamefont
  {Kasieczka}, \citenamefont {Krause}, \citenamefont {Nachman}, \citenamefont
  {Salamani}, \citenamefont {Shih},\ and\ \citenamefont {Zaborowska}}]{d2}%
  \BibitemOpen
  \bibfield  {author} {\bibinfo {author} {\bibfnamefont {M.}~\bibnamefont
  {Faucci~Giannelli}}, \bibinfo {author} {\bibfnamefont {G.}~\bibnamefont
  {Kasieczka}}, \bibinfo {author} {\bibfnamefont {C.}~\bibnamefont {Krause}},
  \bibinfo {author} {\bibfnamefont {B.}~\bibnamefont {Nachman}}, \bibinfo
  {author} {\bibfnamefont {D.}~\bibnamefont {Salamani}}, \bibinfo {author}
  {\bibfnamefont {D.}~\bibnamefont {Shih}},\ and\ \bibinfo {author}
  {\bibfnamefont {A.}~\bibnamefont {Zaborowska}},\ }\bibfield  {title}
  {\bibinfo {title} {{Fast Calorimeter Simulation Challenge 2022 - Dataset
  2}},\ }\href {https://doi.org/10.5281/zenodo.6366271}
  {10.5281/zenodo.6366271} (\bibinfo {year} {2022}{\natexlab{a}})\BibitemShut
  {NoStop}%
\bibitem [{\citenamefont {Faucci~Giannelli}\ \emph
  {et~al.}(2022{\natexlab{b}})\citenamefont {Faucci~Giannelli}, \citenamefont
  {Kasieczka}, \citenamefont {Krause}, \citenamefont {Nachman}, \citenamefont
  {Salamani}, \citenamefont {Shih},\ and\ \citenamefont {Zaborowska}}]{d3}%
  \BibitemOpen
  \bibfield  {author} {\bibinfo {author} {\bibfnamefont {M.}~\bibnamefont
  {Faucci~Giannelli}}, \bibinfo {author} {\bibfnamefont {G.}~\bibnamefont
  {Kasieczka}}, \bibinfo {author} {\bibfnamefont {C.}~\bibnamefont {Krause}},
  \bibinfo {author} {\bibfnamefont {B.}~\bibnamefont {Nachman}}, \bibinfo
  {author} {\bibfnamefont {D.}~\bibnamefont {Salamani}}, \bibinfo {author}
  {\bibfnamefont {D.}~\bibnamefont {Shih}},\ and\ \bibinfo {author}
  {\bibfnamefont {A.}~\bibnamefont {Zaborowska}},\ }\bibfield  {title}
  {\bibinfo {title} {{Fast Calorimeter Simulation Challenge 2022 - Dataset
  3}},\ }\href {https://doi.org/10.5281/zenodo.6366324}
  {10.5281/zenodo.6366324} (\bibinfo {year} {2022}{\natexlab{b}})\BibitemShut
  {NoStop}%
\bibitem [{\citenamefont {Winci}\ \emph {et~al.}(2020)\citenamefont {Winci},
  \citenamefont {Buffoni}, \citenamefont {Sadeghi}, \citenamefont {Khoshaman},
  \citenamefont {Andriyash},\ and\ \citenamefont {Amin}}]{winci2020path}%
  \BibitemOpen
  \bibfield  {author} {\bibinfo {author} {\bibfnamefont {W.}~\bibnamefont
  {Winci}}, \bibinfo {author} {\bibfnamefont {L.}~\bibnamefont {Buffoni}},
  \bibinfo {author} {\bibfnamefont {H.}~\bibnamefont {Sadeghi}}, \bibinfo
  {author} {\bibfnamefont {A.}~\bibnamefont {Khoshaman}}, \bibinfo {author}
  {\bibfnamefont {E.}~\bibnamefont {Andriyash}},\ and\ \bibinfo {author}
  {\bibfnamefont {M.~H.}\ \bibnamefont {Amin}},\ }\bibfield  {title} {\bibinfo
  {title} {A path towards quantum advantage in training deep generative models
  with quantum annealers},\ }\href@noop {} {\bibfield  {journal} {\bibinfo
  {journal} {Machine Learning: Science and Technology}\ }\textbf {\bibinfo
  {volume} {1}},\ \bibinfo {pages} {045028} (\bibinfo {year}
  {2020})}\BibitemShut {NoStop}%
\bibitem [{\citenamefont {Gehring}\ \emph {et~al.}(2017)\citenamefont
  {Gehring}, \citenamefont {Auli}, \citenamefont {Grangier}, \citenamefont
  {Yarats},\ and\ \citenamefont {Dauphin}}]{gehring2017convolutional}%
  \BibitemOpen
  \bibfield  {author} {\bibinfo {author} {\bibfnamefont {J.}~\bibnamefont
  {Gehring}}, \bibinfo {author} {\bibfnamefont {M.}~\bibnamefont {Auli}},
  \bibinfo {author} {\bibfnamefont {D.}~\bibnamefont {Grangier}}, \bibinfo
  {author} {\bibfnamefont {D.}~\bibnamefont {Yarats}},\ and\ \bibinfo {author}
  {\bibfnamefont {Y.~N.}\ \bibnamefont {Dauphin}},\ }\bibfield  {title}
  {\bibinfo {title} {Convolutional sequence to sequence learning},\ }in\
  \href@noop {} {\emph {\bibinfo {booktitle} {International conference on
  machine learning}}}\ (\bibinfo {organization} {PMLR},\ \bibinfo {year}
  {2017})\ pp.\ \bibinfo {pages} {1243--1252}\BibitemShut {NoStop}%
\bibitem [{\citenamefont {Vaswani}\ \emph {et~al.}(2017)\citenamefont
  {Vaswani}, \citenamefont {Shazeer}, \citenamefont {Parmar}, \citenamefont
  {Uszkoreit}, \citenamefont {Jones}, \citenamefont {Gomez}, \citenamefont
  {Kaiser},\ and\ \citenamefont {Polosukhin}}]{vaswani2017attention}%
  \BibitemOpen
  \bibfield  {author} {\bibinfo {author} {\bibfnamefont {A.}~\bibnamefont
  {Vaswani}}, \bibinfo {author} {\bibfnamefont {N.}~\bibnamefont {Shazeer}},
  \bibinfo {author} {\bibfnamefont {N.}~\bibnamefont {Parmar}}, \bibinfo
  {author} {\bibfnamefont {J.}~\bibnamefont {Uszkoreit}}, \bibinfo {author}
  {\bibfnamefont {L.}~\bibnamefont {Jones}}, \bibinfo {author} {\bibfnamefont
  {A.~N.}\ \bibnamefont {Gomez}}, \bibinfo {author} {\bibfnamefont
  {{\L}.}~\bibnamefont {Kaiser}},\ and\ \bibinfo {author} {\bibfnamefont
  {I.}~\bibnamefont {Polosukhin}},\ }\bibfield  {title} {\bibinfo {title}
  {Attention is all you need},\ }\href@noop {} {\bibfield  {journal} {\bibinfo
  {journal} {Advances in neural information processing systems}\ }\textbf
  {\bibinfo {volume} {30}} (\bibinfo {year} {2017})}\BibitemShut {NoStop}%
\bibitem [{\citenamefont {He}\ \emph {et~al.}(2016)\citenamefont {He},
  \citenamefont {Zhang}, \citenamefont {Ren},\ and\ \citenamefont
  {Sun}}]{he2016deep}%
  \BibitemOpen
  \bibfield  {author} {\bibinfo {author} {\bibfnamefont {K.}~\bibnamefont
  {He}}, \bibinfo {author} {\bibfnamefont {X.}~\bibnamefont {Zhang}}, \bibinfo
  {author} {\bibfnamefont {S.}~\bibnamefont {Ren}},\ and\ \bibinfo {author}
  {\bibfnamefont {J.}~\bibnamefont {Sun}},\ }\bibfield  {title} {\bibinfo
  {title} {Deep residual learning for image recognition},\ }in\ \href@noop {}
  {\emph {\bibinfo {booktitle} {Proceedings of the IEEE conference on computer
  vision and pattern recognition}}}\ (\bibinfo {year} {2016})\ pp.\ \bibinfo
  {pages} {770--778}\BibitemShut {NoStop}%
\bibitem [{\citenamefont {Kingma}\ and\ \citenamefont
  {Welling}(2013)}]{kingma2013auto}%
  \BibitemOpen
  \bibfield  {author} {\bibinfo {author} {\bibfnamefont {D.~P.}\ \bibnamefont
  {Kingma}}\ and\ \bibinfo {author} {\bibfnamefont {M.}~\bibnamefont
  {Welling}},\ }\bibfield  {title} {\bibinfo {title} {Auto-encoding variational
  bayes},\ }\href@noop {} {\bibfield  {journal} {\bibinfo  {journal} {arXiv
  preprint arXiv:1312.6114}\ } (\bibinfo {year} {2013})}\BibitemShut {NoStop}%
\bibitem [{Note2()}]{Note2}%
  \BibitemOpen
  \bibinfo {note} {We removed the $\xi $ coefficient due to
  normalization.}\BibitemShut {Stop}%
\bibitem [{\citenamefont {Maddison}\ \emph {et~al.}(2016)\citenamefont
  {Maddison}, \citenamefont {Mnih},\ and\ \citenamefont
  {Teh}}]{maddison2016concrete}%
  \BibitemOpen
  \bibfield  {author} {\bibinfo {author} {\bibfnamefont {C.~J.}\ \bibnamefont
  {Maddison}}, \bibinfo {author} {\bibfnamefont {A.}~\bibnamefont {Mnih}},\
  and\ \bibinfo {author} {\bibfnamefont {Y.~W.}\ \bibnamefont {Teh}},\
  }\bibfield  {title} {\bibinfo {title} {The concrete distribution: A
  continuous relaxation of discrete random variables},\ }\href@noop {}
  {\bibfield  {journal} {\bibinfo  {journal} {arXiv preprint arXiv:1611.00712}\
  } (\bibinfo {year} {2016})}\BibitemShut {NoStop}%
\bibitem [{\citenamefont {Balog}\ \emph {et~al.}(2017)\citenamefont {Balog},
  \citenamefont {Tripuraneni}, \citenamefont {Ghahramani},\ and\ \citenamefont
  {Weller}}]{balog2017lost}%
  \BibitemOpen
  \bibfield  {author} {\bibinfo {author} {\bibfnamefont {M.}~\bibnamefont
  {Balog}}, \bibinfo {author} {\bibfnamefont {N.}~\bibnamefont {Tripuraneni}},
  \bibinfo {author} {\bibfnamefont {Z.}~\bibnamefont {Ghahramani}},\ and\
  \bibinfo {author} {\bibfnamefont {A.}~\bibnamefont {Weller}},\ }\bibfield
  {title} {\bibinfo {title} {Lost relatives of the gumbel trick},\ }in\
  \href@noop {} {\emph {\bibinfo {booktitle} {International Conference on
  Machine Learning}}}\ (\bibinfo {organization} {PMLR},\ \bibinfo {year}
  {2017})\ pp.\ \bibinfo {pages} {371--379}\BibitemShut {NoStop}%
\bibitem [{\citenamefont {Khoshaman}\ and\ \citenamefont
  {Amin}(2018)}]{khoshaman2018gumbolt}%
  \BibitemOpen
  \bibfield  {author} {\bibinfo {author} {\bibfnamefont {A.~H.}\ \bibnamefont
  {Khoshaman}}\ and\ \bibinfo {author} {\bibfnamefont {M.}~\bibnamefont
  {Amin}},\ }\bibfield  {title} {\bibinfo {title} {Gumbolt: Extending gumbel
  trick to boltzmann priors},\ }\href@noop {} {\bibfield  {journal} {\bibinfo
  {journal} {Advances in Neural Information Processing Systems}\ }\textbf
  {\bibinfo {volume} {31}} (\bibinfo {year} {2018})}\BibitemShut {NoStop}%
\bibitem [{\citenamefont {Van~der Maaten}\ and\ \citenamefont
  {Hinton}(2008)}]{van2008visualizing}%
  \BibitemOpen
  \bibfield  {author} {\bibinfo {author} {\bibfnamefont {L.}~\bibnamefont
  {Van~der Maaten}}\ and\ \bibinfo {author} {\bibfnamefont {G.}~\bibnamefont
  {Hinton}},\ }\bibfield  {title} {\bibinfo {title} {Visualizing data using
  t-sne.},\ }\href@noop {} {\bibfield  {journal} {\bibinfo  {journal} {Journal
  of machine learning research}\ }\textbf {\bibinfo {volume} {9}} (\bibinfo
  {year} {2008})}\BibitemShut {NoStop}%
\bibitem [{\citenamefont {Toledo-Mar{\'\i}n}\ and\ \citenamefont
  {Glazier}(2023)}]{toledo2023using}%
  \BibitemOpen
  \bibfield  {author} {\bibinfo {author} {\bibfnamefont {J.~Q.}\ \bibnamefont
  {Toledo-Mar{\'\i}n}}\ and\ \bibinfo {author} {\bibfnamefont {J.~A.}\
  \bibnamefont {Glazier}},\ }\bibfield  {title} {\bibinfo {title} {Using deep
  lsd to build operators in gans latent space with meaning in real space},\
  }\href@noop {} {\bibfield  {journal} {\bibinfo  {journal} {Plos one}\
  }\textbf {\bibinfo {volume} {18}},\ \bibinfo {pages} {e0287736} (\bibinfo
  {year} {2023})}\BibitemShut {NoStop}%
\bibitem [{\citenamefont {Mnih}\ \emph {et~al.}(2012)\citenamefont {Mnih},
  \citenamefont {Larochelle},\ and\ \citenamefont
  {Hinton}}]{mnih2012conditional}%
  \BibitemOpen
  \bibfield  {author} {\bibinfo {author} {\bibfnamefont {V.}~\bibnamefont
  {Mnih}}, \bibinfo {author} {\bibfnamefont {H.}~\bibnamefont {Larochelle}},\
  and\ \bibinfo {author} {\bibfnamefont {G.~E.}\ \bibnamefont {Hinton}},\
  }\bibfield  {title} {\bibinfo {title} {Conditional restricted {Boltzmann}
  machines for structured output prediction},\ }\href@noop {} {\bibfield
  {journal} {\bibinfo  {journal} {arXiv preprint arXiv:1202.3748}\ } (\bibinfo
  {year} {2012})}\BibitemShut {NoStop}%
\bibitem [{\citenamefont {Johnson}\ \emph {et~al.}(2011)\citenamefont
  {Johnson}, \citenamefont {Amin}, \citenamefont {Gildert}, \citenamefont
  {Lanting}, \citenamefont {Hamze}, \citenamefont {Dickson}, \citenamefont
  {Harris}, \citenamefont {Berkley}, \citenamefont {Johansson}, \citenamefont
  {Bunyk} \emph {et~al.}}]{johnson2011quantum}%
  \BibitemOpen
  \bibfield  {author} {\bibinfo {author} {\bibfnamefont {M.~W.}\ \bibnamefont
  {Johnson}}, \bibinfo {author} {\bibfnamefont {M.~H.}\ \bibnamefont {Amin}},
  \bibinfo {author} {\bibfnamefont {S.}~\bibnamefont {Gildert}}, \bibinfo
  {author} {\bibfnamefont {T.}~\bibnamefont {Lanting}}, \bibinfo {author}
  {\bibfnamefont {F.}~\bibnamefont {Hamze}}, \bibinfo {author} {\bibfnamefont
  {N.}~\bibnamefont {Dickson}}, \bibinfo {author} {\bibfnamefont
  {R.}~\bibnamefont {Harris}}, \bibinfo {author} {\bibfnamefont {A.~J.}\
  \bibnamefont {Berkley}}, \bibinfo {author} {\bibfnamefont {J.}~\bibnamefont
  {Johansson}}, \bibinfo {author} {\bibfnamefont {P.}~\bibnamefont {Bunyk}},
  \emph {et~al.},\ }\bibfield  {title} {\bibinfo {title} {Quantum annealing
  with manufactured spins},\ }\href@noop {} {\bibfield  {journal} {\bibinfo
  {journal} {Nature}\ }\textbf {\bibinfo {volume} {473}},\ \bibinfo {pages}
  {194} (\bibinfo {year} {2011})}\BibitemShut {NoStop}%
\bibitem [{\citenamefont {Sakurai}\ and\ \citenamefont
  {Napolitano}(2017)}]{sakurai2017jim}%
  \BibitemOpen
  \bibfield  {author} {\bibinfo {author} {\bibfnamefont {J.}~\bibnamefont
  {Sakurai}}\ and\ \bibinfo {author} {\bibfnamefont {J.}~\bibnamefont
  {Napolitano}},\ }\href@noop {} {\emph {\bibinfo {title} {Modern Quantum
  Mechanics}}}\ (\bibinfo  {publisher} {Cambridge University Press Cambridge},\
  \bibinfo {year} {2017})\ pp.\ \bibinfo {pages} {328--331}\BibitemShut
  {NoStop}%
\bibitem [{\citenamefont {R{\o}nnow}\ \emph {et~al.}(2014)\citenamefont
  {R{\o}nnow}, \citenamefont {Wang}, \citenamefont {Job}, \citenamefont
  {Boixo}, \citenamefont {Isakov}, \citenamefont {Wecker}, \citenamefont
  {Martinis}, \citenamefont {Lidar},\ and\ \citenamefont
  {Troyer}}]{ronnow2014defining}%
  \BibitemOpen
  \bibfield  {author} {\bibinfo {author} {\bibfnamefont {T.~F.}\ \bibnamefont
  {R{\o}nnow}}, \bibinfo {author} {\bibfnamefont {Z.}~\bibnamefont {Wang}},
  \bibinfo {author} {\bibfnamefont {J.}~\bibnamefont {Job}}, \bibinfo {author}
  {\bibfnamefont {S.}~\bibnamefont {Boixo}}, \bibinfo {author} {\bibfnamefont
  {S.~V.}\ \bibnamefont {Isakov}}, \bibinfo {author} {\bibfnamefont
  {D.}~\bibnamefont {Wecker}}, \bibinfo {author} {\bibfnamefont {J.~M.}\
  \bibnamefont {Martinis}}, \bibinfo {author} {\bibfnamefont {D.~A.}\
  \bibnamefont {Lidar}},\ and\ \bibinfo {author} {\bibfnamefont
  {M.}~\bibnamefont {Troyer}},\ }\bibfield  {title} {\bibinfo {title} {Defining
  and detecting quantum speedup},\ }\href@noop {} {\bibfield  {journal}
  {\bibinfo  {journal} {science}\ }\textbf {\bibinfo {volume} {345}},\ \bibinfo
  {pages} {420} (\bibinfo {year} {2014})}\BibitemShut {NoStop}%
\bibitem [{\citenamefont {Hauke}\ \emph {et~al.}(2020)\citenamefont {Hauke},
  \citenamefont {Katzgraber}, \citenamefont {Lechner}, \citenamefont
  {Nishimori},\ and\ \citenamefont {Oliver}}]{hauke2020perspectives}%
  \BibitemOpen
  \bibfield  {author} {\bibinfo {author} {\bibfnamefont {P.}~\bibnamefont
  {Hauke}}, \bibinfo {author} {\bibfnamefont {H.~G.}\ \bibnamefont
  {Katzgraber}}, \bibinfo {author} {\bibfnamefont {W.}~\bibnamefont {Lechner}},
  \bibinfo {author} {\bibfnamefont {H.}~\bibnamefont {Nishimori}},\ and\
  \bibinfo {author} {\bibfnamefont {W.~D.}\ \bibnamefont {Oliver}},\ }\bibfield
   {title} {\bibinfo {title} {Perspectives of quantum annealing: Methods and
  implementations},\ }\href@noop {} {\bibfield  {journal} {\bibinfo  {journal}
  {Reports on Progress in Physics}\ }\textbf {\bibinfo {volume} {83}},\
  \bibinfo {pages} {054401} (\bibinfo {year} {2020})}\BibitemShut {NoStop}%
\bibitem [{\citenamefont {Benedetti}\ \emph {et~al.}(2016)\citenamefont
  {Benedetti}, \citenamefont {Realpe-G{\'o}mez}, \citenamefont {Biswas},\ and\
  \citenamefont {Perdomo-Ortiz}}]{benedetti2016estimation}%
  \BibitemOpen
  \bibfield  {author} {\bibinfo {author} {\bibfnamefont {M.}~\bibnamefont
  {Benedetti}}, \bibinfo {author} {\bibfnamefont {J.}~\bibnamefont
  {Realpe-G{\'o}mez}}, \bibinfo {author} {\bibfnamefont {R.}~\bibnamefont
  {Biswas}},\ and\ \bibinfo {author} {\bibfnamefont {A.}~\bibnamefont
  {Perdomo-Ortiz}},\ }\bibfield  {title} {\bibinfo {title} {Estimation of
  effective temperatures in quantum annealers for sampling applications: A case
  study with possible applications in deep learning},\ }\href@noop {}
  {\bibfield  {journal} {\bibinfo  {journal} {Physical Review A}\ }\textbf
  {\bibinfo {volume} {94}},\ \bibinfo {pages} {022308} (\bibinfo {year}
  {2016})}\BibitemShut {NoStop}%
\bibitem [{\citenamefont {Amin}\ \emph {et~al.}(2023)\citenamefont {Amin},
  \citenamefont {King}, \citenamefont {Raymond}, \citenamefont {Harris},
  \citenamefont {Bernoudy}, \citenamefont {Berkley}, \citenamefont {Boothby},
  \citenamefont {Smirnov}, \citenamefont {Altomare}, \citenamefont {Babcock}
  \emph {et~al.}}]{amin2023quantum}%
  \BibitemOpen
  \bibfield  {author} {\bibinfo {author} {\bibfnamefont {M.~H.}\ \bibnamefont
  {Amin}}, \bibinfo {author} {\bibfnamefont {A.~D.}\ \bibnamefont {King}},
  \bibinfo {author} {\bibfnamefont {J.}~\bibnamefont {Raymond}}, \bibinfo
  {author} {\bibfnamefont {R.}~\bibnamefont {Harris}}, \bibinfo {author}
  {\bibfnamefont {W.}~\bibnamefont {Bernoudy}}, \bibinfo {author}
  {\bibfnamefont {A.~J.}\ \bibnamefont {Berkley}}, \bibinfo {author}
  {\bibfnamefont {K.}~\bibnamefont {Boothby}}, \bibinfo {author} {\bibfnamefont
  {A.}~\bibnamefont {Smirnov}}, \bibinfo {author} {\bibfnamefont
  {F.}~\bibnamefont {Altomare}}, \bibinfo {author} {\bibfnamefont
  {M.}~\bibnamefont {Babcock}}, \emph {et~al.},\ }\bibfield  {title} {\bibinfo
  {title} {Quantum error mitigation in quantum annealing},\ }\href@noop {}
  {\bibfield  {journal} {\bibinfo  {journal} {arXiv preprint arXiv:2311.01306}\
  } (\bibinfo {year} {2023})}\BibitemShut {NoStop}%
\bibitem [{\citenamefont {Amin}(2015)}]{amin2015searching}%
  \BibitemOpen
  \bibfield  {author} {\bibinfo {author} {\bibfnamefont {M.~H.}\ \bibnamefont
  {Amin}},\ }\bibfield  {title} {\bibinfo {title} {Searching for quantum
  speedup in quasistatic quantum annealers},\ }\href@noop {} {\bibfield
  {journal} {\bibinfo  {journal} {Physical Review A}\ }\textbf {\bibinfo
  {volume} {92}},\ \bibinfo {pages} {052323} (\bibinfo {year}
  {2015})}\BibitemShut {NoStop}%
\bibitem [{\citenamefont {Debenedetti}\ and\ \citenamefont
  {Stillinger}(2001)}]{debenedetti2001supercooled}%
  \BibitemOpen
  \bibfield  {author} {\bibinfo {author} {\bibfnamefont {P.~G.}\ \bibnamefont
  {Debenedetti}}\ and\ \bibinfo {author} {\bibfnamefont {F.~H.}\ \bibnamefont
  {Stillinger}},\ }\bibfield  {title} {\bibinfo {title} {Supercooled liquids
  and the glass transition},\ }\href@noop {} {\bibfield  {journal} {\bibinfo
  {journal} {Nature}\ }\textbf {\bibinfo {volume} {410}},\ \bibinfo {pages}
  {259} (\bibinfo {year} {2001})}\BibitemShut {NoStop}%
\bibitem [{\citenamefont {Marshall}\ \emph {et~al.}(2019)\citenamefont
  {Marshall}, \citenamefont {Venturelli}, \citenamefont {Hen},\ and\
  \citenamefont {Rieffel}}]{marshall2019power}%
  \BibitemOpen
  \bibfield  {author} {\bibinfo {author} {\bibfnamefont {J.}~\bibnamefont
  {Marshall}}, \bibinfo {author} {\bibfnamefont {D.}~\bibnamefont
  {Venturelli}}, \bibinfo {author} {\bibfnamefont {I.}~\bibnamefont {Hen}},\
  and\ \bibinfo {author} {\bibfnamefont {E.~G.}\ \bibnamefont {Rieffel}},\
  }\bibfield  {title} {\bibinfo {title} {Power of pausing: Advancing
  understanding of thermalization in experimental quantum annealers},\
  }\href@noop {} {\bibfield  {journal} {\bibinfo  {journal} {Physical Review
  Applied}\ }\textbf {\bibinfo {volume} {11}},\ \bibinfo {pages} {044083}
  (\bibinfo {year} {2019})}\BibitemShut {NoStop}%
\bibitem [{\citenamefont {{D-Wave Systems}}(2022)}]{DWaveSystems3}%
  \BibitemOpen
  \bibfield  {author} {\bibinfo {author} {\bibnamefont {{D-Wave Systems}}},\
  }\href@noop {} {\bibinfo {title} {Advantage processor overview}},\ \bibinfo
  {howpublished}
  {\url{https://www.dwavesys.com/media/3xvdipcn/14-1058a-a_advantage_processor_overview.pdf}}
  (\bibinfo {year} {2022}),\ \bibinfo {note} {accessed: 2023-11-07}\BibitemShut
  {NoStop}%
\bibitem [{\citenamefont {{D-Wave
  Systems}}(2023{\natexlab{a}})}]{DWaveSystems2}%
  \BibitemOpen
  \bibfield  {author} {\bibinfo {author} {\bibnamefont {{D-Wave Systems}}},\
  }\href@noop {} {\bibinfo {title} {Advantage: The first and only computer
  built for business}},\ \bibinfo {howpublished}
  {\url{https://www.dwavesys.com/media/htjclcey/advantage_datasheet_v10.pdf}}
  (\bibinfo {year} {2023}{\natexlab{a}}),\ \bibinfo {note} {accessed:
  2023-11-07}\BibitemShut {NoStop}%
\bibitem [{\citenamefont {{D-Wave
  Systems}}(2023{\natexlab{b}})}]{DWaveSystems}%
  \BibitemOpen
  \bibfield  {author} {\bibinfo {author} {\bibnamefont {{D-Wave Systems}}},\
  }\href@noop {} {\bibinfo {title} {D-wave system documentation}},\ \bibinfo
  {howpublished}
  {\url{https://docs.dwavesys.com/docs/latest/doc_physical_properties.html#anneal-schedules}}
  (\bibinfo {year} {2023}{\natexlab{b}}),\ \bibinfo {note} {accessed:
  2023-11-07}\BibitemShut {NoStop}%
\bibitem [{\citenamefont {Raymond}\ \emph {et~al.}(2016)\citenamefont
  {Raymond}, \citenamefont {Yarkoni},\ and\ \citenamefont
  {Andriyash}}]{raymond2016global}%
  \BibitemOpen
  \bibfield  {author} {\bibinfo {author} {\bibfnamefont {J.}~\bibnamefont
  {Raymond}}, \bibinfo {author} {\bibfnamefont {S.}~\bibnamefont {Yarkoni}},\
  and\ \bibinfo {author} {\bibfnamefont {E.}~\bibnamefont {Andriyash}},\
  }\bibfield  {title} {\bibinfo {title} {Global warming: Temperature estimation
  in annealers},\ }\href@noop {} {\bibfield  {journal} {\bibinfo  {journal}
  {Frontiers in ICT}\ }\textbf {\bibinfo {volume} {3}},\ \bibinfo {pages} {23}
  (\bibinfo {year} {2016})}\BibitemShut {NoStop}%
\bibitem [{\citenamefont {Harris}\ \emph {et~al.}(2010)\citenamefont {Harris},
  \citenamefont {Johnson}, \citenamefont {Lanting}, \citenamefont {Berkley},
  \citenamefont {Johansson}, \citenamefont {Bunyk}, \citenamefont {Tolkacheva},
  \citenamefont {Ladizinsky}, \citenamefont {Ladizinsky}, \citenamefont {Oh}
  \emph {et~al.}}]{harris2010experimental}%
  \BibitemOpen
  \bibfield  {author} {\bibinfo {author} {\bibfnamefont {R.}~\bibnamefont
  {Harris}}, \bibinfo {author} {\bibfnamefont {M.~W.}\ \bibnamefont {Johnson}},
  \bibinfo {author} {\bibfnamefont {T.}~\bibnamefont {Lanting}}, \bibinfo
  {author} {\bibfnamefont {A.}~\bibnamefont {Berkley}}, \bibinfo {author}
  {\bibfnamefont {J.}~\bibnamefont {Johansson}}, \bibinfo {author}
  {\bibfnamefont {P.}~\bibnamefont {Bunyk}}, \bibinfo {author} {\bibfnamefont
  {E.}~\bibnamefont {Tolkacheva}}, \bibinfo {author} {\bibfnamefont
  {E.}~\bibnamefont {Ladizinsky}}, \bibinfo {author} {\bibfnamefont
  {N.}~\bibnamefont {Ladizinsky}}, \bibinfo {author} {\bibfnamefont
  {T.}~\bibnamefont {Oh}}, \emph {et~al.},\ }\bibfield  {title} {\bibinfo
  {title} {Experimental investigation of an eight-qubit unit cell in a
  superconducting optimization processor},\ }\href@noop {} {\bibfield
  {journal} {\bibinfo  {journal} {Physical Review B}\ }\textbf {\bibinfo
  {volume} {82}},\ \bibinfo {pages} {024511} (\bibinfo {year}
  {2010})}\BibitemShut {NoStop}%
\bibitem [{\citenamefont {{Raghav Kansal, Javier Duarte1, Carlos Pareja, Lint
  Action, Zichun Hao, mova}}(2023)}]{jetlib}%
  \BibitemOpen
  \bibfield  {author} {\bibinfo {author} {\bibnamefont {{Raghav Kansal, Javier
  Duarte1, Carlos Pareja, Lint Action, Zichun Hao, mova}}},\ }\href@noop {}
  {\bibinfo {title} {jet-net/jetnet: v0.2.3.post3}},\ \bibinfo {howpublished}
  {\url{https://doi.org/10.5281/zenodo.5597892}} (\bibinfo {year} {2023}),\
  \bibinfo {note} {online; accessed July 2024}\BibitemShut {NoStop}%
\bibitem [{\citenamefont {Salakhutdinov}(2008)}]{salakhutdinov2008learning}%
  \BibitemOpen
  \bibfield  {author} {\bibinfo {author} {\bibfnamefont {R.}~\bibnamefont
  {Salakhutdinov}},\ }\bibfield  {title} {\bibinfo {title} {Learning and
  evaluating boltzmann machines},\ }\href@noop {} {\bibfield  {journal}
  {\bibinfo  {journal} {Utml Tr}\ }\textbf {\bibinfo {volume} {2}},\ \bibinfo
  {pages} {21} (\bibinfo {year} {2008})}\BibitemShut {NoStop}%
\bibitem [{\citenamefont {Burda}\ \emph {et~al.}(2015)\citenamefont {Burda},
  \citenamefont {Grosse},\ and\ \citenamefont
  {Salakhutdinov}}]{burda2015accurate}%
  \BibitemOpen
  \bibfield  {author} {\bibinfo {author} {\bibfnamefont {Y.}~\bibnamefont
  {Burda}}, \bibinfo {author} {\bibfnamefont {R.}~\bibnamefont {Grosse}},\ and\
  \bibinfo {author} {\bibfnamefont {R.}~\bibnamefont {Salakhutdinov}},\
  }\bibfield  {title} {\bibinfo {title} {Accurate and conservative estimates of
  mrf log-likelihood using reverse annealing},\ }in\ \href@noop {} {\emph
  {\bibinfo {booktitle} {Artificial Intelligence and Statistics}}}\ (\bibinfo
  {organization} {PMLR},\ \bibinfo {year} {2015})\ pp.\ \bibinfo {pages}
  {102--110}\BibitemShut {NoStop}%
\bibitem [{\citenamefont {Decelle}\ \emph {et~al.}(2021)\citenamefont
  {Decelle}, \citenamefont {Furtlehner},\ and\ \citenamefont
  {Seoane}}]{decelle2021equilibrium}%
  \BibitemOpen
  \bibfield  {author} {\bibinfo {author} {\bibfnamefont {A.}~\bibnamefont
  {Decelle}}, \bibinfo {author} {\bibfnamefont {C.}~\bibnamefont
  {Furtlehner}},\ and\ \bibinfo {author} {\bibfnamefont {B.}~\bibnamefont
  {Seoane}},\ }\bibfield  {title} {\bibinfo {title} {Equilibrium and
  non-equilibrium regimes in the learning of restricted boltzmann machines},\
  }\href@noop {} {\bibfield  {journal} {\bibinfo  {journal} {Advances in Neural
  Information Processing Systems}\ }\textbf {\bibinfo {volume} {34}},\ \bibinfo
  {pages} {5345} (\bibinfo {year} {2021})}\BibitemShut {NoStop}%
\bibitem [{\citenamefont {Fernandez-de Cossio-Diaz}\ \emph
  {et~al.}(2023)\citenamefont {Fernandez-de Cossio-Diaz}, \citenamefont
  {Cocco},\ and\ \citenamefont {Monasson}}]{fernandez2023disentangling}%
  \BibitemOpen
  \bibfield  {author} {\bibinfo {author} {\bibfnamefont {J.}~\bibnamefont
  {Fernandez-de Cossio-Diaz}}, \bibinfo {author} {\bibfnamefont
  {S.}~\bibnamefont {Cocco}},\ and\ \bibinfo {author} {\bibfnamefont
  {R.}~\bibnamefont {Monasson}},\ }\bibfield  {title} {\bibinfo {title}
  {Disentangling representations in restricted boltzmann machines without
  adversaries},\ }\href@noop {} {\bibfield  {journal} {\bibinfo  {journal}
  {Physical Review X}\ }\textbf {\bibinfo {volume} {13}},\ \bibinfo {pages}
  {021003} (\bibinfo {year} {2023})}\BibitemShut {NoStop}%
\bibitem [{\citenamefont {Boothby}\ \emph {et~al.}(2021)\citenamefont
  {Boothby}, \citenamefont {King},\ and\ \citenamefont
  {Raymond}}]{boothby2021zephyr}%
  \BibitemOpen
  \bibfield  {author} {\bibinfo {author} {\bibfnamefont {K.}~\bibnamefont
  {Boothby}}, \bibinfo {author} {\bibfnamefont {A.~D.}\ \bibnamefont {King}},\
  and\ \bibinfo {author} {\bibfnamefont {J.}~\bibnamefont {Raymond}},\
  }\bibfield  {title} {\bibinfo {title} {Zephyr topology of d-wave quantum
  processors},\ }\href@noop {} {\bibfield  {journal} {\bibinfo  {journal}
  {D-Wave Technical Report Series}\ } (\bibinfo {year} {2021})}\BibitemShut
  {NoStop}%
\bibitem [{\citenamefont {King}\ \emph {et~al.}(2024)\citenamefont {King},
  \citenamefont {Nocera}, \citenamefont {Rams}, \citenamefont {Dziarmaga},
  \citenamefont {Wiersema}, \citenamefont {Bernoudy}, \citenamefont {Raymond},
  \citenamefont {Kaushal}, \citenamefont {Heinsdorf}, \citenamefont {Harris}
  \emph {et~al.}}]{king2024computational}%
  \BibitemOpen
  \bibfield  {author} {\bibinfo {author} {\bibfnamefont {A.~D.}\ \bibnamefont
  {King}}, \bibinfo {author} {\bibfnamefont {A.}~\bibnamefont {Nocera}},
  \bibinfo {author} {\bibfnamefont {M.~M.}\ \bibnamefont {Rams}}, \bibinfo
  {author} {\bibfnamefont {J.}~\bibnamefont {Dziarmaga}}, \bibinfo {author}
  {\bibfnamefont {R.}~\bibnamefont {Wiersema}}, \bibinfo {author}
  {\bibfnamefont {W.}~\bibnamefont {Bernoudy}}, \bibinfo {author}
  {\bibfnamefont {J.}~\bibnamefont {Raymond}}, \bibinfo {author} {\bibfnamefont
  {N.}~\bibnamefont {Kaushal}}, \bibinfo {author} {\bibfnamefont
  {N.}~\bibnamefont {Heinsdorf}}, \bibinfo {author} {\bibfnamefont
  {R.}~\bibnamefont {Harris}}, \emph {et~al.},\ }\bibfield  {title} {\bibinfo
  {title} {Computational supremacy in quantum simulation},\ }\href@noop {}
  {\bibfield  {journal} {\bibinfo  {journal} {arXiv preprint arXiv:2403.00910}\
  } (\bibinfo {year} {2024})}\BibitemShut {NoStop}%
\bibitem [{\citenamefont {Box}\ and\ \citenamefont
  {Muller}(1958)}]{box1958note}%
  \BibitemOpen
  \bibfield  {author} {\bibinfo {author} {\bibfnamefont {G.~E.}\ \bibnamefont
  {Box}}\ and\ \bibinfo {author} {\bibfnamefont {M.~E.}\ \bibnamefont
  {Muller}},\ }\bibfield  {title} {\bibinfo {title} {A note on the generation
  of random normal deviates},\ }\href@noop {} {\bibfield  {journal} {\bibinfo
  {journal} {The annals of mathematical statistics}\ }\textbf {\bibinfo
  {volume} {29}},\ \bibinfo {pages} {610} (\bibinfo {year} {1958})}\BibitemShut
  {NoStop}%
\end{thebibliography}%
\onecolumngrid
\newpage
\appendix

\setcounter{figure}{0}
\section{Variational Autoencoder} \label{App:1}
In this section we describe the VAE framework first proposed by Kingma and Welling in \cite{kingma2013auto}.
Suppose we have a data set $\lbrace \bm{x}^{(i)} \rbrace_{i=1}^{|\mathcal{D}|}$, where each element in the data set lives in $\mathcal{R}^{N}$.
The goal in training a Variational Autoencoder (VAE) on this data set is to fit a probability distribution, $p(\bm{x})$, to the data. This is done by maximizing the log-likelihood (LL) of $p(\bm{x})$ over the data set. A key component in generative models is the introduction of latent variables, $\bm{z}$, such that the joint distribution can be expressed as $p(\bm{x},\bm{z})=p(\bm{x}|\bm{z})p(\bm{z})$, where $p(\bm{z})$ is the \textit{prior} distribution of $\bm{z} \in \mathcal{R}^{M}$. VAEs are composed by an encoder and a decoder, and are trained using the Evidence Lower Bound (ELBO) as a proxy loss function for the LL. To understand the relationship between the LL and the ELBO, we first write the following identity:
\begin{eqnarray}
    \ln p_\theta (\bm{x}) = \langle \ln p_\theta (\bm{x}) \rangle_{q_\phi (\bm{z}|\bm{x})} \label{eq:identity}
\end{eqnarray}
where $\langle \bullet  \rangle_{q_\phi (\bm{z}|\bm{x})}$ denotes expectation value of $\bullet$ over $q_\phi (\bm{z}|\bm{x})$. Here, $q_{\phi}(\bm{z}|\bm{x})$ is the encoding function, also known as the approximate posterior. This function encodes the data $\bm{x}$ into $\bm{z}$ in the latent space. We can further manipulate the r.h.s. in Eq. \eqref{eq:identity}, \textit{viz.},
\begin{eqnarray}
    \ln p_\theta (\bm{x}) &=& \langle \ln \frac{ p_\theta (\bm{x},\bm{z})}{p_\theta (\bm{z}|\bm{x})} \rangle_{q_\phi (\bm{z}|\bm{x})}  \nonumber \\
    &=& \langle \ln \frac{ p_\theta (\bm{x},\bm{z}) q_\phi(\bm{z}|\bm{x})}{q_\phi(\bm{z}|\bm{x}) p_\theta (\bm{z}|\bm{x})} \rangle_{q_\phi (\bm{z}|\bm{x})}  \nonumber \\
    &=& \langle \ln \frac{ p_\theta (\bm{x},\bm{z}) }{q_\phi(\bm{z}|\bm{x})} \rangle_{q_\phi (\bm{z}|\bm{x})} + \langle \ln \frac{ q_\phi(\bm{z}|\bm{x})}{ p_\theta (\bm{z}|\bm{x})} \rangle_{q_\phi (\bm{z}|\bm{x})} \nonumber \\
    &=& \mathcal{L}_{\phi, \theta}(\bm{x}) + D_{kl}( q_\phi(\bm{z}|\bm{x}) ||  p_\theta (\bm{z}|\bm{x})) \; .
\end{eqnarray}
In the last line in the previous Eq., $\mathcal{L}_{\phi, \theta}(\bm{x})$ is the ELBO and $D_{kl}( q_\phi(\bm{z}|\bm{x}) ||  p_\theta (\bm{z}|\bm{x}))$ is the Kullback-Liebler (KL) divergence. The KL divergence is a positive functional and equals zero when both distributions are the same. Therefore:
\begin{equation}
    \mathcal{L}_{\phi, \theta}(\bm{x}) = \ln p_\theta (\bm{x}) - D_{kl}( q_\phi(\bm{z}|\bm{x}) ||  p_\theta (\bm{z}|\bm{x})) \leq \ln p_\theta (\bm{x})
\end{equation}
The previous Eq. shows that maximizing the ELBO implies maximizing the LL (since the LL is the upper bound), as well as to minimizing the KL divergence between $q_\phi(\bm{z}|\bm{x})$ and $p_\theta (\bm{z}|\bm{x})$.

We can express the ELBO in a more tractable way:
\begin{eqnarray}
     \mathcal{L}_{\phi, \theta}(\bm{x}) &=& \langle \ln \frac{ p_\theta (\bm{x},\bm{z}) }{q_\phi(\bm{z}|\bm{x})} \rangle_{q_\phi (\bm{z}|\bm{x})}  \nonumber \\
     &=& \langle \ln p_\theta (\bm{x}|\bm{z})  \rangle_{q_\phi (\bm{z}|\bm{x})} 
     - \langle \ln \frac{ q_\phi(\bm{z}|\bm{x})}{ p_\theta (\bm{z}) } \rangle_{q_\phi (\bm{z}|\bm{x})} \; .
\end{eqnarray}
The first term in the last equality is called the reconstruction term since it is a measure of how well the model is able to reconstruct the input $\bm{x}$ from a latent vector $\bm{z}$. The second term in the last equality is called a regularizer and measures the divergence between the prior and the approximate posterior.

The legacy VAE \cite{kingma2013auto} assumes the functional forms:
\begin{subequations}
\begin{align}
    p_\theta(\bm{x}|\bm{z}) = \prod_{i=1}^N \frac{1}{\sqrt{2\pi \sigma_{\hat{x}_i}^2}} \exp\left(-\frac{(x_i - \hat{x}_i)^2}{2\sigma_{\hat{x}_i}^2}\right) \\
    p_\theta (\bm{z}) = \prod_{i=1}^M \frac{1}{\sqrt{2\pi}} \exp(z_i^2/2) \\
    q_\phi (\bm{z}|\bm{x}) = \prod_{i=1}^M \frac{1}{\sqrt{2\pi \sigma_{i}^2}} \exp \left(-\frac{(z_i-\mu_i)^2}{2\sigma_i^2} \right) \; .
\end{align}
\end{subequations}

This leads to the following expression for the ELBO:
\begin{eqnarray}
    \mathcal{L}_{\phi, \theta}(\bm{x}) &=& - \sum_{i=1}^N \langle (x_i - \hat{x}_i)^2  \rangle_{q_\phi (z|x)}
    - \sum_{i=1}^M \frac{1}{2} \left(\mu_i^2 + \sigma_i^2 - 1 - \ln \sigma_i^2 \right) + \text{const} \; .
\end{eqnarray}
To obtain the previous equality we assumed $\bm{\sigma}_{\hat{x}} = \bm{1}$. We also used the fact that $\langle z_i^2 \rangle_{q_\phi} = \mu_i^2 + \sigma_i^2$. While the goal is to maximize the ELBO, it is common practice to minimize the negative ELBO during training.
An important step in the context of VAEs is the so-called \textit{reparameterization trick}. When dealing with a finite data set and optimizing the loss function we need to compute gradients of expectations with respect to distributions that may not be explicitly expressible. For instance, consider computing the gradient of an estimator:
\begin{eqnarray}
    \nabla_\phi \langle f_\phi(z) \rangle_{q_\phi (z)} &=& \nabla_\phi \int dzq_\phi (z) f_\phi (z) \\
    &\sim& \nabla_\phi \sum_{z\sim q_\phi (z)} f_\phi (z)
\end{eqnarray}
The issue here is that taking the gradient over $\sum_{z\sim q_\phi (z)}$ is rather ill-defined because the samples are drawn from a distribution parameterized by $\phi$. To circumvent this, the reparameterization trick consists in a change in variable in a way that makes the sampling process differentiable with respect to $\phi$. This variable change needs to preserve the metric, \textit{i.e.}, the distribution in the old variable and that in the new variable need to both be normalized:
\begin{equation}
    \int dz q_\phi (z) = \int d\epsilon \rho(\epsilon) \implies \rho(\epsilon) = |\frac{dz}{d\epsilon}| q_\phi (z)
\end{equation}
The simplest change in variable is $z = \mu + \sigma \epsilon$ with $\epsilon \sim \mathcal{N}(1,0)$, which leads to
\begin{equation}
    \nabla_\phi \sum_{z\sim q_\phi (z)} f_\phi (z) \rightarrow \sum_{\epsilon\sim \mathcal{N}(0,1)} \nabla_\phi f_\phi (\mu_\phi + \sigma_\phi \epsilon)
\end{equation}
The \textit{reparameterization trick} is merely a change in variable akin to that used in the Box-Muller method to generate Gaussian distributed random numbers from Uniform distributed random numbers \cite{box1958note}. This is quite often used in the context of deep generative models in order to be able to take the gradient over an estimator.

\section{Discrete Variational Autoencoder}

Discrete Variational Autoencoders (DVAEs) are a type of VAE where the latent space is discrete. The main two challenges with DVAEs are \textit{i}) how does one backpropagate the gradient since the latent space is discrete? \textit{ii}) what reparameterization can be employed to enable gradient-based optimization? To address the former, one can simply relax the discrete condition by introducing annealed sigmoids. Specifically, we replace the Heaviside function $\Theta(x)$ with the sigmoid function $\sigma(x \beta)$, where $\beta$ is the annealing parameter. Notice that $\lim_{\beta \rightarrow \infty} \sigma(x\beta) = \Theta(x)$. To address the latter issue one can employ the \textit{Gumbel trick}. The \textit{Gumbel trick} has become an umbrella term which refers to a set of methods to sample from discrete probabilities or to estimate its partition function. In our case, we simply generate latent variables $\zeta$ via
\begin{equation}
    \zeta = \sigma( (l(\phi, x) + \sigma^{-1}(\rho)) \beta) \; ,
\end{equation}
where $\rho$ is a uniform random number, and $l(\phi,x)$ is a logit, \textit{i.e.}, the inverse of a sigmoid function, such that in the discrete regime of $\zeta$ (\textit{i.e.}, $\beta \rightarrow \infty$) $P(\zeta = 1) = \sigma(l(\phi,x))$. Notice that in this approach, we generate the random variable $\zeta$ using a deterministic equation, $\sigma$; a logit, $l(\phi,x)$; and a uniformly-distributed random number, $\rho$. The connection with Gumbel distributed random numbers is due to the fact that $\sigma^{-1}(\rho) \sim G_1 - G_2$, where $G_1$ and $G_2$ are two Gumbel distributed random numbers \cite{maddison2016concrete, balog2017lost, khoshaman2018gumbolt}. 
    
\section{Bipartite Restricted Boltzmann Machines}
Suppose a data set $\lbrace \bm{v}^{(i)} \rbrace_{i=1}^{|\mathcal{D}|}$, and each element in the data set lives in $\lbrace 0, 1 \rbrace^{ N}$.
The goal behind training a Restricted Boltzmann Machine (RBM) over this data set consists on fitting a probability mass function, $p(\bm{v})$, that models the distribution of the data. This is achieved by maximizing the log-likelihood (LL) of $p(\bm{v})$ over the data set. We denote the joint probability of the dataset as $P_{\mathcal{D}} = \left( \prod_{\bm{v} \in \mathcal{D}} p(\bm{v}) \right)^{1/|\mathcal{D}|}$. Maximizing the LL corresponds to:
\begin{equation}
    \argmax_{\bm{\Omega}} \ln P_{\mathcal{D}}
\end{equation}
By design, $p(\bm{v})$ follows a Boltzmann distribution \textit{viz}.
\begin{equation}
    p(\bm{v}) = \frac{\sum_{\bm{h}} e^{-E(\bm{v},\bm{h}; \bm{a},\bm{b},\bm{W})}}{Z(\bm{a},\bm{b},\bm{W})} \; ,
\end{equation}
where $Z(\bm{a},\bm{b},\bm{W})$ is the partition function and $E(\bm{v},\bm{h};\bm{a},\bm{b},\bm{W})$ is the energy function defined as 
\begin{equation}
    E(\bm{v},\bm{h};\bm{a},\bm{b},\bm{W}) = - \sum_i a_i v_i - \sum_j b_j h_j - \sum_{i,j} v_i W_{ij} h_j \; . \label{eq:energy}
\end{equation}
The parameters $\bm{a}$, $\bm{b}$ and $\bm{W}$ are fitting parameters and $\bm{h}$ is called the hidden vector such that $\bm{h} \in \lbrace 0, 1 \rbrace^{ M}$. Notice that the matrix $\bm{W}$ couples the nodes in $\bm{v}$ with the nodes in $\bm{h}$, while there are no explicit couplings between nodes in $\bm{v}$ nor between nodes in $\bm{h}$, which is the same to say that the RBM is a bipartite graph.

Notice that
\begin{equation}
    \frac{\partial E}{\partial w} = 
    \begin{cases}
        -v_k \qquad w = a_k, \\
        - h_k \qquad w = b_k, \\
        -v_k h_l \qquad w = W_{kl} \; .
    \end{cases}
\end{equation}

Taking the derivative of the LL with respect to some generic parameter $w$ yields:
\begin{equation}
    \frac{\partial \ln P_{\mathcal{D}} }{\partial w} = \frac{1}{|\mathcal{D}|} \sum_{i=1}^{|\mathcal{D}|} \langle - \frac{\partial E}{\partial w} \rangle_{p(\bm{h} | \bm{v}^{(i)})} - \langle - \frac{\partial E}{\partial w} \rangle_{p(\bm{v}, \bm{h})}
\end{equation}

The previous simplifies to
\begin{subequations}
\begin{align}
    \frac{\partial \ln P_{\mathcal{D}}}{\partial a_k} = \frac{1}{|\mathcal{D}|} \sum_{i=1}^{|\mathcal{D}|} \langle v_k \rangle_{p(\bm{h} | \bm{v}^{(i)})} - \langle v_k \rangle_{p(\bm{v}, \bm{h})} \\
    \frac{\partial \ln P_{\mathcal{D}}}{\partial b_k} = \frac{1}{|\mathcal{D}|} \sum_{i=1}^{|\mathcal{D}|} \langle h_k \rangle_{p(\bm{h} | \bm{v}^{(i)})} - \langle h_k \rangle_{p(\bm{v}, \bm{h})} \\
    \frac{\partial \ln P_{\mathcal{D}}}{\partial w_{kl}} = \frac{1}{|\mathcal{D}|} \sum_{i=1}^{|\mathcal{D}|} \langle v_k h_l \rangle_{p(\bm{h} | \bm{v}^{(i)})} - \langle v_k h_l \rangle_{p(\bm{v}, \bm{h})}
\end{align}
\end{subequations}
where 
\begin{equation}
    \langle \bullet \rangle_{p(\bm{h} | \bm{v}^{(i)})} = \frac{ \sum_{\bm{h}} \bullet e^{-E(\bm{v}^{(i)},\bm{h})} }{\sum_{\bm{h}} e^{-E(\bm{v}^{(i)},\bm{h})}} \label{eq:p(h|v)}
\end{equation}
and 
\begin{equation}
    \langle \bullet \rangle_{p(\bm{v},\bm{h})} = \frac{ \sum_{\bm{v},\bm{h}} \bullet e^{-E(\bm{v},\bm{h})} }{ \sum_{\bm{v},\bm{h}} e^{-E(\bm{v},\bm{h})}} \; . \label{eq:p(v,h)}
\end{equation}

Since the number of $\bm{v}$ and $\bm{h}$ states are $2^N$ and $2^M$, respectively, the number of terms in the summations in Eqs. \eqref{eq:p(h|v)} and \eqref{eq:p(v,h)} are $|\mathcal{D}| \times 2^M$ and $2^{N+M}$, respectively. This exponential dependence on the dimensionality makes computing these averages intractable. To overcome this challenge, importance sampling is employed.

Notice that $q(\bm{h}|\bm{v}) = p(\bm{v},\bm{h})/p(\bm{v})$, from which it is straightforward showing
\begin{equation}
    q(\bm{h}|\bm{v}) = \prod_{j=1}^M q(h_j|\bm{v}) \; ,
\end{equation}
where
\begin{eqnarray}
    q(h_j|\bm{v}) = \frac{e^{h_j C_j(\bm{v})}}{1+e^{C_j(\bm{v})}} \label{eq:twoLevelEq}
\end{eqnarray}
and $C_j(\bm{v}) = \sum_i v_i W_{ij} + b_j$. Therefore, $q(h_j=1|\bm{v})=\sigma(C_j(\bm{v}))$. Conversely, $p(v_i=1 |\bm{h}) = \sigma(D_i(\bm{h}))$, with $D_i(\bm{h}) = \sum_{j} W_{ij} h_j + a_i$. Hence, we can employ the expressions $\sigma(D_i(\bm{h}))$ and $\sigma(C_j(\bm{v}))$ to perform Gibbs sampling. 

We can further simplify the expectation values from Eq. \eqref{eq:p(h|v)}:
\begin{subequations}
\begin{align}
    \langle v_k \rangle_{p(\bm{h} | \bm{v}^{(i)})} = v_k \\
    \langle h_j \rangle_{p(\bm{h} | \bm{v}^{(i)})} = \frac{e^{C_j(\bm{v})}}{1+e^{C_j(\bm{v})}} \\
    \langle v_k h_j \rangle_{p(\bm{h} | \bm{v}^{(i)})} = v_k \frac{e^{C_j(\bm{v})}}{1+e^{C_j(\bm{v})}}\\
\end{align}
\end{subequations}

Note that $p(\bm{v},\bm{h})=p(\bm{v}|\bm{h})p(\bm{h})$ and $p(\bm{v},\bm{h})=q(\bm{h}|\bm{v})p(\bm{v})$. Therefore, starting from an observed data point, $\bm{v}$, we can generate samples of the hidden units, $\bm{v}$, via $p(\bm{h}|\bm{v})$. These samples can then be used as prior samples to generate new samples of $\bm{v}$ vectors. We can repeat the process a number of times $K$, called the number of Gibbs sampling steps. We denote this process as $(\bm{v},\bm{h}) \sim \left[ q(\bm{h}|\bm{v})p(\bm{v}|\bm{h}) \right]^K$

Therefore, for very large K, we can estimate Eq. \eqref{eq:p(v,h)} as:
\begin{equation}
    \langle \bullet \rangle_{p(\bm{v},\bm{h})} \simeq \frac{1}{N} \sum_{(v,h) \sim \left[ q(h|v)p(v|h) \right]^K} \bullet  \; . \label{eq:p(v,h)2}
\end{equation}
The Gibbs sampling number of steps ultimately should be larger than the mixing time. The mixing time will depend on the size of the RBM, the data set being used and, interestingly, it has been shown that as the number of updates during training increases, the Gibbs sampling number of steps must increase for the RBM to reach equilibrium and avoid getting stuck in an non-equilibrium state \cite{decelle2021equilibrium}.

The standard procedure to train an RBM involves partitioning the data set $\mathcal{D}$ into mini-batches $\mathcal{D}_\alpha$, such that $\mathcal{D} = \cup_\alpha \mathcal{D}_\alpha$.
Then the RBM parameters are updated by:
\begin{subequations}
\begin{align}
    a_{k}^{(t)} = a_{k}^{(t-1)} + \eta \frac{\partial \ln P_{\mathcal{D}_\alpha}}{\partial a_{k}} \; , \\
    b_{k}^{(t)} = b_{k}^{(t-1)} + \eta \frac{\partial \ln P_{\mathcal{D}_\alpha}}{\partial b_{k}} \; , \\
    W_{kl}^{(t)} = W_{kl}^{(t-1)} + \eta \frac{\partial \ln P_{\mathcal{D}_\alpha}}{\partial W_{kl}} \; ,
\end{align}
\end{subequations}
where $\eta$ is the learning rate.
When performing importance sampling, it is common to generate Markov chains of the size of the mini-batch, $|\mathcal{D}_\alpha |$.

There are three primary methods for training RBMs in the literature. Each one mainly differs from the others in the manner in which the Markov chains are initialized. The simplest one correspond to the case where for each parameter update, the initial state is randomly sampled from a $1/2$-Bernoulli distribution and is called \textit{Rdm-K}, where K is the number for Gibbs sampling steps. Another way shown to yield more robust RBMs is called \textit{Contrastive Divergence} (CD), whereby the MArkov chain is initialized from a point in the dataset. Lastly, persistent contrastive divergence (PCD) is very similar to CD in the sense that the Markov chain is started using a data point in the data set for the first parameter update, while for the remaining parameter updates, the Markov chains are initialized using the last state in the previous parameter update. This is similar to the traditional way to sample from an Ising model when decreasing the temperature. Instead of restarting the Markov chain from a random state after each temperature update, the chain is restarted from the previous state before the temperature update. 

\section{High temperature gradient approximation} \label{sec:HTGA}
The previous section shows the derivation of the block Gibbs sampling Eqs. used to trained RBM. A quite common approach to training RBMs consists in replacing $\ln Z$ with the average energy before computing the gradient. The basis comes from noticing that the gradient of the logarithm of the partition function w.r.t. the RBM parameters is equal to the average value of the gradient of the energy w.r.t. the RBM parameters, \textit{viz.},
\begin{equation}
    - \frac{\partial \ln Z}{\partial \phi} = \langle \frac{\partial E}{\partial \phi} \rangle
\end{equation}
This expectation value is over the ensemble and it is approximated by an arithmetic average over samples obtained via block Gibbs sampling, \textit{i.e.},
\begin{equation}
    \langle \frac{\partial E}{\partial \phi} \rangle \simeq \frac{1}{N} \sum_{\bm{z} \sim BGS} \frac{\partial E}{\partial \phi} = \frac{\partial}{\partial \phi} \frac{1}{N} \sum_{\bm{z} \sim BGS} E(\bm{z}) \; .
\end{equation}
The last equality is not general and only holds in certain scenarios, as we show here.
In the following we show that this approximation corresponds to the high temperature gradient approximation where thermal energy is larger than typical spin interactions, such that the specific heat is zero and the only contribution to the entropy is configurational.

By definition, the average energy is given by $\langle E(\bm{z}) \rangle = \sum_{\bm{z}} E(\bm{z})e^{-\beta E(\bm{z})}/Z(\beta)$. Deriving the energy w.r.t. some parameter $\phi$ the energy depends on leads to:
\begin{equation}
    \frac{\partial }{\partial \phi} \langle E(\bm{z}) \rangle = \langle \frac{\partial E(\bm{z})}{\partial \phi} \rangle + \beta \left( \langle E(\bm{z}) \rangle \langle \frac{ \partial E(\bm{z})}{\partial \phi } \rangle - \langle E(\bm{z}) \frac{\partial E(\bm{z})}{\partial \phi} \rangle \right)
\end{equation}
Hence $\frac{\partial }{\partial \phi} \langle E(\bm{z}) \rangle = \langle \frac{\partial E(\bm{z})}{\partial \phi} \rangle$ implies the equality:
\begin{equation}
    \langle E(\bm{z}) \rangle \langle \frac{ \partial E(\bm{z})}{\partial \phi } \rangle - \langle E(\bm{z}) \frac{\partial E(\bm{z})}{\partial \phi} \rangle = 0\; . \label{eq:condT}
\end{equation}
Recall the specific heat relates to the second cumulant via $C_T = \frac{1}{kT^2}\sigma_E^2$. Notice that $C_T$ depends on $\phi$. 
The derivative of the second cumulant w.r.t. $\phi$ leads to
\begin{eqnarray}
    \frac{\partial \sigma_E^2}{\partial \phi} = \beta \langle \frac{\partial E(\bm{z})}{\partial \phi} \rangle \left( \langle E(\bm{z})^2 \rangle - 2 \langle E(\bm{z}) \rangle^2 \right) + 2 \left( \langle E(\bm{z}) \frac{\partial E(\bm{z})}{\partial \phi} \rangle - \langle E(\bm{z}) \rangle \langle \frac{\partial E(\bm{z})}{\partial \phi} \rangle \right) \\
    - \beta \left( \langle E(\bm{z})^2 \frac{\partial E(\bm{z})}{\partial \phi} \rangle - 2 \langle E(\bm{z}) \rangle \langle E(\bm{z}) \frac{\partial E(\bm{z})}{\partial \phi} \rangle \right)
\end{eqnarray}
From the previous Eq. it is easy to show that when Eq. \eqref{eq:condT} is satisfied, $\frac{\partial C_T}{\partial \phi} = 0$. In general, the previous occurs at very large temperatures where $C_T=0$, \textit{i.e.}, the energy of the system saturates such that increasing the temperature does not increase the energy. In such regime, the spins are uncorrelated and the entropy is solely defined by the logarithm of possible configurations.

\section{Quadripartite RBM numerical verification}
\renewcommand{\thefigure}{E\arabic{figure}}

\begin{figure}[hbtp]
\centering
\includegraphics[width=3.2in]{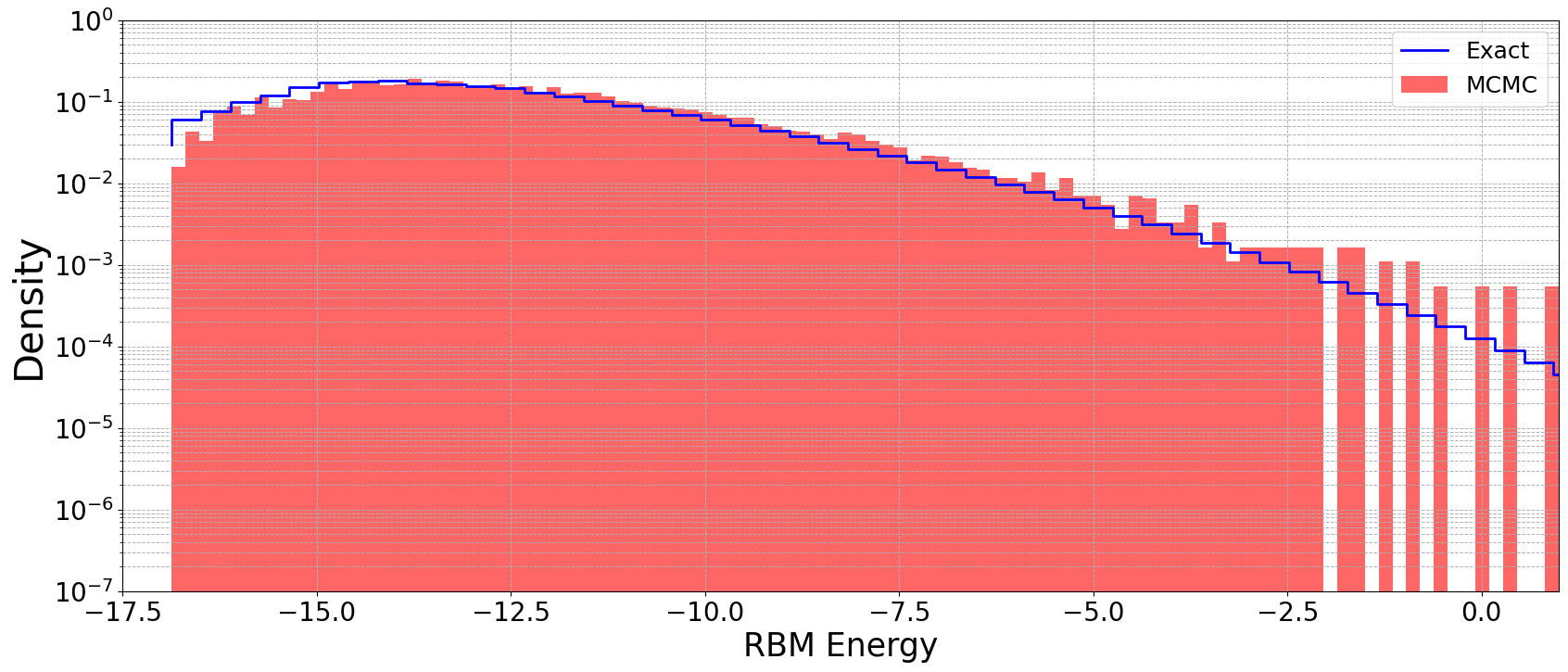}
\includegraphics[width=3.2in]{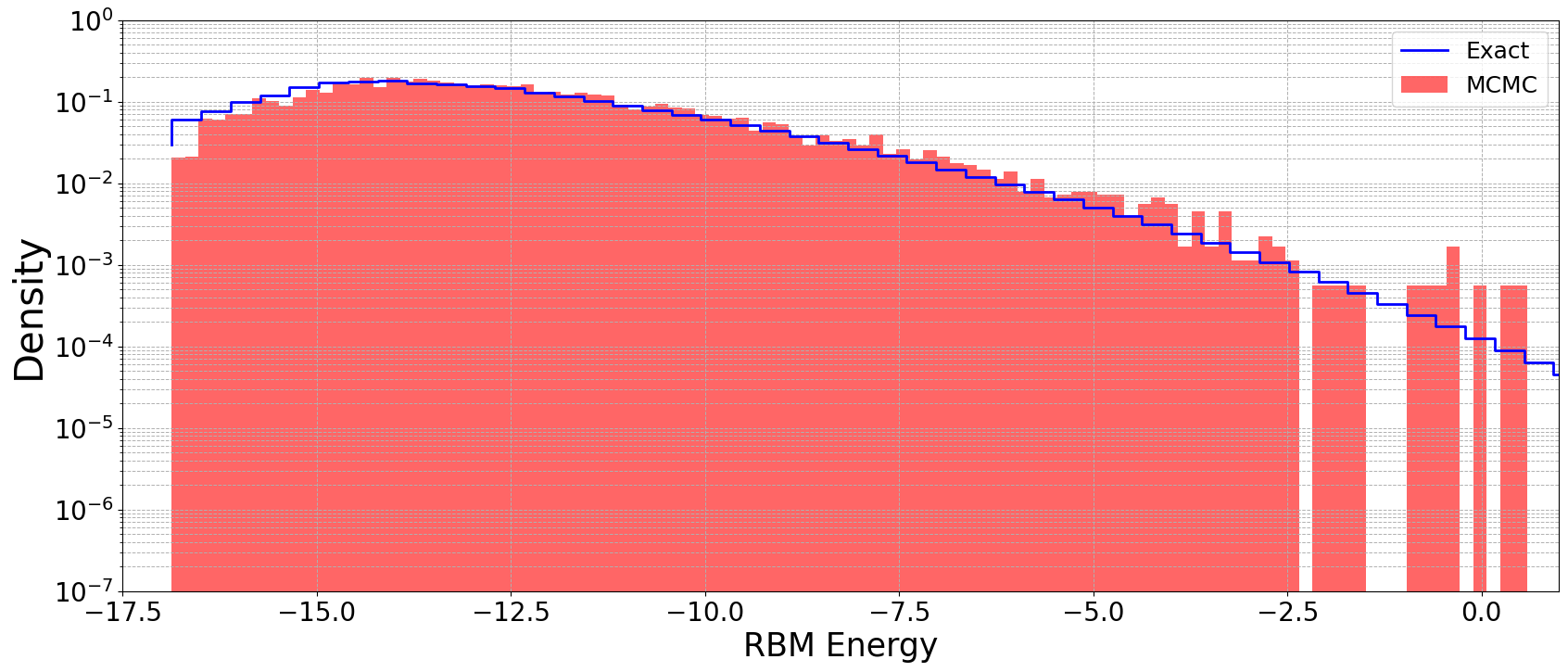}
\includegraphics[width=3.2in]{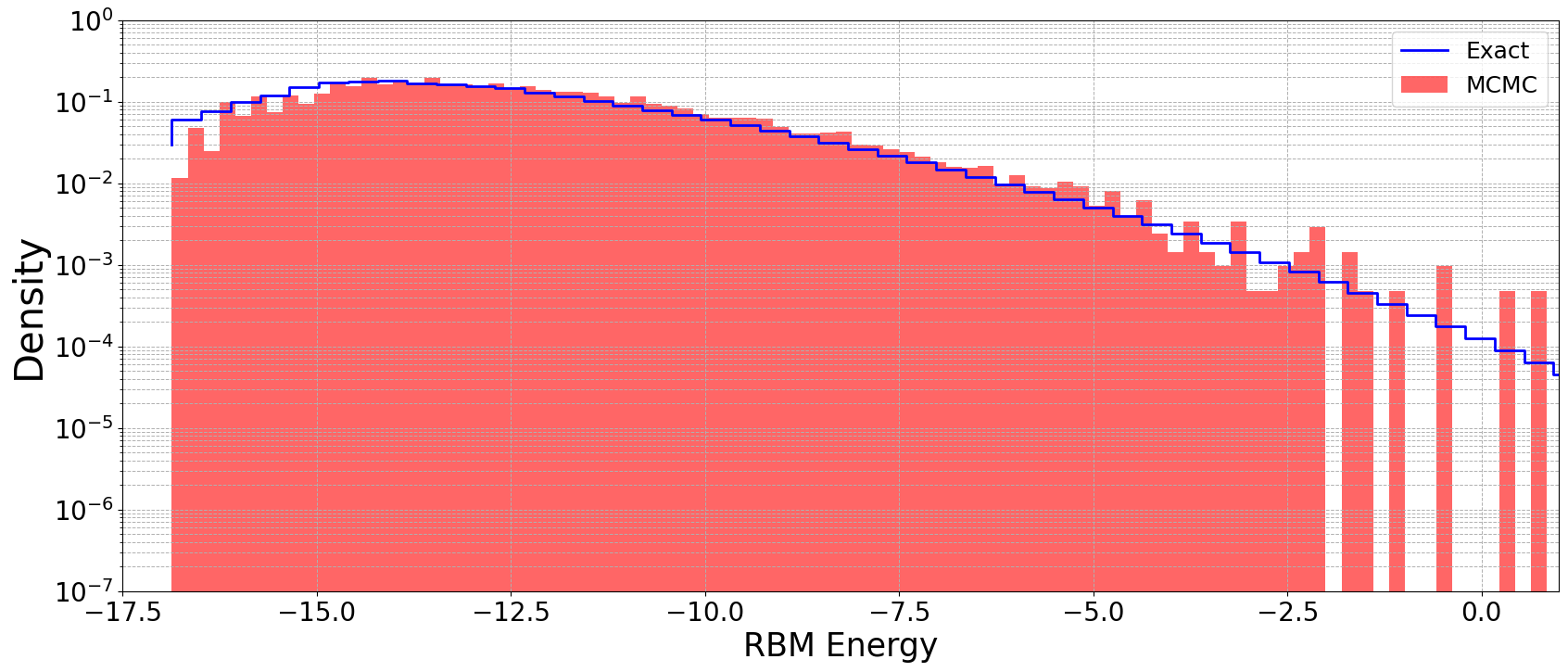}
\caption{Density of states for 4-partite RBMs with $6 \times 4$ nodes and $N = 10240$ samples, with Gibbs sampling steps set to \textbf{(upper left panel)} 200, \textbf{(upper right panel)} 500, and \textbf{(lower panel)} 3000 units.}\label{fig:rbm4pverification}
\end{figure}

We considered 4-partite RBM with six nodes per partition. This setting allows for the explicit enumeration of all feasible states, facilitating the precise computation of the partition function. To rigorously assess the accuracy of our approach, we conducted a comparative analysis of the density of energy states. This entailed a direct comparison between the utilization of all feasible states and the implementation of the 4-partite Gibbs sampling method, as elaborated upon in Section \ref{sec:4pRBM}. In Figure \ref{fig:rbm4pverification}, we present a detailed visual comparison of the density of states obtained through both methodologies, across various iterations of the Gibbs sampling process. This comparison shows the convergence and consistency of these two approaches.

\section{Quantum Annealers} \label{App:QA}
\subsection{Adiabatic Approximation} \label{sec:adApprox}
Here we derive the adiabatic approximation following \cite{sakurai2017jim}, which is the theoretical foundation of quantum annealers. Let us suppose a time-dependent Hamiltonian $H(t)$, we denote the time-dependent eigenstates as $|n; t \rangle$ and the eigenvalues as $E_n(t)$, such that,
\begin{equation}
    H(t)|n;t\rangle = E_n(t)|n,t\rangle \; , \label{eq:char_eq}
\end{equation}
which simply states that at any particular time $t$, the eigenstate and eigenvalue may change. Notice that one can write the general solution to Schr\"odinger's Eq., \textit{viz.},
\begin{eqnarray}
    i \hbar \frac{\partial }{\partial t} |\alpha; t \rangle = H(t) | \alpha; t \rangle \label{eq:schro}
\end{eqnarray}
as
\begin{equation}
    |\alpha; t \rangle = \sum_n c_n(t) e^{i\theta(t)} |n;t\rangle \label{eq:alpha_state}
\end{equation}
with
\begin{eqnarray}
    \theta_n(t) = - \frac{1}{\hbar} \int_0^t E_n(t')dt' \; . \label{eq:theta_energy}
\end{eqnarray}
By substituting Eqs. \eqref{eq:alpha_state} and \eqref{eq:theta_energy} in Eq. \eqref{eq:schro} we reach the following equality:
\begin{equation}
    \sum_n e^{i \theta_n(t)} \left[ \dot{c}_n(t) |n;t \rangle + c_n(t) \frac{\partial}{\partial t} | n;t\rangle \right] = 0
\end{equation}
By taking the inner product w.r.t. $\langle m;t|$ and invoking orthonormality, yields the following differential equation for the time-dependent coefficients:
\begin{equation}
    \dot{c}_m(t) = -\sum_n c_n(t) e^{i (\theta_n(t)- \theta_m(t))} \langle m;t|\frac{\partial}{\partial t}|n;t\rangle \; . \label{eq:c_n_diff}
\end{equation}
To get a better sense of the $\langle m;t|\frac{\partial}{\partial t}|n;t\rangle$ term, let us take the time derivative of the characteristic Eq. \eqref{eq:char_eq}:
\begin{eqnarray}
    \langle m; t|\frac{\partial}{\partial t} \left[H(t)|n;t\rangle = E_n(t)|n,t\rangle \right] \implies \nonumber \\
    \langle m; t| \left[ \dot{H}(t)|n;t\rangle + H(t)\frac{\partial}{\partial t}|n;t\rangle = \dot{E}_n(t)|n;t\rangle  \right. \\ 
    \left. + E_n(t)\frac{\partial}{\partial t}|n;t\rangle \right] \implies \nonumber \\
    \langle m; t| \dot{H}(t)|n;t\rangle = \left(E_n(t) - E_m(t) \right)\langle m;t| \frac{\partial}{\partial t}|n;t\rangle \nonumber  \\
\end{eqnarray}
For $m\neq n$, we can write
\begin{equation}
    \langle m;t | \frac{\partial}{\partial t}|n;t\rangle = \frac{\langle m; t| \dot{H}(t)|n;t\rangle}{E_n(t) - E_m(t)} \; .
\end{equation}
Substituting the previous Eq. in Eq. \eqref{eq:c_n_diff} leads to:
\begin{eqnarray}
    \dot{c}_m(t) =  - c_m(t) \langle m;t|\frac{\partial}{\partial t}|m;t\rangle \\
    -\sum_{n\neq m} c_n(t) e^{i (\theta_n(t)- \theta_m(t))}\frac{\langle m; t| \dot{H}(t)|n;t\rangle}{E_n(t) - E_m(t)} \; . 
\end{eqnarray}
The previous Eq. shows that states with $n\neq m$ will mix with $|m;t\rangle$ due to the time dependence of the Hamiltonian.

The adiabatic approximation consists in neglecting the mixing terms which correspond to the regime whereby
\begin{equation}
    \frac{|\langle m; t| \dot{H}(t)|n;t\rangle|}{E_n(t) - E_m(t)} \equiv \frac{1}{\tau} \ll \langle m;t|\frac{\partial}{\partial t}|m;t\rangle \sim \frac{E_m}{\hbar} \; . \label{eq:adiabatic_approx}
\end{equation}
The previous gives us the condition where the adiabatic approximation holds, \textit{i.e.}, that in which the timescale $\tau$ for changes in the Hamiltonian is much larger than the inverse of the characteristic frequency of the state phase factor. In such regime,
\begin{equation}
    c_n(t) = e^{i \gamma(t)}c_n(0)
\end{equation}
with
\begin{equation}
    \gamma_n(t) \equiv i \int_0^t dt' \langle n;t' | \frac{\partial}{\partial t'} |n;t'\rangle
\end{equation}
Notice that
\begin{equation}
    0 = \frac{\partial}{\partial t} \langle n;t' | n;t'\rangle = \left[\frac{\partial}{\partial t} \langle n;t' | \right] | n;t'\rangle + \langle n;t | \frac{\partial}{\partial t} |n;t\rangle
\end{equation}
which implies that
\begin{equation}
    \left(  \langle n;t | \frac{\partial}{\partial t} |n;t\rangle \right)^* = -  \langle n;t | \frac{\partial}{\partial t} |n;t\rangle
\end{equation}
Therefore, the integral argument is imaginary in which case $\gamma(t)$ is real. Hence, in the adiabatic approximation ($\tau \gg 1/\omega_n$), if the system starts out in eigenstate $|n;0\rangle$, it will remain there since $c_n(t) = e^{i \gamma(t)} c_n(0)$ and $c_l(t)=0$ for all $l \neq n$. Finally, it is important to stress that the adiabatic approximation does not correspond to short time regimes, \textit{i.e.}, the time $t$ is not relevant here but only the Hamiltonian change rate and the characteristic time of the state phase factor.


\subsection{Dwave parameter mapping}
This subsection shows the explicit parameter mapping between an RBM and a QA. Recall the quadripartite RBM energy function is:
\begin{eqnarray}
    E(\mathbf{v},\mathbf{h},\mathbf{s},\mathbf{t}) = - a_i v_i - b_i h_i - c_i s_i - d_i t_i \nonumber \\
    - v_i W^{(0,1)}_{ij} h_j - v_i W^{(0,2)}_{ij} s_j \nonumber \\
    - v_i W^{(0,3)}_{ij} t_j - h_i W^{(1,2)}_{ij} s_j \nonumber \\
    - h_i W^{(1,3)}_{ij} t_j - s_i W^{(2,3)}_{ij} t_j \; , \label{eq:4pRBMEnergy}
\end{eqnarray}
where we are using the double indices convention for summation. Since RBM data values are $0$s and $1$s, while qubits can take the values $\lbrace -1, 1 \rbrace$, we map the quantum states to RBM states as:
\begin{equation}
    \begin{pmatrix}
        \mathbf{v} \\
        \mathbf{h} \\
        \mathbf{s} \\
        \mathbf{t} \\
    \end{pmatrix}
    = \frac{1}{2}
    \begin{pmatrix}
        \mathbf{z_v} + \mathbf{1} \\
        \mathbf{z_h} + \mathbf{1} \\
        \mathbf{z_s} + \mathbf{1} \\
        \mathbf{z_t} + \mathbf{1} \\
    \end{pmatrix} \label{eq:QAtoRBMMapping}
\end{equation}
By substituting Eq. \eqref{eq:QAtoRBMMapping} in Eq. \eqref{eq:4pRBMEnergy} and reading out the new couplings and biases, we obtain:
\begin{eqnarray}
    \Delta_i &=& - \frac{a_i}{2} \nonumber \\ &-& \frac{1}{4} \left( \sum_j W_{ij}^{(01)} + W_{ij}^{(02)} + W_{ij}^{(03)} \right) , i \in \Phi_0 \nonumber \\
    \Delta_i &=& - \frac{b_i}{2} \nonumber \\ &-& \frac{1}{4} \left( \sum_j (W^{(01)})^{t}_{ij} + W_{ij}^{(12)} + W_{ij}^{(13)} \right), i \in \Phi_1 \nonumber \\
    \Delta_i &=& - \frac{c_i}{2} \nonumber \\ &-& \frac{1}{4} \left( \sum_j (W^{(02)})^{t}_{ij} + (W^{(12)})^{t}_{ij} + W_{ij}^{(23)} \right), i \in \Phi_2 \nonumber \\
    \Delta_i &=& - \frac{d_i}{2} \nonumber \\ &-& \frac{1}{4} \left( \sum_j (W^{(03)})^{t}_{ij} + (W^{(13)})^{t}_{ij} + (W^{(23)})^{t}_{ij} \right), i \in \Phi_3 \nonumber \\
    J_{ij} &=& - \frac{W_{ij}^{(\gamma,\delta)}}{4}, i \in \Phi_\gamma \text{ and } j \in \Phi_\delta \label{eq:DwaveMap}
\end{eqnarray}
where $\Phi_i$ denotes the partition  $i$ and $(\bullet)^{t}$ denotes transpose. In addition, this mapping introduces an energy offset, $H_{o}$, in the Hamiltonian
\begin{equation}
    H_{o} = - \left( \frac{1}{2} \sum_i a_i + b_i + c_i + d_i + \frac{1}{4} \sum_j \sum_{\gamma<\delta} W_{ij}^{(\gamma, \delta)}   \right) \; ,
\end{equation}
which we ignore  since it does not contribute to the state probability distribution. After applying this transformation, we obtain the new RBM Hamiltonian:
\begin{eqnarray}
    H_{RBM} = \sum_i \Delta^{RBM}_i z_i + \sum_{ij} J_{ij}^{RBM} z_i z_j \; .
\end{eqnarray}

\subsection{DWave $\beta_{QA}$ parameter estimation: Method 1}
Let us assume two RBMs which we denote as $QA$ and $B$, both, described by the same Hamiltonian at different temperatures, \textit{viz.}: 
\begin{eqnarray}
    P_{QA}(x) &=& \frac{e^{-\beta_{QA} H(x)}}{Z(\beta_{QA})} \; , \\
    P_{B}(x) &=& \frac{e^{-\beta H(x)}}{Z(\beta)} \; .
\end{eqnarray}
We denote as $\beta_{QA}$ and $\beta$ the inverse temperatures of system $QA$ and $B$, respectively. The Kullback-Liebler divergence associated to these two system yields:
\begin{equation}
    D_{KL}(P_{QA} || P_{B} ) = (\beta - \beta_{QA}) \langle H \rangle_{QA} + \ln \frac{Z(\beta)}{Z(\beta_{QA})} \; ,
\end{equation}
from which it is trivial to show that $\beta=\beta_{QA}$ yields zero in the KL divergence. The KL divergence derivative w.r.t. $\beta$ yields
\begin{equation}
    \frac{\partial D_{KL}}{\partial \beta} = \langle H \rangle_{QA} - \langle H \rangle_{B(\beta)} \; ,
\end{equation}
where we have made explicit the $\beta$ dependence of system $B$. We can fit $\beta$ through gradient descent using the KL divergence, which leads to:
\begin{eqnarray}
    \beta_{t+1} = \beta_t - \eta \left( \langle H(x) \rangle_{QA} - \langle H(x) \rangle_{B(\beta)} \right) \label{eq:beta_app}
\end{eqnarray}
The pseudo-algorithm to fit $\beta$ using Eq. \eqref{eq:beta_app} is:
\begin{enumerate}
    \item Fix learning rate $\eta$ and initialize $\beta_0$. Parse the RBM parameters onto the Quantum Annealer via Eqs. \eqref{eq:DwaveMap}.
    \item Sample from system $QA$ RBM and from system $B$ at temperature $1/\beta_0$.
    \item Compute expectation of $H(x)$ using the samples from $QA$ and from $B$, respectively.
    \item Update $\beta$ using Eq. \eqref{eq:beta_app}.
    \item Repeat steps 2 to 4 until a convergence criterion is fulfilled..
\end{enumerate}
Afterwards, $\beta_T \approx \beta_{QA}$. Therefore, rescaling the Hamiltonian by $1/\beta_T$ and then parsing the new Hamiltonian parameters onto the $QA$ via Eqs. \eqref{eq:DwaveMap} will ensure that we are effectively sampling from $H$.
Notice that in the previous method, $\langle H(x) \rangle_{QA}$ is independent of $\beta$, hence in the previous algorithm we need to generate samples only from system $B$ every time we update $\beta$. This can be rather inconvenient, for instance, in the case where our interest is in having the QA mimic B, where the latter is a trained RBM. Changing the temperature of system $B$ to match that of QA will affect the performance of the model. We might be tempted to invert the method and fit $\beta_{QA}$, but this is not a viable approach, since one does not have control over $\beta_{QA}$ let alone a measurement of it. Instead, one can do the following: Replace the original Hamiltonian with that scaled by $\beta$, \textit{i.e.}, $H(x) \rightarrow H(x)/\beta$, which leads to:
\begin{eqnarray}
    \beta_{t+1} = \beta_t - \eta \left( \langle H(x) \rangle_{QA^{(r)}} - \langle H(x) \rangle_{B(1)} \right) \label{eq:beta_app_2}
\end{eqnarray}
where $QA^{(r)}$ correspond to rescaling $H(x)$ by $1/\beta_t$. In reaching the previous equation, we redefined $\frac{\eta}{\beta_t} \rightarrow \eta$ where $\eta$ is fixed.
The pseudo-algorithm to fit $\beta$ using Eq. \eqref{eq:beta_app_2} is:
\begin{enumerate}
    \item Fix learning rate $\eta$ and initialize $\beta_0$. Parse the RBM parameters onto the Quantum Annealer via multiplying Eqs. \eqref{eq:DwaveMap} by $1/\beta_0$.
    \item Sample from the $QA$ RBM and from the $B$ RBM at temperature $1$.
    \item Compute expectation of $H(x)/\beta_0$ using the samples from $QA$ and from $B$, respectively.
    \item Update $\beta_1$ using Eq. \eqref{eq:beta_app_2}.
    \item Repeat steps 2 to 4 until a convergence criterion is fulfilled.
\end{enumerate}

The previous method is one of the common approaches used to estimate the $\beta_{QA}$ parameter in QAs. However, it can be slow to converge which is why we propose a simple mapping with a stable fixed point at $\beta_{QA}$. We describe the method in full detail in the following.

\subsection{DWave $\beta_{QA}$ parameter estimation: Method 2}

Once again, let us assume two RBMs which we denote as $QA$ and $B$, both, described by the same Hamiltonian at different temperatures, \textit{viz.}: 
\begin{eqnarray}
    P_{QA}(x) &=& \frac{e^{-\beta_{QA} H(x)}}{Z(\beta_{QA})} \; , \\
    P_{B}(x) &=& \frac{e^{-\beta H(x)}}{Z(\beta)} \; .
\end{eqnarray}
We denote as $\beta_{QA}$ and $\beta$ the inverse temperatures of system $QA$ and $B$, respectively. Now, let us denote as $S_{QA}$ and $S_{B}$ as the entropy of QA and B, respectively, and assume $S_{QA} = S_B$, from which after some straightforward algebra:
\begin{eqnarray}
    \beta = \beta_{QA} \frac{\langle H \rangle_{QA}}{\langle H \rangle_{B(\beta)}} + \frac{\ln \frac{Z(\beta_{QA})}{Z(\beta)}}{\langle H \rangle_{B(\beta)}} \; .
\end{eqnarray}
We can further simplify the previous expression by introducing the variable $\Delta \beta = \beta_{QA} - \beta$:
\begin{eqnarray}
    \beta = \beta_{QA} \frac{\langle H \rangle_{QA}}{\langle H \rangle_{B(\beta)}} + \frac{\ln \langle e^{- \Delta \beta H} \rangle_{B(\beta)}}{\langle H \rangle_{B(\beta)}} \; . \label{eq:beta_mapping}
\end{eqnarray}
Notice that the r.h.s. of Eq. \eqref{eq:beta_mapping} has a fixed point at $\beta=\beta_{QA}$. Here on we will only keep the first term in the r.h.s. and we will show that the fixed point is stable. In addition, same as we did when deriving the previous method, we replace $H(x) \rightarrow H(x)/\beta$. Since we do not have any control over $\beta_{QA}$ nor we know the value \textit{a priori}, we replace the prefactor in the first term of the r.h.s. with $\beta$ since it does not affect the fixed point value and we further introduce a stability parameter $\delta (> 0)$. After the previous considerations, we propose the following mapping:
\begin{equation}
    \beta_{t+1} = f_{\delta}(\beta_t)\equiv \beta_t \left( \frac{\langle H \rangle_{QA^{(r)}}}{\langle H \rangle_{B(1)}} \right)^{\delta}
\end{equation}
The function $f_{\delta}$ has a fixed point at $\beta=\beta_{QA}$. The stability condition close to the fixed point correspond to $|f_{\delta}'(\beta_{QA})|< 1$. The first derivative at the fixed point yields:
\begin{eqnarray}
    |f_{\delta}'(\beta_{QA})| = 
    \begin{cases}
        |1+ \frac{\sigma^2_{QA}}{\langle H \rangle_{B(1)}}|, \; \delta=1 \\
        |1+ \delta \frac{\sigma^2_{QA}}{\langle H \rangle_{QA}}|, \; \delta \neq 1 \; .
    \end{cases} \label{eq:stability}
\end{eqnarray}
In Fig. \ref{fig:stability_plot} we have plotted Eq. \eqref{eq:stability} \textit{vs} $\beta$ for different values of $\delta$. The values of $\beta$ chosen for this plot correspond to where we typically find the fixed point. We call $\delta$ a stability parameter since we can tune it to stabilize the mapping per iteration. 

A similar analysis can be done for the previous method. Specifically, the stability condition becomes:
\begin{equation}
    |1-\frac{\sigma_{QA}^2}{\beta_{QA}/\eta}| < 1 \; .
\end{equation}
From the previous it is easy to notice that the fixed point is unstable when the learning rate, $\eta$, such that $\eta > \beta_{QA}/\sigma_{QA}^2$ ($\beta_{QA}/\sigma_{QA}^2 \sim 2 \cdot 10^{-2}$).

\section{Incident energy conditioning} \label{AppSec:IncEn}
In the conditioned Calo4pQVAE framework, we condition the latent space RBM using the incident energy. We perform the condition as follows:
\begin{enumerate}
    \item By applying a floor function on the incident energy in MeV, bin the incident energy, $e$, $e_{bin} = \text{floor}(e)$; the logarithm of the incident energy multiplied by $10$, $e_{bin}^{\ln} = \text{floor}(10\cdot \ln e)$; and the square root of the incident energy multiplied by $10$, $e_{bin}^{\sqrt{}} = \text{floor}(10 \cdot \sqrt{e})$.
    \item Convert to binary number the three previous binned numbers, $B_{e}=\text{binary}(e_{bin})$, $B_{\ln e}=\text{binary}(e_{bin}^{\ln})$, $B_{\sqrt{e}}=\text{binary}(e_{bin}^{\sqrt{}})$. We allocate 20 bits for each of these binary numbers.
    \item Concatenate the three binary numbers, $B=\text{cat}(B_{e},B_{\ln e}, B_{\sqrt{e}})$.
    \item Use one partition to fit as many repetitions of the concatenated binary number $B$.
    \item Set residual nodes to zero.
\end{enumerate}
We fixed the number of nodes per partition to $512$, hence the binary number $B$ fits 8 times and the number of residual nodes is 32.

\section{Gaussian approximation to shower logits} \label{AppSec:GaussApprox}
In the main text we describe the data transformation used to train our model, where we first reduce the voxel energy per event by dividing it by the incident energy and we afterwards construct logits based on the reduced energy random variable. The number of particles in the electromagnetic shower follows approximately a Poisson distribution. Furthermore, via the saddle point approximation, for large number of particles in the shower the multivariate Poisson distribution becomes a multivariate Gaussian distribution with the mean equal to the variance. Here we show that to zeroth approximation, the logits are Gaussian distributed.

Let us consider a Gaussian positive distributed random variable $r$ with mean and variance $\Lambda$, \textit{i.e.},
\begin{equation}
    f(r)=\frac{\mathcal{N}(r|\Lambda, \Lambda)}{\Omega} \; , \forall r>0
\end{equation}
where $\Omega$ is a normalization constant. We define $u=\ln \frac{x}{1-x}$ with $x=r/R$ and $R\gg r$. To zeroth order approximation, $u \approx \ln r - \ln R$. To obtain the distribution of $u$ we first introduce an auxiliary random variable $z=\ln r$ with distribution $g(z)$. By equating the cumulatives of $r$ and $z$ we obtain:
\begin{equation}
    g(z)=e^z f(e^z) \; .
\end{equation}
The distribution of $u$, $h(u)$, is simply the distribution of $z$ shifted by $\ln R$, namely, $h(u)=g(u+\ln R)$:
\begin{equation}
    h(u)=\frac{R e^u}{\Omega} \frac{1}{\sqrt{2\pi \Lambda}} e^{-\frac{(Re^u - \Lambda)^2}{2\Lambda}} \; .
\end{equation}
The previous distribution is highly sensitive to $u$ and the main contribution comes from $Re^u \approx \Lambda$. Hence, we can expand $\ln r$ around $\ln \Lambda$:
\begin{equation}
    \ln r \approx \ln \Lambda + \frac{r-\Lambda}{\Lambda} \; ,
\end{equation}
which translates to $u \approx \ln \frac{\Lambda}{R}+\frac{r-\Lambda}{\Lambda}$.
Notice that since $\frac{r-\Lambda}{\Lambda} \sim \mathcal{N}(0,\frac{1}{\Lambda})$, then:
\begin{equation}
    u \sim \mathcal{N}(\ln \frac{\Lambda}{R}, \frac{1}{\Lambda}) \; .
\end{equation}




\end{document}